%% file: main.tex
\documentclass{article}

\PassOptionsToPackage{numbers, compress}{natbib}

\usepackage[dandb, final]{neurips_2025}

\usepackage[table]{xcolor}  

\usepackage[utf8]{inputenc} 
\usepackage[T1]{fontenc}    
\usepackage{hyperref}       
\usepackage{url}            
\usepackage{booktabs}       
\usepackage{amsfonts}       
\usepackage{nicefrac}       
\usepackage{microtype}      
\usepackage{longtable}
\usepackage{graphicx}
\usepackage{array}
\usepackage{multirow}
\usepackage{xspace}
\usepackage{makecell}
\usepackage{boldline}
\usepackage{pifont}
\usepackage{tabularx}
\usepackage{graphicx}
\usepackage{wrapfig}
\usepackage[export]{adjustbox}
\usepackage{fancyvrb}
\usepackage{fvextra}
\usepackage[most,breakable]{tcolorbox}
\usepackage{caption}
\usepackage{pgfplots}
\pgfplotsset{compat=newest}
\usepgfplotslibrary{units}
\usepackage{subcaption}
\usepackage{listings}
\usepackage{siunitx}
\usepackage{hhline}
\usepackage{float}

\definecolor{lightblue}{rgb}{0.88, 0.95, 1.0}  
\definecolor{lightgreen}{rgb}{0.56, 0.93, 0.56} 
\definecolor{lightyellow}{rgb}{1.0, 1.0, 0.88}  
\definecolor{lightpink}{rgb}{1.0, 0.71, 0.76}   
\definecolor{lightorange}{rgb}{1.0, 0.83, 0.62} 
\definecolor{lightcyan}{rgb}{0.88, 1.0, 1.0}   
\definecolor{lightpurple}{rgb}{0.9, 0.8, 1.0}   
\definecolor{lightgray}{rgb}{0.83, 0.83, 0.83}  
\definecolor{lightgrey}{rgb}{0.83, 0.83, 0.83}  
\definecolor{lavender}{rgb}{0.9, 0.8, 1.0}       
\definecolor{lilac}{rgb}{0.78, 0.64, 0.8}        
\definecolor{periwinkle}{rgb}{0.8, 0.8, 1.0}     
\definecolor{mauve}{rgb}{0.87, 0.63, 0.87}       
\definecolor{orchid}{rgb}{0.85, 0.44, 0.84}      
\definecolor{amethyst}{rgb}{0.6, 0.4, 0.8}       
\definecolor{wisteria}{rgb}{0.79, 0.63, 0.86}    
\definecolor{dustylavender}{rgb}{0.76, 0.7, 0.86} 
\definecolor{frenchlavender}{rgb}{0.8, 0.6, 0.7} 
\definecolor{heliotrope}{rgb}{0.87, 0.73, 1.0}   
\definecolor{plum}{rgb}{0.87, 0.63, 0.87}        

\newcommand{\ours}[0]{\textsc{Jedi}\xspace}
\newcommand{\ourmodelthreebillion}[0]{\textsc{Jedi}-3B\xspace}
\newcommand{\ourmodelsevenbillion}[0]{\textsc{Jedi}-7B\xspace}
\newcommand{\ourbenchmarkname}[0]{\textsc{OSWorld-G}\xspace}

\newcommand{\numexamples}[0]{564\xspace}

\newcommand{\numinfeasibleexamples}[0]{54\xspace}
\newcommand{\numdataexamples}[0]{4 million\xspace}

\definecolor{softred}{rgb}{0.7,0.2,0.2}
\definecolor{softgreen}{rgb}{0.2,0.6,0.3}
\newcommand{\greencheck}{\textcolor{softgreen}{\checkmark}}
\newcommand{\redcross}{\textcolor{softred}{\ding{55}}}

\lstdefinelanguage{json}{
    basicstyle=\ttfamily\small,
    numbers=left,
    numberstyle=\tiny,
    stepnumber=1,
    numbersep=5pt,
    showstringspaces=false,
    breaklines=true,
    frame=single,
    backgroundcolor=\color{gray!10},
    string=[s]{"}{"},
    morestring=[b]',
    literate=
     *{0}{{{\color{blue}0}}}{1}
      {1}{{{\color{blue}1}}}{1}
      {2}{{{\color{blue}2}}}{1}
      {3}{{{\color{blue}3}}}{1}
      {4}{{{\color{blue}4}}}{1}
      {5}{{{\color{blue}5}}}{1}
      {6}{{{\color{blue}6}}}{1}
      {7}{{{\color{blue}7}}}{1}
      {8}{{{\color{blue}8}}}{1}
      {9}{{{\color{blue}9}}}{1},
}

\title{
Scaling Computer-Use Grounding via User Interface Decomposition and Synthesis
}

\author{
    \small
  {\bf
    Tianbao Xie
    \thanks{\ \ Equal contribution. \dag Corresponding authors. Work mainly done during TX's internship in Salesforce.}\ \ $^{{\color{purple}\boldsymbol{h}}}$ 
    Jiaqi Deng
    $^*$$^{{\color{purple}\boldsymbol{h}}}$
    Xiaochuan Li
    $^*$$^{{\color{purple}\boldsymbol{h}}}$
    Junlin Yang
    $^*$$^{{\color{purple}\boldsymbol{h}}}$
    Haoyuan Wu
    $^{{\color{purple}\boldsymbol{h}}}$
    Jixuan Chen
    $^{{\color{purple}\boldsymbol{h}}}$
    \vspace{3pt}
  } \\
    \small
  {
  \bf
    Wenjing Hu
    $^{{\color{purple}\boldsymbol{h}}}$
    Xinyuan Wang
    $^{{\color{purple}\boldsymbol{h}}}$
    Yuhui Xu
    $^{{\color{purple}\boldsymbol{s}}}$
    Zekun Wang
    $^{{\color{purple}\boldsymbol{h}}}$
    Yiheng Xu
    $^{{\color{purple}\boldsymbol{h}}}$
    Junli Wang
    $^{{\color{purple}\boldsymbol{h}}}$
    \vspace{3pt}
  } \\
    \small
  {
  \bf
    Doyen Sahoo
    $^{{\color{purple}\boldsymbol{s}}}$
    \vspace{1pt}
    Tao Yu
    $^{\dagger}$$^{{\color{purple}\boldsymbol{h}}}$
    \vspace{1pt}
    Caiming Xiong
    $^{\dagger}$$^{{\color{purple}\boldsymbol{s}}}$
    \vspace{1pt}
  } \\
\small
    {
    \hspace{-12pt}$^{\color{purple}\boldsymbol{h}}$ The University of Hong Kong \quad
    $^{\color{purple}\boldsymbol{s}}$Salesforce AI Research \quad
    }
}

\begin{document}

\makeatletter

\maketitle

\input{text/abstract}
\input{text/intro}
\input{text/approach}

\input{text/experiments}
\input{text/analysis}
\input{text/related}

\input{text/conclusion}
\input{text/acknowledgements}
\clearpage
\input{text/limitations}

\bibliographystyle{plainnat}
\bibliography{bibliography}

\clearpage



\newpage
\appendix
\input{text/appendix}

\end{document}

%% file: text/abstract.tex
\begin{abstract}

Graphical user interface (GUI) grounding, the ability to map natural language instructions to specific actions on graphical user interfaces, remains a critical bottleneck in computer use agent development. 
Current benchmarks oversimplify grounding tasks as short referring expressions, failing to capture the complexity of real-world interactions that require software commonsense, layout understanding, and fine-grained manipulation capabilities. 
To address these limitations, we introduce \ourbenchmarkname, a comprehensive benchmark comprising \numexamples finely annotated samples across diverse task types including text matching, element recognition, layout understanding, and precise manipulation. 
Additionally, we synthesize and release the largest computer use grounding dataset \ours, which contains \numdataexamples examples through multi-perspective decoupling of tasks. 
Our multi-scale models trained on \ours demonstrate its effectiveness by outperforming existing approaches on ScreenSpot-v2, ScreenSpot-Pro, and our \ourbenchmarkname. 
Furthermore, we demonstrate that improved grounding with \ours directly enhances agentic capabilities of general foundation models on complex computer tasks with state-of-the-art performance, improving from 23\% to 51\% on OSWorld. 
Through detailed ablation studies, we identify key factors contributing to grounding performance and verify that combining specialized data for different interface elements enables compositional generalization to novel interfaces. 
All benchmark, data, checkpoints, and code are open-sourced and available at \url{https://osworld-grounding.github.io}.

\end{abstract}

%% file: text/intro.tex
\section{Introduction}
Graphical user interface (GUI) grounding, the ability to accurately map natural language instructions to specific actions (including the positions of on-screen elements), is a cornerstone for computer use agents to effectively interact with GUIs on devices such as mobile phones and desktop computers. 
It plays a critical role, whether as an isolated component of human-machine interaction, a facilitator of multi-model collaboration agents, or a means to enhance end-to-end models. 

Achieving practical GUI grounding requires software commonsense (e.g., understanding the meaning of icons, the functions of components, and specific software knowledge), layout understanding (e.g., interpreting a sidebar on one side or elements under a panel) and fine-grained component manipulation (e.g., adjusting a slider or selecting text on character level). 
Knowledge and grounding together enable comprehension and interaction.
Additionally, rejecting infeasible instructions (e.g., mistaking Thunderbird for Firefox) is necessary to avoid unrecoverable states.
Previous work around GUI grounding oversimplify these tasks as short referring expressions.
Such descriptions are clear but leave a gap with real-world requirements.
As a result, existing benchmarks like ScreenSpot-v2~\citep{cheng2024seeclickharnessingguigrounding, Wu2024OSATLASAF} show saturation at early stages (\textasciitilde 90\%) accuracy by recent approaches~\citep{qin2025uitarspioneeringautomatedgui} together with the progress of vision-language models (VLMs)~\citep[see][{\emph{i.a.}}]{chen2024fargpt4vclosinggap, wang2024qwen2vlenhancingvisionlanguagemodels, bai2025qwen25vltechnicalreport}, primarily focusing on simple instructions to locate referenced elements in screenshots.
Current evaluation approaches either lack nuance in their assessment criteria or artificially inflate difficulty through unnatural conditions, 
such as ScreenSpot-Pro's extreme resolutions that rarely occur in typical computing environments.
Achieving practical grounding requires software context awareness and fine-grained manipulation capabilities for diverse GUI elements including dropdown menus, tabbed interfaces, scrollbars, and context-sensitive controls that have not been adequately measured or explored.
On the data side, the primary capabilities of current grounding models arise from structured text and screenshot correspondences found on webpages(e.g., SeeClick~\cite{cheng2024seeclickharnessingguigrounding}, UGround~\cite{gou2024navigatingdigitalworldhumans}, OmniParser~\cite{lu2024omniparser}, OS-Atlas~\cite{Wu2024OSATLASAF}, Aria-UI~\cite{yang2024aria}).
Alternatively, they rely on manually annotated data (e.g., Aguvis~\cite{xu2024aguvis}, UI-TARS~\cite{qin2025uitarspioneeringautomatedgui}).
The former can capture coarse-grained element understanding signals for webpage but lacks fine-grained operational capabilities for UI elements.
The latter, due to high manual annotation costs, struggles to scale effectively.

\begin{figure}[t]
    \centering
    \includegraphics[width=\linewidth]{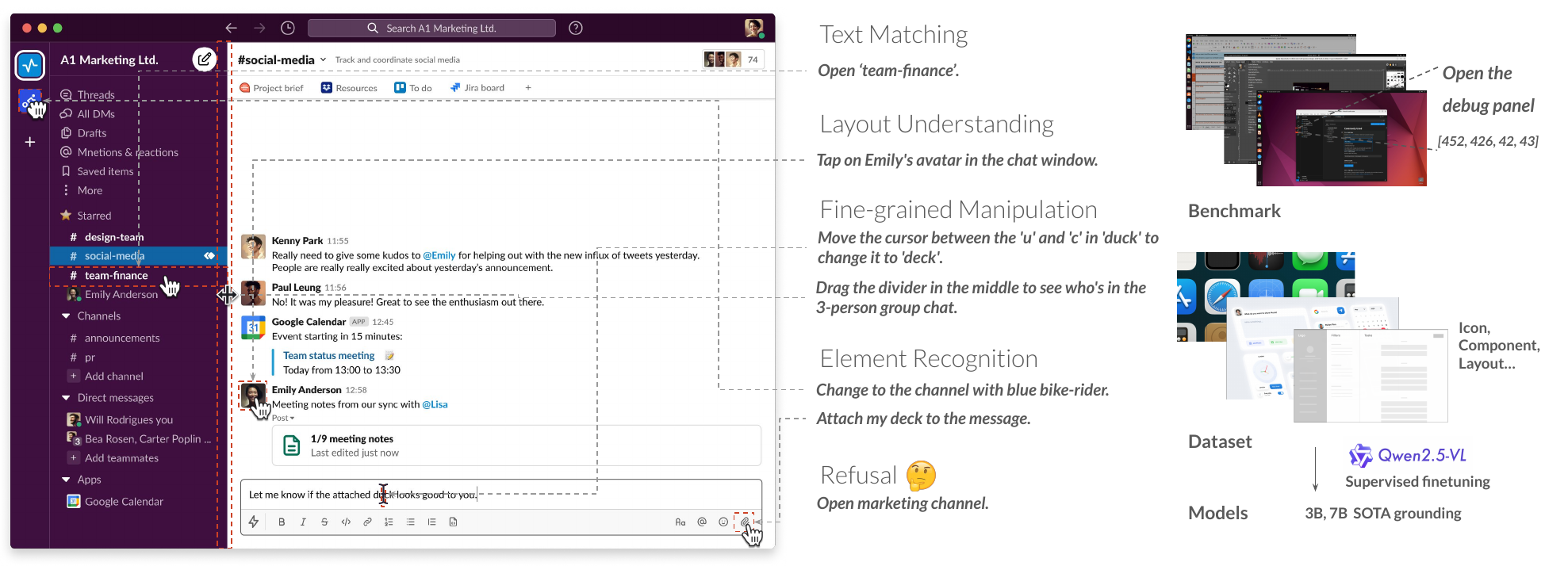}
    \caption{We have developed a comprehensive benchmark comprising \numexamples examples that cover the diverse task types that previous work has overlooked. Additionally, we synthesize and release the largest computer use grounding dataset containing \numdataexamples examples, and train models that achieve state-of-the-art performance on this dataset.}
    \label{fig:grounding_examples}
\end{figure}

To better assist the community in addressing GUI grounding challenges, we start with benchmarks and data as shown in Figure~\ref{fig:grounding_examples}. 
We develop the \ourbenchmarkname, comprising \numexamples finely annotated samples that systematically cover text matching, element recognition, layout understanding, fine-grained manipulation and infeasibility, with annotations for the element types required to solve each task. 
On the data side, we collect and synthesize the largest-scale open grounding dataset \ours in the web and desktop domain through multi-perspective decoupling of tasks. 
Additionally, we train multi-scale models on this dataset to validate its effectiveness.

Our evaluation on ScreenSpot-v2, ScreenSpot-Pro and \ourbenchmarkname demonstrates that our approach significantly outperforms existing models in aspect of grounding ability. 
Beyond standalone grounding performance, we show that improved grounding directly translates to enhanced agentic capabilities on complex tasks in OSWorld~\citep{xie2024osworldbenchmarkingmultimodalagents} and WindowsAgentArena~\citep{Bonatti2024WindowsAA} benchmarks. 
Through detailed ablation studies, we identify key factors that most significantly contribute to grounding performance, providing insights for future data collection and training efforts to enhance such abilities. 
Our case studies verify the effectiveness of our decomposition hypothesis, demonstrating that combining specialized data for different interface elements enables compositional generalization to novel interfaces.

%% file: text/approach.tex
\section{Approach}

\input{text/definition}
\input{text/benchmark}

\input{text/data_construction}

%% file: text/definition.tex
\paragraph{Task Definition} 
A \textbf{Multimodal Agent} is an AI system that visually perceives the GUI from the environment. At each step \( t \), it receives a visual observation \( O_t \) (e.g., pixel data \( \in \mathbb{R}^{H \times W \times C} \)) and executes an action \( a_t \) based on a natural language instruction \( I \) and its current observation (and potentially history). 
The agent learns a policy \( \pi: (O_t, I, \text{state}_t) \rightarrow a_t \) to generate the sequence of actions \( A = \{a_1, \dots, a_n\} \), purely from visual perception without access to the GUI's underlying code or APIs. 
An action \( a_t \) consists of an action type (e.g., \texttt{click}, \texttt{move\_to}, \texttt{type}) and action parameters that typically involve coordinates, represented as either a point \( (x, y) \) or a bounding box \( (x, y, w, h) \) to specify the target GUI element.
\textbf{GUI Grounding} represents the core capability enabling the policy \( \pi \) to function effectively at each step \( t \). 
Given a potentially step-specific interpretation or sub-instruction \( I_t \) (derived explicitly or implicitly from \( I \)) and the current observation \( O_t \), grounding is the process of mapping these inputs to the specific, executable action \( a_t \). 
Achieving accurate grounding for each \( (I_t, O_t) \) pair is a fundamental objective in training the agent and a key determinant of the policy's success on the overall task.

%% file: text/benchmark.tex
\subsection{\ourbenchmarkname}
\label{sec:osworld_g}

\subsubsection{Benchmark Construction}
We sample screenshots from the rollout of previous models on OSWorld~\cite{xie2024osworldbenchmarkingmultimodalagents}, as this is currently one of the most widely adopted benchmark environments for evaluating computer use agents, covering diverse elements, fine-grained components, and rich layouts. 
The screen size is set to 720p and 1080p.
Following ScreenSpot and ScreenSpot-Pro, we annotate these screenshots with instructions and corresponding bounding boxes. 
Even for fine-grained manipulation tasks such as text editing, we can identify specific pixel regions that are sufficient for creating appropriate bounding boxes. 
For evaluation, we determine whether the coordinates in the agent's predicted actions fall within the annotated bounding boxes, and calculate accuracy based on this spatial containment criterion.
We utilize the CVAT~\footnote{\url{https://app.cvat.ai/}} platform to collect annotations of objects corresponding to instructions. 
Each annotation is performed by individuals highly familiar with the software details and is verified through actual testing in the real software, particularly for edge cases. 
Following the initial annotations, we conduct multiple verification rounds based on feedback from preliminary experiments. For each example in \ourbenchmarkname, we assign a fine-grained tag that identifies the element types required to solve the example. Additionally, we provide a refined annotation for each example that rephrases the original instructions to decompose the necessary GUI knowledge needed to complete the task.
In total, we collect \numexamples samples, annotated with 32 different UI-types, each with a paraphrased instruction that requires no software knowledge to execute.
The average annotation time per sample is approximately 0.5 human-hours. We provide the annotation workflow in the Appendix \ref{sub:annotation_details}.

\subsubsection{Data Types}
\input{tables/bmk_classification}
Leveraging the fine-grained element type tags, we categorize tasks into capability dimensions that directly reflect core model competencies: \textit{text matching}, \textit{element recognition}, \textit{layout understanding}, \textit{fine-grained manipulation}, and \textit{refusal handling}, as presented in Table~\ref{tab:capabilities-classification}.

\paragraph{Text Matching \& Element Recognition} 
Most cases in GUI grounding require simply text matching and element recognition as two fundamental capabilities.
Text matching involves grounding actions according to explicit textual information provided in instructions (e.g., ``Select `As Attachment`''). 
This requires matching the specified text to locate the appropriate screen region. 
Element recognition encompasses multiple aspects of visual understanding: identifying visual patterns such as icons or images (e.g., ``Click on Ellipse icon''), and importantly, recognizing elements based on their implied functionality rather than explicit labels. 
For example, recognizing a ``Save'' button by its floppy disk icon, a ``Settings'' option by its gear icon, or a ``Search'' function by its magnifying glass symbol—all cases where the agent must associate visual elements with their functional purpose, even when no explicit text label is present.

\paragraph{Layout Understanding}

\begin{wrapfigure}{r}{0.5\textwidth}
    \centering
    \vspace{-15pt}
    \includegraphics[width=\linewidth]{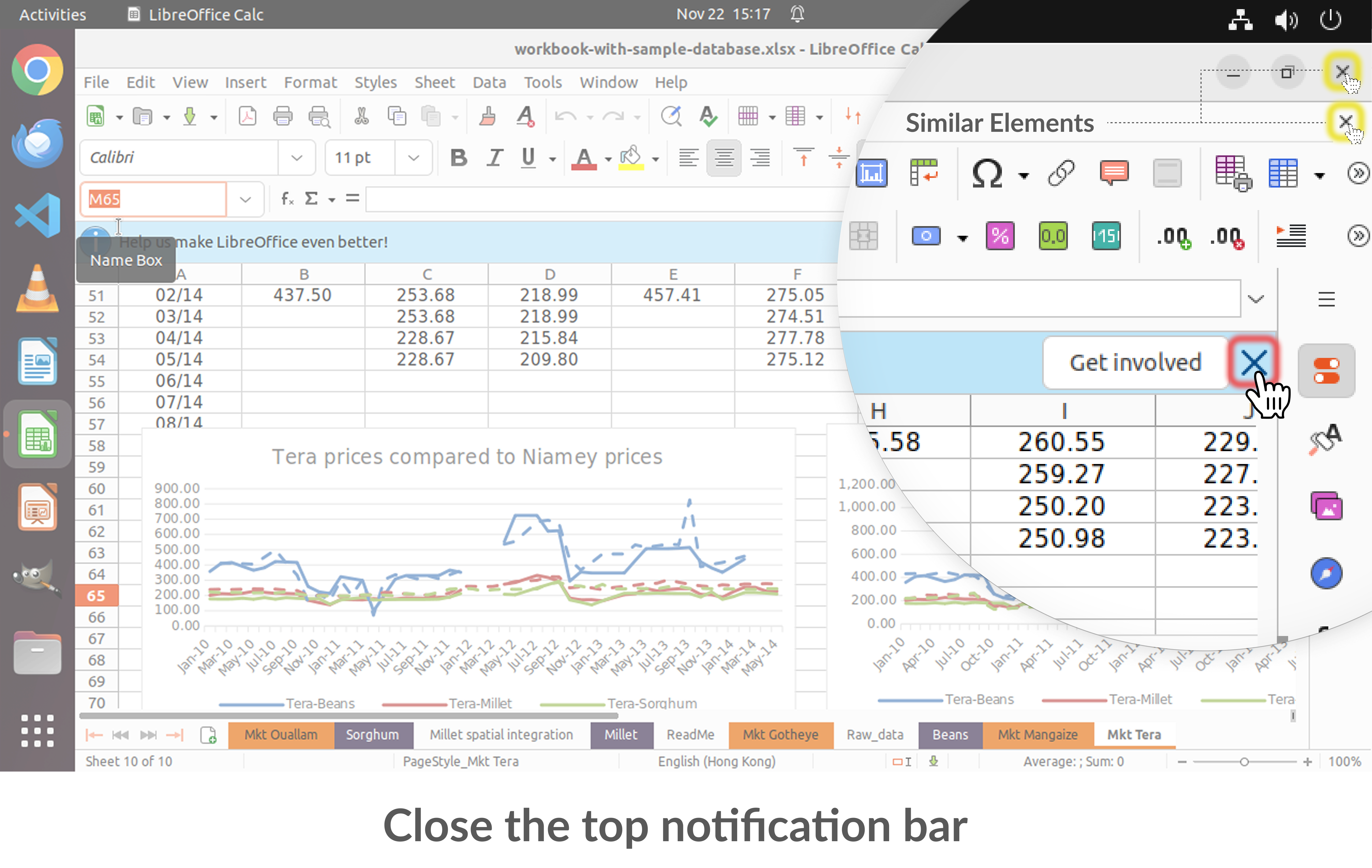}
    \caption{Example of layout understanding case in \ourbenchmarkname.}
    \vspace{-25pt}
    \label{fig:benchmark_case_layout_understanding}
\end{wrapfigure}

GUIs are typically designed with modular structures. 
Knowledge of layout hierarchy is critical to locate elements precisely, as visually similar elements may exist across different modules, and describing elements often requires referencing their position within the layout.
For instance, instructions like ``Close the top notification bar'' require correct identification of the notification bar area, as multiple similar close buttons may appear throughout the interface. 
Other cases require identification of toolbars, panels, pop-up windows, and other common GUI modules.

\paragraph{Fine-grained Manipulation}

\begin{wrapfigure}{r}{0.5\textwidth}
    \centering
    \vspace{-15pt}
    \includegraphics[width=\linewidth]{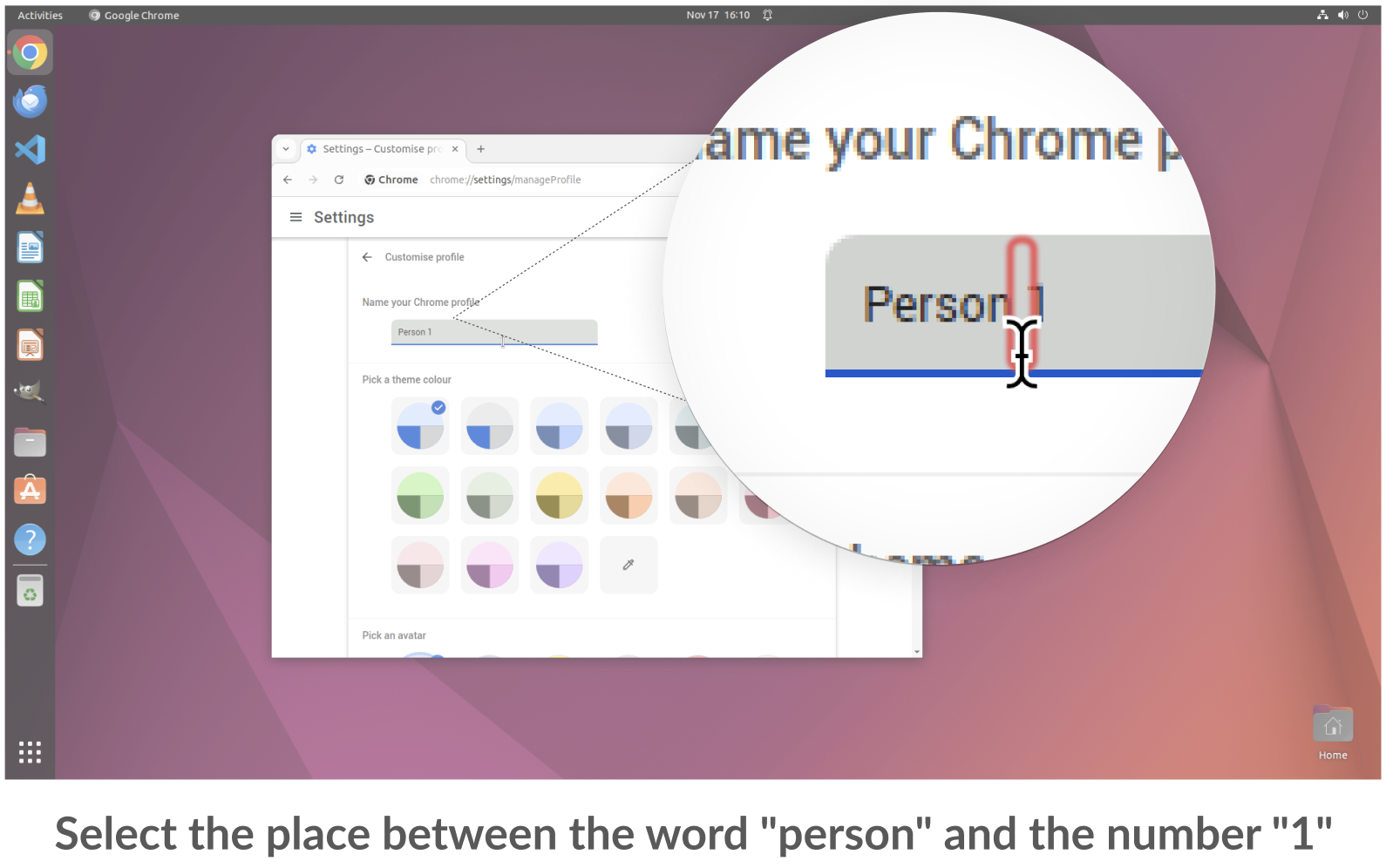}
    \caption{Example of fine-grained manipulation case in \ourbenchmarkname.}
    \vspace{-30pt}
    \label{fig:benchmark_case_finegrained_manipulation}
\end{wrapfigure}

Computer use agent tasks frequently involve text editing operations. 
Instructions such as ``Select the place between the world 'person' and the number '1'' require precise cursor placement between specific letters, which may occupy only a small portion of the screen. 
Such actions demand the ability to perform operations with high precision within relatively small screen regions. 
Beyond text, this capability extends to interaction with compact components like sliders, steppers, table cell and other small elements.

\vspace{15pt}

\paragraph{Infeasible}
Certain tasks may arise from hallucinated or incorrect low-level user instructions or automated planning suggestions. 
An example could be an instruction like, ``Click to open the Firefox browser,'' when the shown screenshot does not contain a Firefox icon or any visible reference to it.
A distinct subset of \ourbenchmarkname tasks with \numinfeasibleexamples examples explicitly highlights these infeasible scenarios. 
These tasks are valuable for evaluating a system’s ability to reject impossible instructions gracefully, preventing errors and ensuring safer, more robust interactions.

%% file: tables/bmk_classification.tex
\begin{wraptable}{r}{0.6\textwidth}
\centering
\vspace{-15pt}
\caption{Distribution of examples in the \ourbenchmarkname benchmark categorized by GUI grounding capabilities and their corresponding interface element types. Full table can be refer to Appendix \ref{app:og_data_types}}
\scalebox{0.8}{
\footnotesize
\begin{tabular}{p{3.5cm}p{4cm}c}
\toprule
\textbf{Capabilities} & \textbf{Element Types} & \textbf{\# of Examples} \\
\midrule
Text Matching & Label & 268 \\
\addlinespace
Element Recognition & Icon, Image, Button & 337 \\
\addlinespace
Layout Understanding & Tab, Menu Bar, Dropdown Menu, Panel/Container, … & 252 \\
\addlinespace
Fine-grained Manipulation & Slider, Stepper, Text Field, Input Box, Divider, Table, … & 154 \\
\addlinespace
Refusal & -- & 54 \\
\bottomrule
\end{tabular}
}
\vspace{-15pt}
\label{tab:capabilities-classification}
\end{wraptable}

%% file: text/data_construction.tex
\subsection{\ours Data Construction}\label{sec:data-construction}
\begin{figure}[h]
    \centering
    \includegraphics[width=\linewidth]{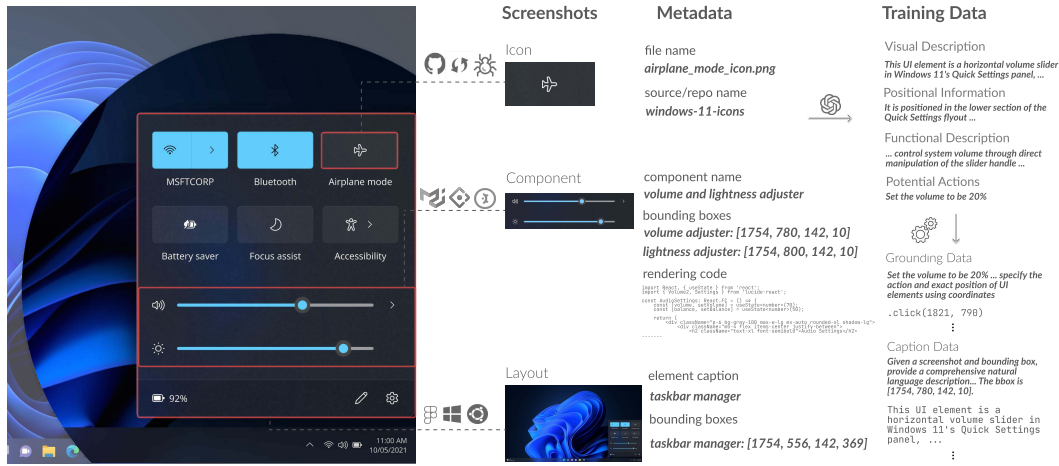}
    \caption{An overview of the synthetic data generation pipeline, demonstrating how screenshots and metadata are collected and synthesized, and subsequently converted into training data.}
    \label{fig:data-construction}
\end{figure}

To enable robust GUI grounding, we construct the world’s largest multimodal dataset tailored for computer-use grounding scenarios, containing \numdataexamples newly synthesized examples.
Our grounding data collection process centers on gathering pairs of \textit{screenshots} and \textit{metadata} (information such as filename, rendering code, element bounding box, \textit{etc}.), which are then further transformed into \textit{training data} which contains queries and corresponding answers for VLMs to learn from it. 
Previous methods in Figure~\ref{fig:data-construction} provides an overview of this pipeline.

\subsubsection{Icon}
\textit{Icons} are essential visual elements in graphical user interfaces that convey functionality through compact, recognizable imagery.
To create a comprehensive collection of icons and corresponding metadata for grounding, we employ three complementary data collection strategies.

\paragraph{GitHub Repositories and Specialized Icon Websites}

Many open-source software projects archive their design icons within GitHub repositories.
To acquire a varied collection, we systematically mine repositories containing the key term such as ``icon'' applying filtering criteria including star count, quantity of icon images, and temporal relevance.
This yield icons representing various design paradigms such as flat design, fluent design, and skeuomorphism.
To supplement our collection with production website icons, we implement a targeted web crawling pipeline that identifies and extracts icon elements from popular websites across various categories, capturing both visual assets and associated metadata including class names, aria labels, and contextual information.
We leverage these icons by generating detailed descriptions through LLMs and creating training scenarios where models identify target icons based on textual descriptions.
This comprehensive approach provides access to contemporary icons in their natural context, allowing us to capture emerging design patterns not yet available in open-source repositories.

\paragraph{Reverse Engineering Software}
To address the gap in desktop software icons, we employ reverse engineering techniques using specialized tools like IconsExtract to extract icons directly from executable files, DLLs, and system libraries across Windows, macOS and Ubuntu. 
We target a diverse range of software including Windows system applications and commonly used desktop applications. 
This method provide access to thousands of production-quality icons representing real-world software functionality.

\vspace{-5pt}
\subsubsection{Component}
\vspace{-5pt}
A \textit{component} refers to a functional unit composed of icons, UI elements, and text, collectively enabling specific modes of computer-based interaction. 
Components serve as fundamental interaction units essential for user engagement in digital environments.

\paragraph{Synthesis Process}
We collect screenshots and associated metadata primarily through a code-and-rendering pipeline. 
By leveraging mainstream production-level UI component libraries commonly used in front-end development (e.g., Material UI), we select components and use their example code as the base code. 
We then employ LLMs to synthesize functional cases for specific tasks (such as a slider for air conditioning control) using the base code as context. 
We render these within a React application to obtain visual screenshots and extract corresponding metadata, such as the element position tree, built-in component names, and coordinates. This approach allows us to generate diverse component examples with precise ground truth source code.

\paragraph{Real-world Augmentation}
We observe that common interactive behaviors such as scrolling a webpage, clicking a cell in spreadsheet or resizing a text box in slides are underrepresented in code-based libraries. To address this gap, we further source real-world screenshots from existing websites and applications. 
For these, we utilize HTML parsing and application-specific tools (e.g., \texttt{python-pptx}) to extract structured metadata. 
All the implementation details can be found in the Appendix \ref{sub:component_data_sources_and_pipelines}.

\vspace{-5pt}
\subsubsection{Layout}
\vspace{-5pt}
A \textit{layout} refers to the spatial arrangement of UI elements and components within an application or across the entire operating system. 
Layout understanding is crucial for tasks that require reasoning about the overall structure of the screen, enabling agents to interpret and interact with complex, multi-element interfaces at the application or system level.

\paragraph{Prototype Designs}
UI prototype platforms such as Figma~\footnote{\url{https://www.figma.com/}} provide numerous website and application design templates, including authentic specifications for production applications like VSCode, Zoom, and Microsoft 365.
These designs offer valuable ground truth information, as each element includes designer-specified bounding boxes, component types, and functional descriptions.
By using the official APIs of these platforms, we exported the designs as high-quality images while preserving their structured metadata, including hierarchical relationships between elements and positional data.

\paragraph{Real-World Application Screenshots}
To further improve scalability and diversity, we supplement our dataset with raw screenshots captured from real-world applications running on operating systems. 
We collect these screenshots by leveraging agent rollout data from OSWorld and WindowsAgentArena. 
Subsequently, we utilize the object detection model from OmniParser-v2 to generate bounding boxes for interface elements, thereby obtaining the necessary metadata.

\vspace{-5pt}
\subsubsection{Data Processing}\label{sec:data-processing}
\vspace{-5pt}
After obtaining screenshots (icons, components, layouts) and metadata (filenames, paths, rendered source code, UI designer annotations, \textit{etc}.), we convert them into an image-text-to-text multimodal question-answering format, creating richer and more natural language-oriented data suitable for VLM training.
Our processing approach remain consistent across the different data types. 
We employ a VisualSketchpad~\cite{hu2024visual}-like prompting methodology with models such as GPT-4o and Claude to generate enriched annotations based on the original screenshots and metadata, describing both appearance and functionality.
We construct two complementary training formats: 
(1) \textit{grounding format}, where the model receives a screenshot with instructions and must predict actions or relevant bounding boxes; 
and (2) \textit{description format}, where the model receives a screenshot with bounding boxes and must provide descriptive information.
For screenshots with multiple potential query-answer pairs, we compress them into single conversation to improve training efficiency.

\vspace{-5pt}
\subsubsection{Supplementary Training Data}
\vspace{-5pt}
To enhance the model's ability to identify and reject infeasible actions, we construct a refusal part in out dataset by mismatching existing instructions with unrelated screenshots, yield over 2.6 million examples. 
We further sample and manually inspect a subset of these examples to verify that the vast majority indeed reflects truly infeasible actions. 
In addition, we integrate and unify new datasets from previous work (human-labeled or synthesized) such as SeeClick, OS-Atlas, follow the practice from Aguvis~\cite{xu2024aguvis}.
We observe that synthetic data obtained directly from the Internet such as SeeClick, OS-Atlas contain noisy examples, we use UI-TARS-72B to filter them and keep the labeled and predicted matching part of the data.
Full data statistics in Table~\ref{tab:data_stats}.

%% file: text/experiments.tex
\vspace{-5pt}
\section{Experiments}
\label{sec:experiments}
\vspace{-5pt}

We first adapt previous benchmarks for testing our data effectiveness.
We adapt different sizes of the latest Qwen2.5-VL~\cite{bai2025qwen25vltechnicalreport} as our backbone model, set the maximum pixel limit to approximately 1080p.
Model finetuning takes approximately 20 hours for the 3B model, and 30 hours for the 7B model, conducted using cluster of 128 CPU cores, 512GB memory, and 64 NVIDIA H100 GPUs.

\vspace{-7pt}
\subsection{Grounding Ability}
\vspace{+2pt}

\input{tables/screenspot_v2}
\input{tables/screenspot_pro}

We select several benchmarks for GUI grounding. 
The most commonly used benchmarks in the past include ScreenSpot-v2 (Table~\ref{tab:screenspot_v2_comparison}), ScreenSpot-Pro (Table~\ref{tab:screenspot_pro_comparison}), which focuses on high-resolution and professional software charts, UI-Vision~\cite{nayak2025uivisiondesktopcentricguibenchmark} (Table~\ref{tab:uivision_comparison}), which focuses on fine-grained evaluation of computer use agents in real-world desktop environments, and \ourbenchmarkname (Table~\ref{tab:osworld_g_comparison}), which we use to evaluate model performance on fine-grained and functional components. 

The results show that fine-tuning existing open-source models on our data achieves state-of-the-art performance, surpassing other dedicated computer use model such as Operator (unpublished data and model) and UI-TARS (unpublished data) with a small model size.
On \ourbenchmarkname, we observe that models generally achieve the highest accuracy on examples involving text matching, outperforming their abilities in element recognition and layout understanding, with the lowest performance observed in fine-grained manipulation tasks. 
Notably, although we included refusal data during training to encourage the model to reject instructions referring to elements not present on the screen, the model rarely produces refusal responses. 
Similarly, in all models except Gemini-2.5-Pro, especially those specifically trained for computer-use tasks, refusal predictions are consistently absent. 
\input{tables/uivision}
\input{tables/osworld_g}

\vspace{-5pt}
\subsection{Agentic Ability}
\vspace{-5pt}
We hope that the data and benchmark we provide will ultimately serve as a critical signal in fostering the agentic capabilities required, rather than merely enhancing specific grounding abilities. 
We evaluate our approach on the computer use benchmarks in online environments, namely OSWorld~\cite{xie2024osworldbenchmarkingmultimodalagents,osworld_verified} and WindowsAgentArena~\citep{Bonatti2024WindowsAA}.
We employ foundation models like GPT-4o or o3 as the planner model, which receives high-level instructions and, at each step, predicts the next low-level natural language instruction based on the current observation and action history. 
Our \ours model then takes these low-level instructions and predicts the concrete actions to execute. 
To control for confounding variables, we do not introduce any specialized agent architecture or model scheduling~\cite{agashe2024agentsopenagentic}.

\input{tables/osworld}

The results demonstrate that, when using our model as the grounding component, a simple agent with foundation models that are not specialized in computer use tasks can achieve state-of-the-art performance, surpassing previous approaches that used 72B-scale models for grounding, and matching the performance of specialized models. 
Additionally, our agent system exhibits a similar trend to Operator, with performance improving as deployment scale increases. 
These findings suggest that, given the current reasoning capabilities of large language models, supplementing them with enhanced grounding ability—either through additional data or external systems—can be a starting point to build highly effective agentic systems.

%% file: tables/screenspot_v2.tex
\begin{table}[htbp]
    \vspace{-10pt}
    \caption{Comparison of various planners and grounding methods on ScreenSpot-v2. The highlighted column presents the overall average performance across all categories}
    \centering
     \resizebox{0.85\textwidth}{!}{%
    \begin{tabular}{llcccccc >{\columncolor{heliotrope}}c}
        \toprule
        \multirow{2}{*}{\textbf{Planner}} & \multirow{2}{*}{\textbf{Grounder}} & \multicolumn{2}{c}{\textbf{Mobile}} & \multicolumn{2}{c}{\textbf{Desktop}} & \multicolumn{2}{c}{\textbf{Web}} & \multirow{2}{*}{\cellcolor{white}\textbf{Avg}} \\
        \cmidrule(lr){3-4} \cmidrule(lr){5-6} \cmidrule(lr){7-8} 
        & & \textbf{Text} & \textbf{Icon/Widget} & \textbf{Text} & \textbf{Icon/Widget} & \textbf{Text} & \textbf{Icon/Widget} & \cellcolor{white}{} \\
        \midrule
        \multirow{6}{*}{-}
        & SeeClick & 78.4 & 50.7 & 70.1 & 29.3 & 55.2 & 32.5 & 55.1 \\
        & OS-Atlas-Base-7B & 95.2 & 75.8 & 90.7 & 63.6 & 90.6 & 77.3 & 85.1 \\
        & UI-TARS-7B & 96.9 & 89.1 & 95.4 & 85.0 & 93.6 & 85.2 & 91.6 \\
        & UI-TARS-72B & 94.8 & 86.3 & 91.2 & 87.9 & 91.5 & 87.7 & 90.3 \\
        & Operator & 47.3 & 41.5  & 90.2 & 80.3 & 92.8 & 84.3 & 70.5 \\
        & Qwen2.5-VL-3B & 93.4 & 73.5 & 88.1 & 58.6 & 88.0 & 71.4 & 80.9 \\
        & Qwen2.5-VL-7B & 97.6 & 87.2 & 90.2 & 74.2 & 93.2 & 81.3 & 88.8 \\
        & Qwen2.5-VL-32B & 97.9 & 88.2 & 98.5 & 79.3 & 91.2 & 86.2 & 91.3 \\
        \midrule
        \multirow{2}{*}{GPT-4o}
        & OS-Atlas-Base-7B & 96.2 & 83.4 & 89.7 & 69.3 & 94.0 & 79.8 & 87.1 \\
        & OmniParser-v2 & 95.5 & 74.6 & 92.3 & 60.9 & 88.0 & 59.6 & 80.7 \\
        \midrule
        \multicolumn{2}{c}{\ourmodelthreebillion} & 96.6 & 81.5 & 96.9 & 78.6 & 88.5 & 83.7 & 88.6 \\
        \multicolumn{2}{c}{\ourmodelsevenbillion} & 96.9 & 87.2 & 95.9 & 87.9 & 94.4 & 84.2 & 91.7 \\
        \bottomrule
    \end{tabular}
    }
    \label{tab:screenspot_v2_comparison}
\end{table}
\vspace{-5pt}

%% file: tables/screenspot_pro.tex
\begin{table}[htbp] 
\vspace{-5pt}
    \centering
    \caption{Comparison of models on ScreenSpot-Pro. The highlighted column presents the overall average performance across all categories.} 
    \label{tab:agent_comparison} 
    \resizebox{\textwidth}{!}{
    \begin{tabular}{@{}l rrr rrr rrr rrr rrr rrr rr>{\columncolor{heliotrope}}r @{}}
        \toprule
        \textbf{Agent Model} & \multicolumn{3}{c}{\textbf{Development}} & \multicolumn{3}{c}{\textbf{Creative}} & \multicolumn{3}{c}{\textbf{CAD}} & \multicolumn{3}{c}{\textbf{Scientific}} & \multicolumn{3}{c}{\textbf{Office}} & \multicolumn{3}{c}{\textbf{OS}} & \multicolumn{3}{c}{\textbf{Avg}} \\
        \cmidrule(lr){2-4} \cmidrule(lr){5-7} \cmidrule(lr){8-10} \cmidrule(lr){11-13} \cmidrule(lr){14-16} \cmidrule(lr){17-19} \cmidrule(lr){20-22}
        & Text & Icon & Avg & Text & Icon & Avg & Text & Icon & Avg & Text & Icon & Avg & Text & Icon & Avg & Text & Icon & Avg & Text & Icon & \cellcolor{white}{Avg} \\
        \midrule
        SeeClick~\cite{cheng2024seeclickharnessingguigrounding} & 0.6 & 0.0 & 0.3 & 1.0 & 0.0 & 0.6 & 2.5 & 0.0 & 1.9 & 3.5 & 0.0 & 2.0 & 1.1 & 0.0 & 0.9 & 2.8 & 0.0 & 1.5 & 1.8 & 0.0 & 1.1 \\
        Qwen2-VL-7B~\cite{wang2024qwen2vlenhancingvisionlanguagemodels} & 2.6 & 0.0 & 1.3 & 1.5 & 0.0 & 0.9 & 0.5 & 0.0 & 0.4 & 6.3 & 0.0 & 3.5 & 3.4 & 1.9 & 3.0 & 0.9 & 0.0 & 0.5 & 2.5 & 0.2 & 1.6 \\
        ShowUI-2B~\cite{lin2024showui} & 16.9 & 1.4 & 9.4 & 9.1 & 0.0 & 5.3 & 2.5 & 0.0 & 1.9 & 13.2 & 7.3 & 10.6 & 15.3 & 7.5 & 13.5 & 10.3 & 2.2 & 6.6 & 10.8 & 2.6 & 7.7 \\
        CogAgent-18B~\cite{hong2023cogagent} & 14.9 & 0.7 & 8.0 & 9.6 & 0.0 & 5.6 & 7.1 & 3.1 & 6.1 & 22.2 & 1.8 & 13.4 & 13.0 & 0.0 & 10.0 & 5.6 & 0.0 & 3.1 & 12.0 & 0.8 & 7.7 \\
        Aria-UI~\cite{yang2024aria} & 16.2 & 0.0 & 8.4 & 23.7 & 2.1 & 14.7 & 7.6 & 1.6 & 6.1 & 27.1 & 6.4 & 18.1 & 20.3 & 1.9 & 16.1 & 4.7 & 0.0 & 2.6 & 17.1 & 2.0 & 11.3 \\
        Claude~\cite{AnthropicModelCA} & 22.0 & 3.9 & 12.6 & 25.9 & 3.4 & 16.8 & 14.5 & 3.7 & 11.9 & 33.9 & 15.8 & 25.8 & 30.1 & 16.3 & 26.9 & 11.0 & 4.5 & 8.1 & 23.4 & 7.1 & 17.1 \\
        Operator~\cite{cua2025} & 50.0 & 19.3 & 35.1 & 51.5 & 23.1 & 39.6 & 16.8 & 14.1 & 16.1 & 58.3 & 24.5 & 43.7 & 60.5 & 28.3 & 53.0 & 34.6 & 30.3 & 32.7 & 45.0 & 23.0 & 36.6 \\
        OS-Atlas-7B~\cite{Wu2024OSATLASAF}  & 33.1 & 1.4 & 17.7 & 28.8 & 2.8 & 17.9 & 12.2 & 4.7 & 10.3 & 37.5 & 7.3 & 24.4 & 33.9 & 5.7 & 27.4 & 27.1 & 4.5 & 16.8 & 28.1 & 4.0 & 18.9 \\
        UGround-V1-7B~\cite{gou2024navigatingdigitalworldhumans} & -   & -   & 35.5 & -   & -   & 27.8 & -   & -   & 13.5 & -   & -   & 38.8 & -   & -   & 48.8 & -   & -   & 26.1 & -   & -   & 31.1 \\
        UI-TARS-2B~\cite{qin2025uitarspioneeringautomatedgui} & 47.4 & 4.1 & 26.4 & 42.9 & 6.3 & 27.6 & 17.8 & 4.7 & 14.6 & 56.9 & 17.3 & 39.8 & 50.3 & 17.0 & 42.6 & 21.5 & 5.6 & 14.3 & 39.6 & 8.4 & 27.7 \\
        UI-TARS-7B~\cite{qin2025uitarspioneeringautomatedgui} & 58.4 & 12.4& 36.1 & 50.0 & 9.1 & 32.8 & 20.8 & 9.4 & 18.0 & 63.9 & 31.8 & 50.0 & 63.3 & 20.8 & 53.5 & 30.8 & 16.9 & 24.5 & 47.8 & 16.2 & 35.7 \\
        UI-TARS-72B~\cite{qin2025uitarspioneeringautomatedgui} & 63.0 & 17.3 & 40.8 & 57.1 & 15.4 & 39.6 & 18.8 & 12.5 & 17.2 & 64.6 & 20.9 & 45.7 & 63.3 & 26.4 & 54.8 & 42.1 & 15.7 & 30.1 & 50.9 & 17.5 & 38.1 \\
        Qwen2.5-VL-3B & 38.3 & 3.4 & 21.4 & 40.9 & 4.9 & 25.8 & 22.3 & 6.3 & 18.4 & 44.4 & 10.0 & 29.5 & 48.0 & 17.0 & 40.9 & 33.6 & 4.5 & 20.4 & 37.8 & 6.6 & 25.9 \\
        Qwen2.5-VL-7B & 51.9 & 4.8 & 29.1 & 36.9 & 8.4 & 24.9 & 17.8 & 1.6 & 13.8 & 48.6 & 8.2 & 31.1 & 53.7 & 18.9 & 45.7 & 34.6 & 7.9 & 22.4 & 39.9 & 7.6 & 27.6 \\
        Qwen2.5-VL-32B & 74.0 & 21.4 & 48.5 & 61.1 & 13.3 & 41.1 & 38.1 & 15.6 & 32.6 & 78.5 & 29.1 & 57.1 & 76.3 & 37.7 & 67.4 & 55.1 & 27.0 & 42.3 & 63.2 & 22.5 & 47.6 \\
        \midrule
        \ourmodelthreebillion & 61.0 & 13.8 & 38.1 & 53.5 & 8.4 & 34.6 & 27.4 & 9.4 & 23.0 & 54.2 & 18.2 & 38.6 & 64.4 & 32.1 & 57.0 & 38.3 & 9.0 & 25.0 & 49.8 & 13.7 & 36.1 \\
        \ourmodelsevenbillion & 42.9 & 11.0 & 27.4 & 50.0 & 11.9 & 34.0 & 38.0 & 14.1 & 32.2 & 72.9 & 25.5 & 52.4 & 75.1 & 47.2 & 68.7 & 33.6 & 16.9 & 26.0 & 52.6 & 18.2 & 39.5 \\
        \bottomrule
    \end{tabular}%
    } 
    \label{tab:screenspot_pro_comparison}
\end{table}
\vspace{-5pt}

%% file: tables/uivision.tex
\begin{table}[htbp]
\vspace{-10pt}
    \caption{Comparison of models on element grounding tasks in UI-Vision. The highlighted column presents the overall average performance across all categories.}
    \centering
    \resizebox{0.75\textwidth}{!}{  
    \begin{tabular}{lccc >{\columncolor{heliotrope}}c}  
        \toprule
        \textbf{Model} & \textbf{Basic Overall} & \textbf{Functional Overall} & \textbf{Spatial Overall} & \cellcolor{white}\textbf{Final Avg} \\
        \midrule
        Claude-3.7-Sonnet~\cite{TheC3} & 9.48 & 7.73 & 7.60 & 8.27 \\
        Qwen-2.5VL-7B & 1.24 & 0.79 & 0.51 & 0.85 \\
        MiniCPM-V-8B~\cite{yao2024minicpmvgpt4vlevelmllm} & 7.11 & 5.30 & 1.45 & 4.34 \\
        ShowUI-2B & 8.07 & 7.67 & 2.07 & 5.94 \\
        Aria-UI & 12.2 & 14.0 & 3.98 & 10.1 \\
        UGround-v1-7B & 15.4 & 17.1 & 6.25 & 12.9 \\
        OSAtlas-7B & 12.2 & 11.2 & 3.67 & 9.02 \\
        Aguvis-7B & 17.8 & 18.3 & 5.06 & 13.7 \\
        UI-TARS-7B & 20.1 & 24.3 & 8.37 & 17.6 \\
        SeeClick & 9.42 & 4.68 & 2.07 & 5.39 \\
        UI-TARS-72B & 31.4 & 30.5 & 14.7 & 25.5 \\
        \midrule
        {\ourmodelthreebillion} & 22.3 & 25.2 & 9.35 & 18.7 \\
        {\ourmodelsevenbillion} & 32.3 & 30.5 & 12.8 & 24.8 \\
        \bottomrule
    \end{tabular}
    }
    \label{tab:uivision_comparison}
\vspace{-10pt}
\end{table}

%% file: tables/osworld_g.tex
\begin{table}[htbp]
\vspace{-15pt}
    \caption{Performance comparison of models on \ourbenchmarkname across multiple capability dimensions. The highlighted column presents the overall average performance across all categories.}
    \centering
     \resizebox{0.95\textwidth}{!}{%
    \begin{tabular}{lccccc >{\columncolor{heliotrope}}c}
        \toprule
        \textbf{Agent Model} & \textbf{Text Matching} & \textbf{Element Recognition} & \textbf{Layout Understanding} & \textbf{Fine-grained Manipulation} & \textbf{Refusal} & \cellcolor{white}\textbf{Overall} \\
        \midrule
        OS-Atlas-7B & 44.1 & 29.4 & 35.2 & 16.8 & 7.4 & 27.7 \\
        UGround-V1-7B & 51.3 & 40.3 & 43.5 & 24.8 & 0.0 & 36.4 \\
        Aguvis-7B & 55.9 & 41.2 & 43.9 & 28.2 & 0.0 & 38.7 \\
        UI-TARS-7B & 60.2 & 51.8 & 54.9 & 35.6 & 0.0 & 47.5 \\
        Seed1.5-VL \cite{seed2025seed1_5vl} & 73.9 & 66.7 & 69.6 & 47.0 & 18.5 & 62.9 \\
        UI-TARS-72B & 69.4 & 60.6 & 62.9 & 45.6 & 0.0 & 57.1 \\
        Gemini-2.5-Pro & 59.8 & 45.5 & 49.0 & 33.6 & 38.9 & 45.2 \\
        Operator & 51.3 & 42.4 & 46.6 & 31.5 & 0.0 & 40.6 \\
        Qwen2.5-VL-3B & 41.4 & 28.8 & 34.8 & 13.4 & 0.0 & 27.3 \\
        Qwen2.5-VL-7B & 45.6 & 32.7 & 41.9 & 18.1 & 0.0 & 31.4 \\
        Qwen2.5-VL-32B & 63.2 & 47.3 & 49.0 & 36.9 & 0.0 & 46.5 \\
        \midrule
        \ourmodelthreebillion & 67.4 & 53.0 & 53.8 & 44.3 & 7.4 & 50.9 \\
        \ourmodelsevenbillion & 65.9 & 55.5 & 57.7 & 46.9 & 7.4 & 54.1 \\
        \bottomrule
    \end{tabular}
    }
    \label{tab:osworld_g_comparison}
\vspace{-10pt}
\end{table}

%% file: tables/osworld.tex

\begin{wraptable}{r}{0.5\textwidth}
\small
\vspace{-15pt}
\caption{Success rate on the OSWorld and WindowsAgentArena benchmarks. 
\ours with GPT-4o results are the average success rate of 4 runs with standard deviation. 
More detailed performance see~\ref{appendix:detail_bmk_result}.}
\resizebox{0.5\textwidth}{!}{%
    \begin{tabular}{llcc}
    \toprule
    {\textbf{Planner}} & {\textbf{Grounding}} & {\textbf{OS SR}} & {\textbf{WAA SR}} \\
    \midrule
    \multicolumn{2}{c}{GPT-4o} & 5.0 & 9.4 \\
    \multicolumn{2}{c}{Kimi-VL \cite{kimiteam2025kimivltechnicalreport} } & 8.2 & 10.4 \\
    \multicolumn{2}{c}{UI-TARS-72B} & 22.7 & - \\
    \multicolumn{2}{c}{o3} & 23.0 & - \\
    \multicolumn{2}{c}{Operator \cite{cua2025}} & 32.6 & - \\
    \multicolumn{2}{c}{OpenCUA-32B \cite{wang2025opencuaopenfoundationscomputeruse}} & 34.8 & - \\
    \multicolumn{2}{c}{Claude 4 Sonnet} & 43.9 & - \\
    \midrule
    GPT-4o & Aguvis-72B & 17.0 & - \\
    GPT-4o & \ourmodelthreebillion & 24.0 \textsubscript{\scalebox{0.75}{$\pm$ 1.05}} & 33.03 \textsubscript{\scalebox{0.75}{$\pm$ 1.64}}\\
    GPT-4o & \ourmodelsevenbillion & \textbf{27.0} \textsubscript{\scalebox{0.75}{$\pm$ 1.81}} & \textbf{33.7} \textsubscript{\scalebox{0.75}{$\pm$ 0.82}}\\
    o3 & \ourmodelsevenbillion & \textbf{51.0} & - \\
    \bottomrule
\end{tabular}
}
\label{tab:os_world_waa_results}
\vspace{-40pt}
\end{wraptable}

%% file: text/analysis.tex
\vspace{15pt}
\section{Analysis}
\vspace{-5pt}

\vspace{-5pt}
\subsection{Effectiveness of Knowledge}
\vspace{-5pt}
\begin{wrapfigure}{r}{0.48\textwidth}
\centering
\vspace{-10pt}
\begin{tikzpicture}
    \begin{axis}[
        ybar,
        bar width=8pt,
        width=0.48\textwidth,
        height=5cm,
        enlarge x limits=0.12,
        legend style={at={(0.5,1.08)}, anchor=south, legend columns=-1, font=\footnotesize},
        ylabel={Success Rate (\%)},
        ylabel near ticks,
        symbolic x coords={
            UI-TARS-72B, Operator, Gemini-2.5-pro, Qwen2.5-VL-32B, Jedi-3B, Jedi-7B
        },
        xtick=data,
        xticklabels={
            UI-TARS-72B, Operator, Gemini-2.5-pro, Qwen2.5-VL-32B, {\textsc{Jedi-3B}}, {\textsc{Jedi-7B}}
        },
        xticklabel style={rotate=30, anchor=east, font=\footnotesize},
        ymajorgrids=true,
        grid style=dashed,
        nodes near coords,
        nodes near coords align={vertical},
        nodes near coords style={font=\tiny},
    ]
    \addplot+[
    fill=blue!30,
    draw=blue!30,
    every node near coord/.append style={text=black!100}
    ] coordinates {
        (UI-TARS-72B,57.1)
        (Operator,51.8)
        (Gemini-2.5-pro,45.2)
        (Qwen2.5-VL-32B,46.5)
        (Jedi-3B,50.9)
        (Jedi-7B,54.1)
    };
    \addplot+[
    fill=blue!10,
    draw=blue!10,
    every node near coord/.append style={text=black!100}
    ] coordinates {
        (UI-TARS-72B,63.7)
        (Operator,57.5)
        (Gemini-2.5-pro,47.0)
        (Qwen2.5-VL-32B,59.0)
        (Jedi-3B,61.0)
        (Jedi-7B,63.8)
    };
    \legend{Original, Refined}
    \end{axis}
\end{tikzpicture}
\vspace{-15pt}
\caption{Success rates of various models on the \ourbenchmarkname benchmark with original and refined instructions.}
\vspace{-15pt}
\label{fig:osworldg-refined}
\end{wrapfigure}
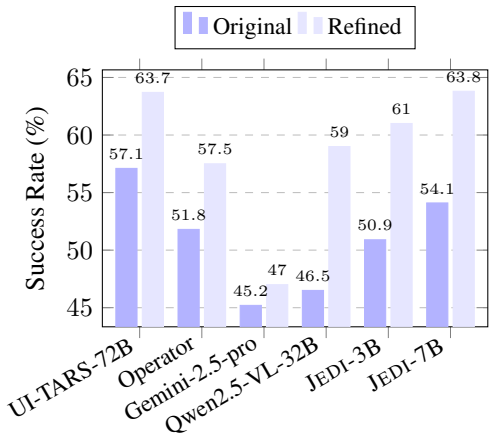

GUI grounding also requires knowledge and even reasoning. 
We aim to investigate the performance of pure grounding when almost no additional knowledge is required. 
To this end, we assume that the instruction recipient possesses minimal prior experience with GUI interactions, and we re-annotate the entire benchmark to minimize the background knowledge needed to understand each instruction. 
This is achieved by relying on easily identifiable universal features such as color and shape.
For example, the instruction ``Open the filter function for search settings.'' is refined, based on the screenshot, to ``Click the button that includes an icon of a funnel on the right of the `search settings' bar.''
We conduct experiments on several models and present the performance comparison before and after instruction refinement in Figure~\ref{fig:osworldg-refined}. 
First, we observe that model performance generally improves after instruction refinement. 
This suggests that if we can supplement models with relevant interaction experience or provide more precise expressions—either manually or via upstream models—grounding performance can be enhanced.
Second, after instruction refinement, our model achieves performance comparable to the largest state-of-the-art model, UI-TARS-72B. 
This indicates that, with appropriate data such as our \ours dataset, smaller models are already sufficient in terms of pure grounding ability, and further advantages may lie in the supplementation of background knowledge. 

\vspace{-5pt}
\subsection{Performance as Data Scaling}
\vspace{-5pt}
\begin{figure}[ht]
    \centering
    \vspace{-10pt}
    \begin{minipage}[b]{0.48\textwidth}
    \begin{tikzpicture}
    \begin{axis}[
        width=\textwidth, height=4cm,
        title={ScreenSpot-v2 Performance},
        title style={font=\footnotesize},
        xlabel={Training Data (\%)},
        xlabel style={yshift=1ex, font=\footnotesize},
        ylabel={Success Rate (\%)},
        ylabel near ticks,
        ylabel style={font=\footnotesize},
        grid=both,
        xmin=1, xmax=4,
        ymin=65, ymax=90,
        xtick={1, 2, 3, 4},
        xticklabels={10\%, 20\%, 50\%, 100\%},
        xticklabel style={font=\footnotesize},
        yticklabel style={font=\footnotesize},
        ymajorgrids=true,
        grid style=dashed,
    ]
    
    \addplot[
        color=blue,
        mark=*,
        mark size=1.5pt,thick
        ]
        coordinates {
        (1,75.94)(2,76.81)(3,78.77)(4,79.40)
        };

    \addplot[
        color=orange,
        mark=square*,
        mark size=1.5pt,thick
        ]
        coordinates {
        (1,71.93)(2,80.27)(3,75.00)(4,77.75)
        };
        
    \addplot[
        color=green,
        mark=triangle*,
        mark size=1.5pt,thick
        ]
        coordinates {
        (1,68.55)(2,77.20)(3,77.04)(4,79.64)
        };
        
    \addplot[
        color=red,
        mark=diamond*,
        mark size=1.5pt,thick
        ]
        coordinates {
        (1,81.53)(2,82.39)(3,83.73)(4,85.33)
        };
        
    \end{axis}
    \end{tikzpicture}
    \end{minipage}
    \hfill
    \begin{minipage}[b]{0.48\textwidth}
    \begin{tikzpicture}
    \begin{axis}[
        width=\textwidth, height=4cm,
        title={OSWorld-G Performance},
        title style={font=\footnotesize},
        xlabel={Training Data (\%)},
        xlabel style={yshift=1ex, font=\footnotesize},
        ylabel={Success Rate (\%)},
        ylabel near ticks,
        ylabel style={font=\footnotesize},
        grid=both,
        xmin=1, xmax=4,
        ymin=25, ymax=45,
        xtick={1, 2, 3, 4},
        xticklabels={10\%, 20\%, 50\%, 100\%},
        xticklabel style={font=\footnotesize},
        yticklabel style={font=\footnotesize},
        ymajorgrids=true,
        grid style=dashed,
    ]
    
    \addplot[
        color=blue,
        mark=*,
        mark size=1.5pt,thick
        ]
        coordinates {
        (1,27.30)(2,28.55)(3,28.72)(4,29.08)
        };

    \addplot[
        color=orange,
        mark=square*,
        mark size=1.5pt,thick
        ]
        coordinates {
        (1,28.55)(2,29.43)(3,26.77)(4,28.72)
        };
        
    \addplot[
        color=green,
        mark=triangle*,
        mark size=1.5pt,thick
        ]
        coordinates {
        (1,27.66)(2,28.72)(3,32.62)(4,33.87)
        };
        
    \addplot[
        color=red,
        mark=diamond*,
        mark size=1.5pt,thick
        ]
        coordinates {
        (1,33.87)(2,35.82)(3,37.94)(4,39.01)
        };
        
    \end{axis}
    \end{tikzpicture}
    \end{minipage}
    \caption{The effect of training data percentage on Qwen2.5-VL-3B model performance across different UI elements. Blue line: Icon; Orange line: Component; Green line: Layout; Red line: All. Left: ScreenSpot-v2 benchmark; Right: \ourbenchmarkname benchmark.}
    \vspace{-10pt}
    \label{fig:training_data_effect}
\end{figure}
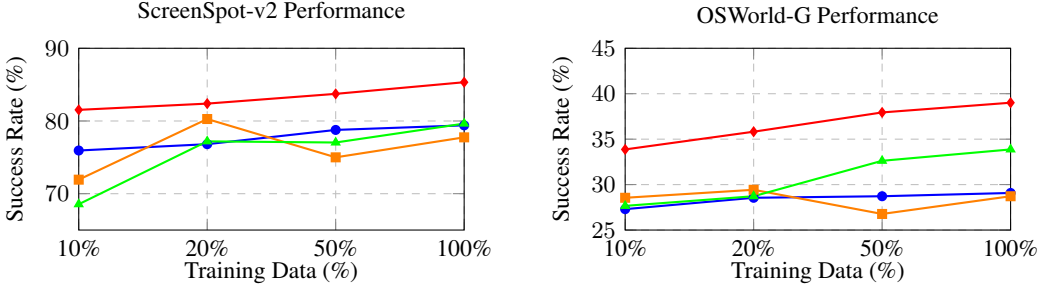

We aim to investigate whether collecting data through our pipeline enables further performance improvements as the data scale increases. 
We sample data of icon, component, and layout at proportions of 10\%, 20\%, 50\%, and 100\%. 
For each data proportion, we train the models for the same number of steps, ensuring that all models are sufficiently trained to allow a fair comparison of final performance under equal computational resources. 
The results are shown in Figure~\ref{fig:training_data_effect}.
First, we observe that as the data scale increases, model performance continues to improve, with no sign of saturation. 
This suggests that further scaling up the data using our proposed approach can yield additional gains.
Second, we note that scaling up a single data type (e.g., component) can lead to performance fluctuations. 
In contrast, scaling up mixed data types results in more stable improvements, indicating that combining data from multiple sources is beneficial.

\vspace{-5pt}
\subsection{Case Study}
\vspace{-5pt}

\input{images/case_study}

We conducted a detailed comparison of \ourmodelsevenbillion and Qwen2.5-VL-7B-Instruct using \ourbenchmarkname. 
To illustrate the improvements of \ours, we selected representative cases where their results differed, as shown in Figure~\ref{fig:case_study}. 
In each subfigure, the green square represents the click position of \ours, which is the correct grounding action, while the red square indicates the erroneous click position of Qwen.
In these examples, \ours showcases exceptional fine-grained operational capabilities and comprehension skills in locating and matching information. 
As illustrated in the subfigure~\ref{fig:case_row1}, \ours successfully identifies the target cell without an explicit location (like ``E19'') by using information from both the timestamp and the table header. 
Similarly, by understanding the paragraph text and accurately identifying relative positions, \ours effectively addresses the case presented in the~\ref{fig:case_row2}.
Furthermore, as illustrated in the~\ref{fig:case_row3}, by learning from web page layouts, \ours exhibits generalization to desktop environments, accurately locating the specified blank cell based on the positional description. Additionally, benefiting from training on extensive icon data, \ours successfully associated the icon (a counter-clockwise arrow) with its corresponding function (``rotate''), as depicted in the ~\ref{fig:case_row4}. 
Further analysis of additional examples can be found in Appendix~\ref{appendix:case_success} and ~\ref{appendix:case_failure}.

%% file: images/case_study.tex
\begin{figure}[htbp]
    \centering

    \begin{subfigure}[t]{0.48\textwidth}
        \centering
        \includegraphics[width=\linewidth,height=0.2\textheight,keepaspectratio]{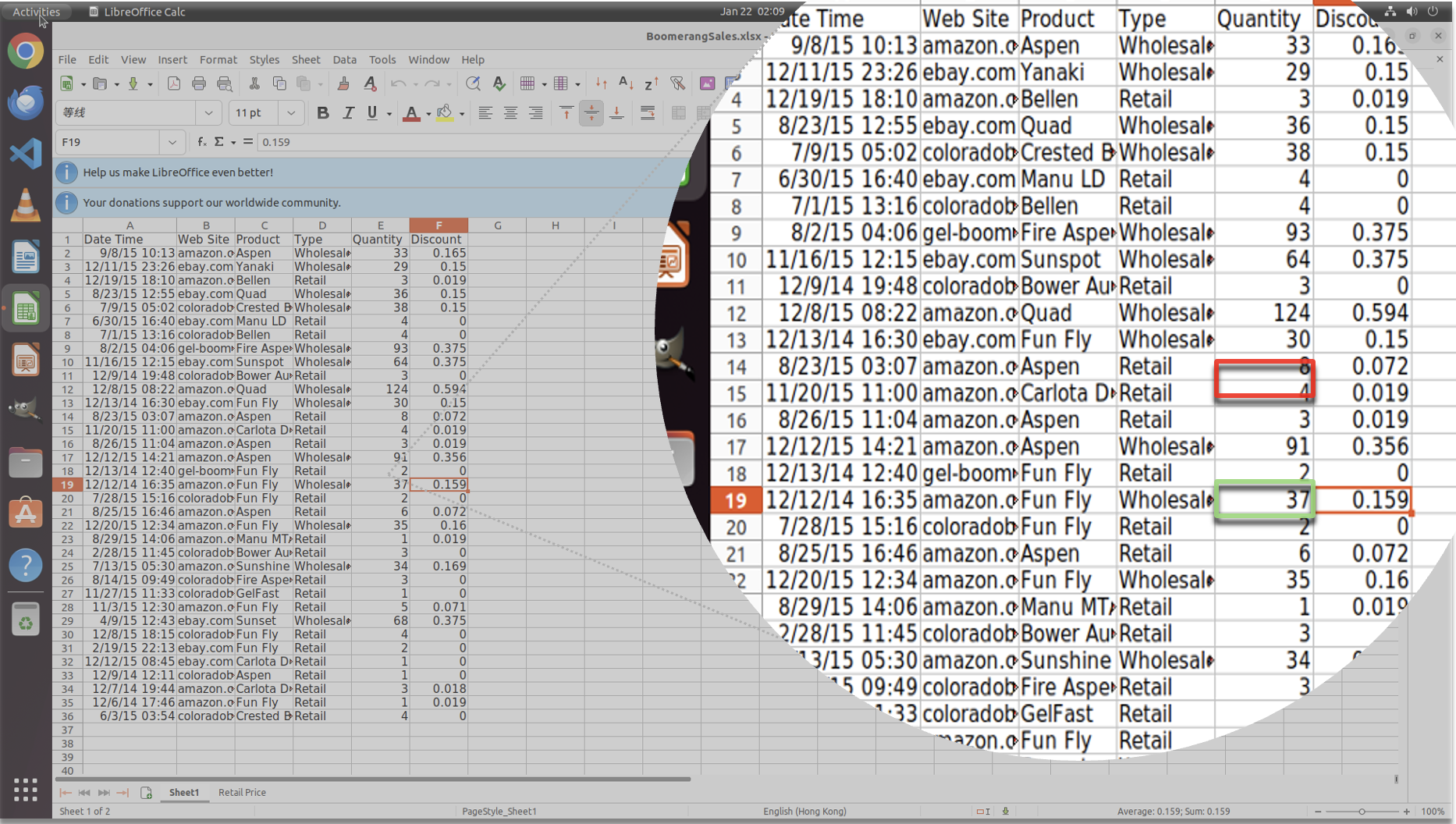}
        \subcaption{Instruction: Click on the quantity of product in 12/12/14 16:35.}
        \label{fig:case_row1}
    \end{subfigure}
    \hfill
    \begin{subfigure}[t]{0.48\textwidth}
        \centering
        \includegraphics[width=\linewidth,height=0.2\textheight,keepaspectratio]{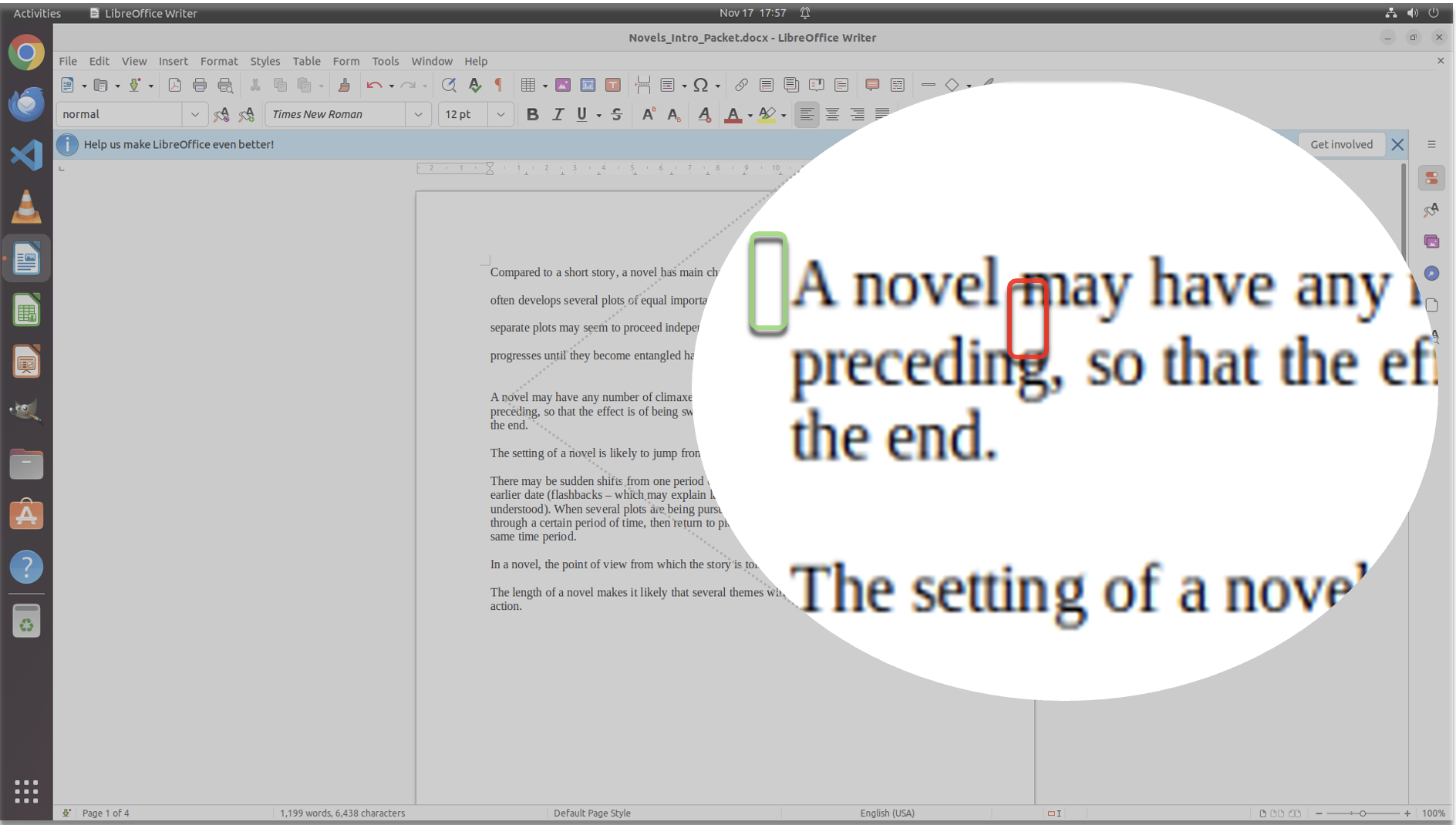}
        \subcaption{Instruction: Place the cursor before the capital 'A' in the paragraph about novel climaxes.}
        \label{fig:case_row2}
    \end{subfigure}

    \vspace{0.5\baselineskip} 

    \begin{subfigure}[t]{0.48\textwidth}
        \centering
        \includegraphics[width=\linewidth,height=0.2\textheight,keepaspectratio]{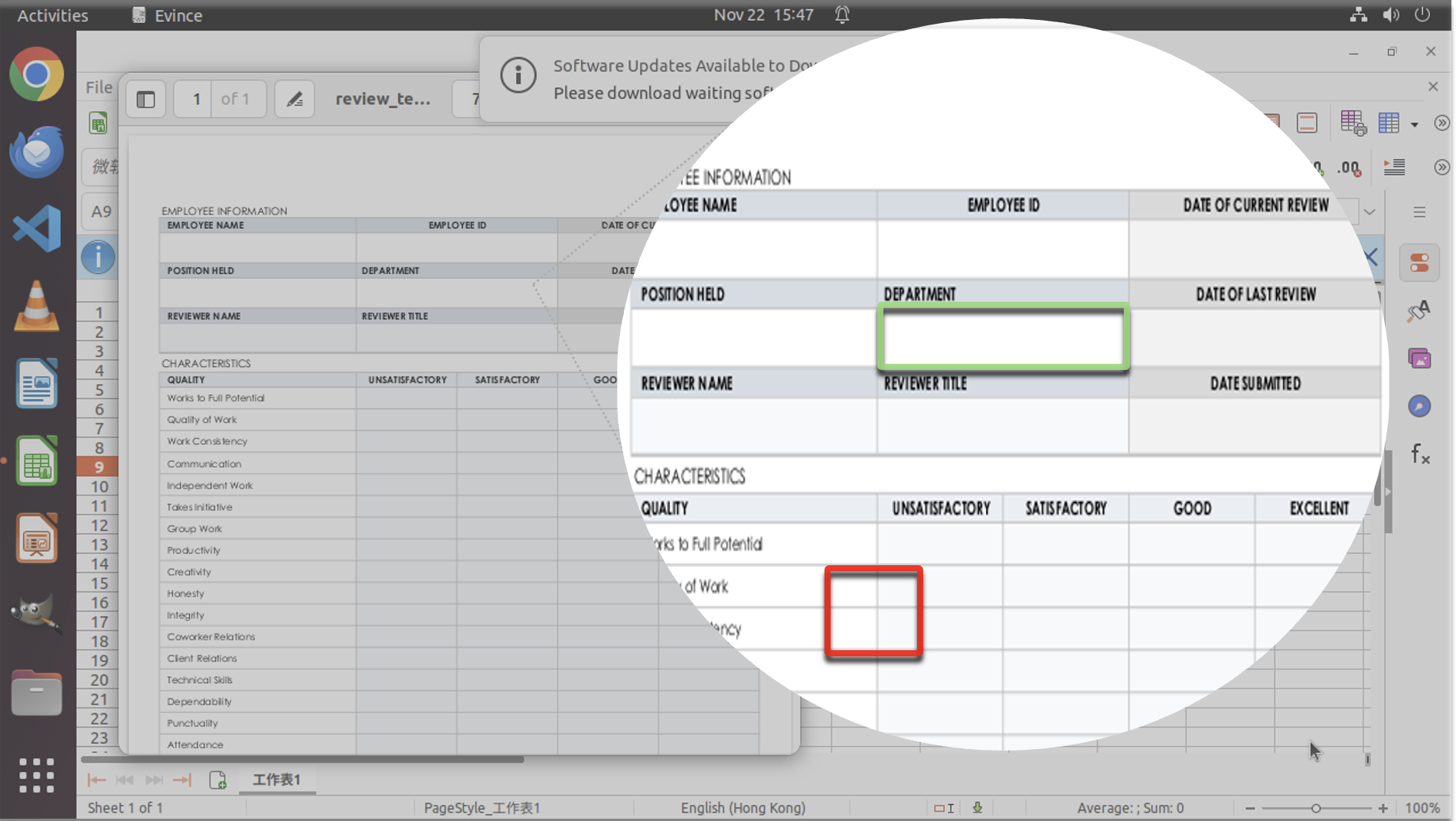}
        \subcaption{Instruction: Fill up the middle space of the second blank line in the visualized information form.}
        \label{fig:case_row3}
    \end{subfigure}
    \hfill
    \begin{subfigure}[t]{0.48\textwidth}
        \centering
        \includegraphics[width=\linewidth,height=0.2\textheight,keepaspectratio]{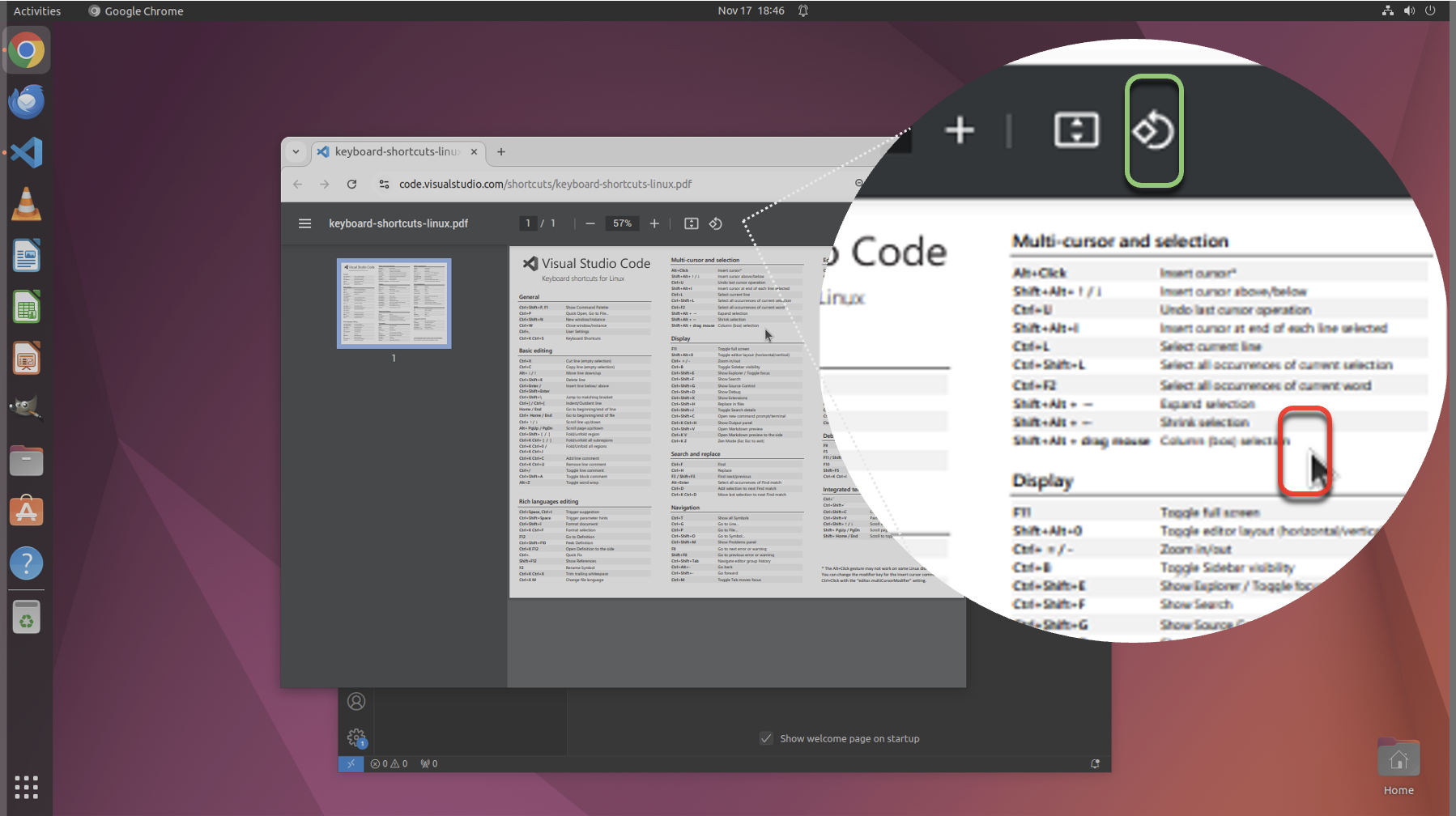}
        \subcaption{Instruction: Button to rotate the pdf.}
        \label{fig:case_row4}
    \end{subfigure}

    \caption{Qualitative comparison showing \ours's enhanced fine-grained operation and GUI understanding compared to Qwen2.5-VL-7B-Instruct across four cases. The green square represents the click position of \ours, while the red square indicates the click position of Qwen.}
    \vspace{-10pt}
    \label{fig:case_study}
\end{figure}

%% file: text/related.tex
\vspace{-5pt}
\section{Related Work}
\label{sec:related}
\vspace{-5pt}

\paragraph{Digital Agents}
Multimodal agents can be broadly categorized into digital and physical agents~\cite{shi2017world, shridhar2020alfred, fan2022minedojo}.
Existing digital agent research focuses on establishing environments for mobile and web interaction~\cite{shi2017world, liu2018reinforcement, nakano2021webgpt, toyama2021androidenv, yao2022webshop, zhou2023webarena, rawles2023android, zhang2023mobile, Koh2024VisualWebArenaEM, drouin2024workarena, wang2025computer}, with subsequent works extending to real-world computer interaction scenarios~\cite{xie2024osworldbenchmarkingmultimodalagents, Bonatti2024WindowsAA}. 
Recent advances include enhanced GUI understanding through visual encoding architectures~\cite{hong2023cogagent, bai2025qwen25vltechnicalreport}, reinforcement learning frameworks introduced to web/mobile operations~\citep{qi2024webrl, bai2025digirl}, agentic-frameworks~\citep{zheng2024gpt, gou2024navigatingdigitalworldhumans, agashe2024agentsopenagentic, yang2024aria, liu2025pc} and joint visual-language modeling~\citep{Wu2024OSATLASAF, xu2024aguvis, qin2025uitarspioneeringautomatedgui}.
However, current methods face precision limitations on grounding due to homogeneous synthetic training data~\cite{cheng2024seeclickharnessingguigrounding, gou2024navigatingdigitalworldhumans, yang2024aria, sun2024genesis}, which overlook the systematic support for fine-grained component operations (e.g., slider adjustments, nested menu selections), finally limits the upper policy execution as well as further learning.
Furthermore, the sources of data which could be beneficial for enhancing GUI interaction abilities are underexplored.

\paragraph{GUI Grounding}
GUI grounding remains a core challenge for digital agents executing actions in real world environment.
Recent approaches have shifted from relying on textual information such as HTML/accessibility information to pure visual solutions~\cite{cheng2024seeclickharnessingguigrounding, zheng2024gpt, gou2024navigatingdigitalworldhumans, lu2024omniparser, lin2024showui, yu2025omniparser}.
However, both existing training data and evaluation paradigms suffer from oversimplification—whether through screenshot-text pairings or manual annotations—failing to capture the complexity of natural language instructions and action execution, particularly in tasks requiring understanding of expressed intent rather than simple referencing, screen-level comprehension (such as identifying active windows), and fine-grained operations (like sliders and drag-and-drop), thus hindering meaningful assessment and advancement in these critical areas.
We point out the problems by proposed benchmark and bridging these gaps through multiple aspects of synthetic data.
The comparison with previous work is shown in Table~\ref{tab:benchmark_comparison}.

%% file: text/conclusion.tex
\vspace{-5pt}
\section{Conclusion}
\label{sec:conclusion}
\vspace{-5pt}

We highlight overlooked GUI grounding challenges such as fine-grained manipulation and layout understanding, introducing \ourbenchmarkname with \numexamples annotated samples for evaluation. 
We set up multiple pipelines to construct a dataset containing \numdataexamples examples to address these challenges. 
Our models trained on this dataset achieve competitive results on ScreenSpot-v2, ScreenSpot-Pro, and \ourbenchmarkname, while also boosting agent performance in OSWorld and WindowsAgentArena. 
These results demonstrate the effectiveness of addressing previously identified gaps in GUI grounding research.

%% file: text/acknowledgements.tex
\section*{Acknowledgements}
We thank Binyuan Hui, Weilu Xu, Dunjie Lu, Zhiyong Wu, Weiyun Wang,  Hao Hu, Bowen Wang, Eric Xin Wang, Yuhao Yang, Junlei Zhang, Victor Zhong, Yujia Qin for their helpful feedback on discussion around this work.

%% file: text/limitations.tex
\section{Limitations}
\label{sec:limitations}

In this work, we mainly discuss the data synthesis methods while figuring out the essential factors.
Screen capture data can be extracted from internet images and videos by neural networks, which can further expand the dataset.
This approach can significantly expand the screenshot metadata, thus enlarging the grounding data.
Due to resource restrictions, we leave this for further scaling through industrial efforts.
Rejecting infeasible actions is crucial, as it helps prevent errors and mitigates the risks associated with incorrect instructions. Refusal modeling in GUI grounding remains a significant challenge, as models show limited improvement due to the inherent limitations in pretraining and the hallucination phenomenon in VLMs. While we find this problem has inherent complexity and challenges, this provides direction for future in-depth research and optimization.
On the other hand, based on our enhanced grounding model, we can construct human-like traversers that interact in the digital world with or without specific purposes, similar to how humans navigate digital environments. 
This approach can further collect interaction data to improve grounding capabilities and even enhance model knowledge. 
We also leave these explorations for future work.

%% file: text/appendix.tex
\section{Appendix}

\subsection{\ourbenchmarkname Statistics}

\subsubsection{Data Types}\label{app:og_data_types}
We categorize the examples into five categories that requires different grounding capabilities. 
And the classification can be refer to their corresponding element types in the Table \ref{tab:app-capabilities-classification}.

\input{tables/appendix_og_types_full}

\subsubsection{Comparison with Previous Work}
We show the comparison between \ourbenchmarkname and previous work in Table~\ref{tab:benchmark_comparison}.

\input{tables/bmk_comparison}

\subsubsection{Data Examples}

\begin{figure}[htbp!]
    \centering
    \begin{subfigure}[b]{0.32\textwidth}
        \centering
        \includegraphics[width=\textwidth]{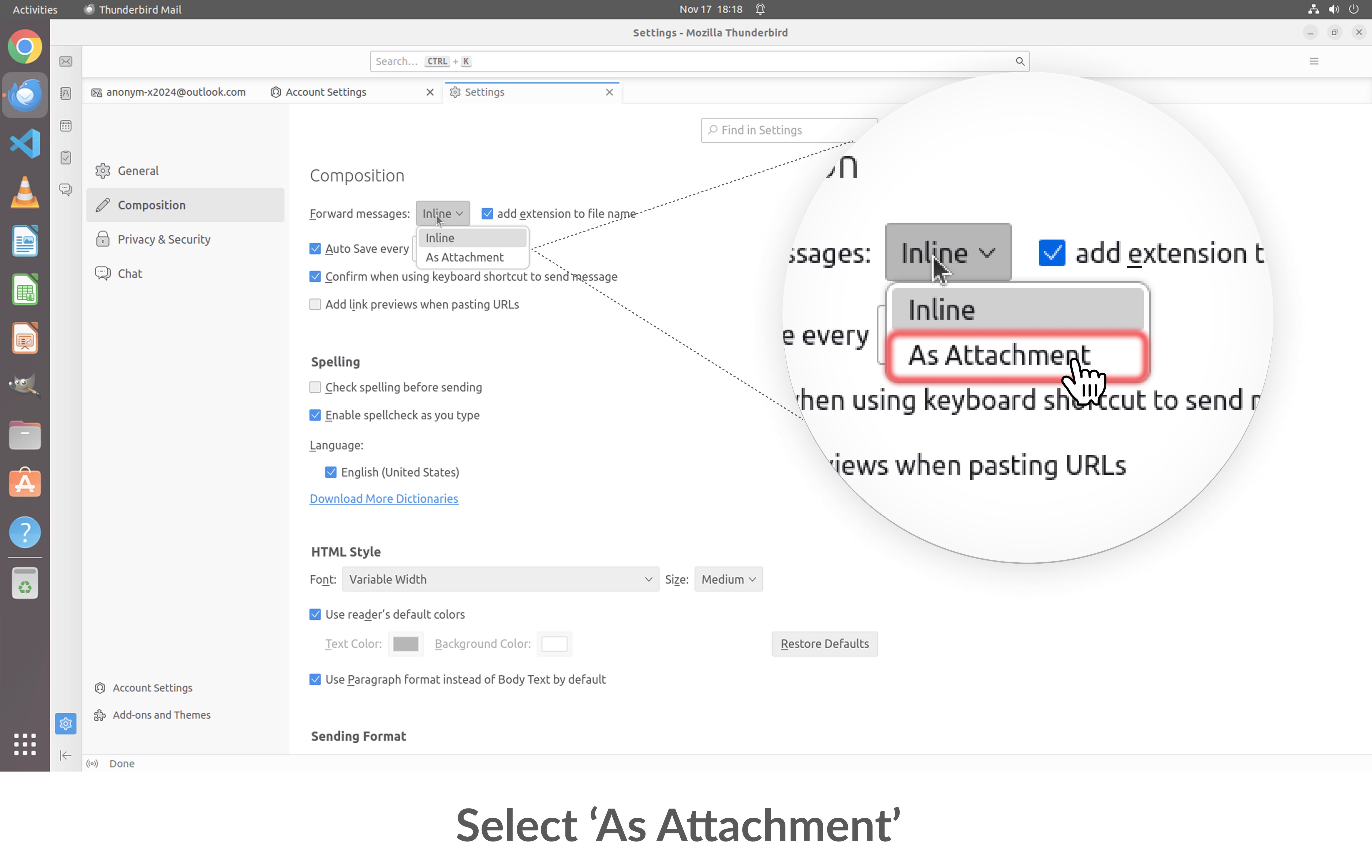}
        \caption{Text matching}
        \label{fig:benchmark_example_text_matching}
    \end{subfigure}
    \hfill
    \begin{subfigure}[b]{0.32\textwidth}
        \centering
        \includegraphics[width=\textwidth]{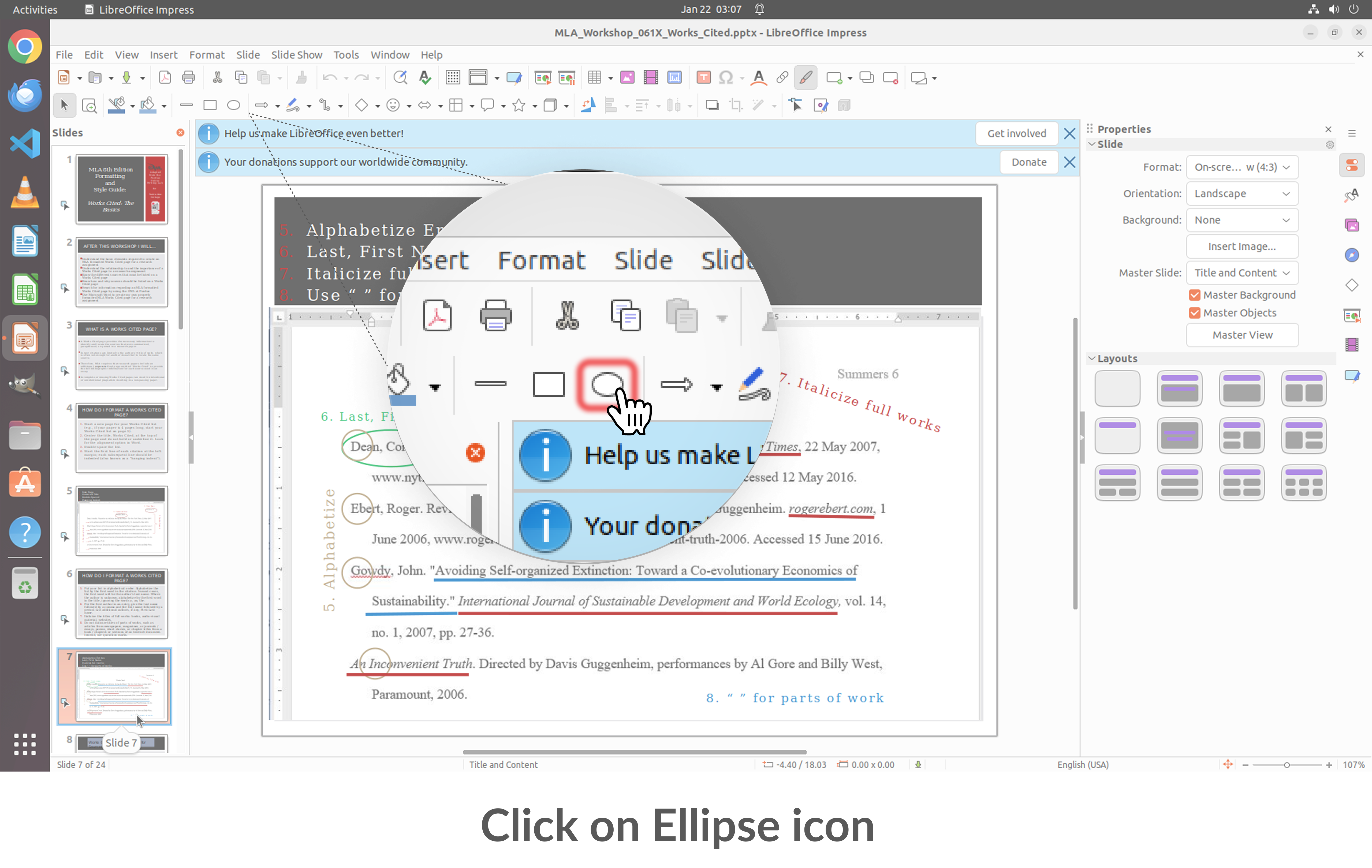}
        \caption{Element recognition}
        \label{fig:benchmark_example_element_recognition}
    \end{subfigure}
    \hfill
    \begin{subfigure}[b]{0.32\textwidth}
        \centering
        \includegraphics[width=\textwidth]{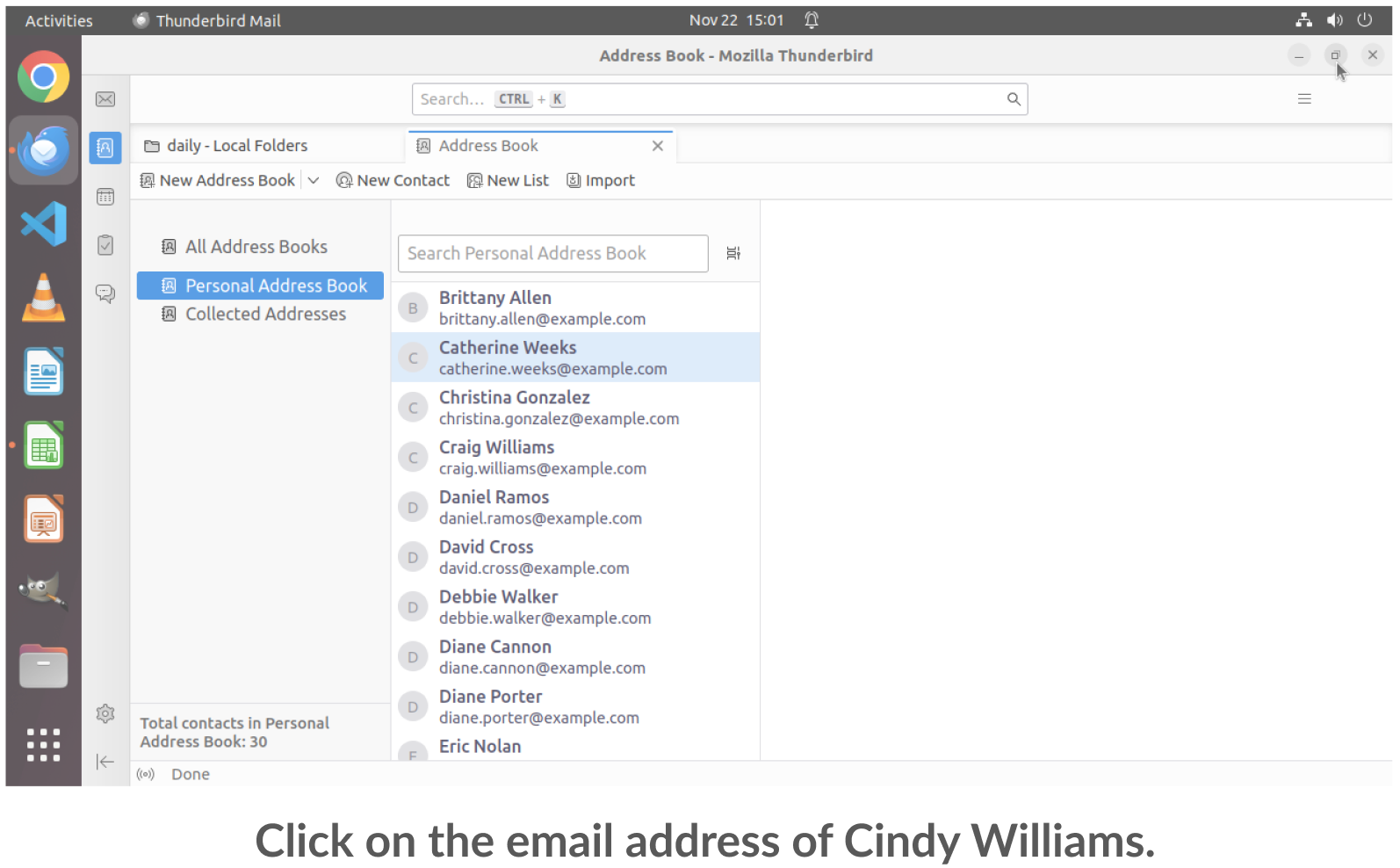}
        \caption{Refusal instruction}
        \label{fig:benchmark_example_refusal_inst}
    \end{subfigure}
    \caption{Examples in \ourbenchmarkname that require text matching and element recognition abilities.}
    \label{fig:benchmark_example_text_matching_element_recognition}
\end{figure}

We show examples of text matching type and element recognition type in Figure ~\ref{fig:benchmark_case_layout_understanding} (layout understanding), ~\ref{fig:benchmark_case_finegrained_manipulation} (fine-grained manipulation), ~\ref{fig:benchmark_example_text_matching_element_recognition} (text matching, element recognition and refusal instruction).

\paragraph{Layout Understanding}
Layout understanding tasks require models to comprehend the hierarchical structure of interface elements. In the example shown in Figure~\ref{fig:benchmark_case_layout_understanding}, closing the top notification bar requires recognizing that such bars typically appear at the top region of the editing area in Libreoffice Calc.

\paragraph{Fine-grained Manipulation}
Fine-grained manipulation tasks demand high-precision actions within small or tightly packed screen regions. In the example in Figure~\ref{fig:benchmark_case_finegrained_manipulation}, selecting the position between the word "person" and the number "1" requires the model to operate at a character-level granularity.

\paragraph{Text Matching}
Text matching tasks involve grounding actions based on explicit textual cues in the instruction. As shown in Figure~\ref{fig:benchmark_example_text_matching}, choosing "As Attachment" requires the model to locate and match this phrase within the screenshot.

\paragraph{Element Recognition}
Element recognition tasks require identifying visual patterns such as icons or images. In the example in Figure~\ref{fig:benchmark_example_element_recognition}, clicking on the ellipse icon involves recognizing the ellipse shape visually within the interface.

\paragraph{Refusal Instruction}
Refusal instruction tasks assess whether the model can recognize when an action is infeasible. In the example in Figure~\ref{fig:benchmark_example_refusal_inst}, the instruction refers to "Cindy Williams," who is not visible on the screen. Therefore, clicking on her email address is not possible, and the model is expected to refrain from taking action.

\subsubsection{Annotation Details}\label{sub:annotation_details}

The annotation process for \ourbenchmarkname{} comprised the following systematic steps:

\begin{enumerate}
    \item \textbf{Failure Case Collection:} We systematically gathered grounding failure cases from state-of-the-art model trajectories, categorizing each failure according to its primary grounding capability requirements.
    
    \item \textbf{Expert Annotation:} Annotators with extensive experience in various software applications performed initial precise annotations. Using the collected failure cases as guidance, they crafted descriptive low-level instructions that were designed to be unambiguous and map uniquely to specific screen actions.
    
    \item \textbf{Bounding Box Annotation:} For each instruction-screenshot pair, corresponding bounding boxes were carefully annotated to indicate the precise regions of interest.
    
    \item \textbf{Quality Verification:} We conducted multi-round verification procedures, leveraging predictions from strong models to resolve cases with inconsistent annotation results.
\end{enumerate}

All examples in \ourbenchmarkname{} were annotated following this rigorous process to ensure high-quality and consistent annotations.

\subsection{\ours Statistics}\label{sub:data_statistics}

\subsubsection{Overview}
Table \ref{tab:data_stats} provides an overview of the statistics for \ours. To enhance the quality of our dataset, we made several improvements upon \textsc{Aguvis}, where we name it \textsc{Aguvis}++. 
First, we manually filter out low-quality annotations and samples unrelated to computer use scenarios. We then augment the dataset by incorporating OS-Atlas data. For synthetic data sources such as SeeClick and OS-Atlas, which inherently contain rendering artifacts and alignment issues, we employ UI-TARS~\footnote{\url{https://huggingface.co/ByteDance-Seed/UI-TARS-72B-DPO}} model for quality control—comparing predicted outputs against ground truth values to ensure deviations remained within acceptable thresholds. 
In-house data is annotated by human workers. 
We ask them to use computers while recording timestamps of their actions and capturing screenshots from their screens as observations.
These are later used with models like GPT to construct input instructions.

\input{tables/data_statistics}

\newpage
\subsubsection{Icon Statistics}\label{sub:icon_data_sources}
The Source Statistics of icon data in \ours are detailed in Table~\ref{tab:icon_data_srcs}.
\input{tables/icon_data_srcs}

\newpage
\subsubsection{Component Statistics}\label{sub:component_data_sources_and_pipelines}
The following Table~\ref{tab:component_data_srcs} provides a detailed list of the component libraries we use, along with the contribution of each component to the \ours dataset.
\input{tables/component_data_srcs}

\subsubsection{Layout Statistics}

We collected layout data from two primary sources: the UI design community Figma and through rollouts across operating systems. Statistics can be found in the following Tables \ref{tab:rollout_stats} \ref{tab:figma_stats}. The number of elements shown in the following table is not the exact number in the final dataset. These elements are filtered in later processing stage.

\input{tables/appendix_layout_os_stats}
\input{tables/appendix_layout_figma_stats}

\subsubsection{Cost Analysis}\label{sub:cost_analysis}

We utilized GPT-4o throughout our data generation pipeline, with all cost calculations based on GPT-4o's token pricing. The comprehensive cost breakdown for the three components of the \ours dataset is detailed below:

\begin{itemize}
    \item \textbf{Icon Data (0.4M samples):} We employed input prompts and images to generate visual and functionality descriptions for icons. Each sample incurred approximately \$0.01 in processing costs, resulting in a total expenditure of \textasciitilde\$4,000.
    
    \item \textbf{Component Data (1M samples):} This category comprised two distinct subsets:
    \begin{itemize}
        \item \textit{Template-based fine-grained operations} (\textasciitilde40K samples): Generated using predefined template rules for slides and sheets data, incurring no additional costs.
        \item \textit{Code-rendered data} (\textasciitilde1M samples): Costs were distributed across component rendering, action generation, and filtering processes, averaging \textasciitilde\$0.025 per sample, totaling \textasciitilde\$25,000.
    \end{itemize}
    
    \item \textbf{Layout Data (2.3M samples from 0.8M captions):} We leveraged GPT-4o to generate comprehensive screenshot captions. Each caption required processing of approximately 3 images (\textasciitilde2,100 tokens) plus prompts (\textasciitilde550 tokens), with an average output of \textasciitilde250 tokens. This resulted in a cost of \textasciitilde\$0.0091 per caption, totaling \textasciitilde\$7,000 for the complete set of 0.8M captions.
\end{itemize}
The aggregate cost for utilizing GPT-4o across all data generation tasks amounted to approximately \textbf{\$36,000}.

\newpage
\subsection{\ours Dataset Construction: A Detailed Pipeline for Component}\label{appendix:data-construction}

\subsubsection{Component Collection and Style Augmentation}

We begin by collecting example components from four mainstream UI libraries hosted on GitHub: \texttt{Material UI}, \texttt{Ant Design}, \texttt{Mantine UI}, and \texttt{Chakra UI}. From each repository, we extract example code snippets(in \texttt{typescript}) that showcase usage of individual components.

To diversify these examples, we apply style augmentation using two LLMs: \texttt{GPT-4o} and \texttt{Claude-3.5-Sonnet}. For each original code snippet, we first ask the model to envision a unique UI usage scenario. Based on the original code and the imagined context, it then generates a stylistically augmented variant code.

This process is repeated multiple times per example, each time with a different context to promote diversity. Previously generated variants are included in the prompt to prevent redundancy across augmented examples.
    
\subsubsection{Rendering and Interaction Preparation}

Each augmented component is rendered on a React application. Components are wrapped in a container with a randomized position to mitigate positional overfitting. Using \texttt{Playwright}, we programmatically open and interact with the rendered pages.

We extract \textbf{screenshots} of the rendered component and\textbf{element tree information} (positioning, hierarchy, etc.) using \texttt{Playwright}'s \texttt{evaluate} method and custom JavaScript.

These outputs are used to generate component-grounded actions via two distinct pipelines.

\subsubsection*{Pipeline 1: Component-level Action Generation}

\paragraph{Step 1: Generate Action Intents}~\par
\index We prompt GPT-4o with component name, component code and screenshot. GPT-4o returns a list of action intents, each representing a high-level user interaction. We use few-shot examples to guide this process.

\paragraph{Step 2: Generate Action Details}~\par
\index For each intent, we generate detailed interaction metadata using component name, component code, screenshot, element position tree and action intent.

Each action detail includes:
\begin{enumerate}
    \item \textbf{Thought Process}: The thinking process of generating an action detail
    \item \textbf{Action Space Type}:
    \begin{itemize}
        \item \texttt{None}: No action space exists, 
        \item \texttt{Unique}: Only one possible action exists (e.g., clicking a button), 
        \item \texttt{Discrete}: Limited/unlimited set of distinct possible actions (e.g., selecting from a list of options), 
        \item \texttt{Continuous}:  Infinite possible actions within a range (e.g., dragging a slider to any position)
    \end{itemize}
    \item \textbf{Action Description}: Describe what the action does, which serves as the instantiation/implementation of the action intent.
    \item \textbf{Action Parameters}: List of all parameter names for the action function(in action code)
    \item \textbf{Discrete Values}: List of all possible parameter values for discrete action spaces (if applicable)
    \item \textbf{Continuous Intervals}: List of interval for all possible parameter values for continuous action spaces (if applicable)
    \item \textbf{Action Code}: A function using \texttt{PyAutoGUI} to represent one action or a kind of actions 
\end{enumerate}

\paragraph{Example}~\par
\begin{lstlisting}[language=json]
{
    "thought_process": "The target element is a slider, which provides a continuous range of values from 0 to 100. The action involves setting a specific value within this range by determining the corresponding position on the slider bar and simulating a click at that position. The slider's endpoints are identified, and linear interpolation will be used to calculate the appropriate position based on the desired value.",
    "action_space_type": "continuous",
    "action_desc": "Set saturation to <saturation>%",
    "action_params": [
        "saturation"
    ],
    "action_discrete_values": null,
    "action_continuous_interval": {
        "saturation": [
            [
                0.0,
                100.0
            ]
        ]
    },
    "action_code": "def action(saturation):\n    x_0, y_0 = 600.5, 830  # Left endpoint\n    x_1, y_1 = 1064.5, 830  # Right endpoint\n    x = x_0 + (x_1 - x_0) * (saturation / 100)\n    pyautogui.click(x, y_0)"
}   
\end{lstlisting}

We then convert the action code (e.g., \texttt{def action(parameter): ...}, often involving \texttt{pyautogui}) into one or more pieces of \textit{grounding data}—such as \texttt{pyautogui.click(x, y)}—by sampling from the corresponding action space. If the action space is \texttt{None}, no sampling is needed. This conversion is guided using few-shot examples. An example of this process can be seen below.

\paragraph{Example}~\par
\textbf{Instruction:} \textit{Set saturation to \texttt{<saturation>}}

\textbf{Action code:}
\begin{verbatim}
def action(saturation):
    x_0, y_0 = 600.5, 830   # Left endpoint of the saturation slider
    x_1, y_1 = 1064.5, 830  # Right endpoint of the saturation slider
    x = x_0 + (x_1 - x_0) * (saturation / 100)
    pyautogui.click(x, y_0)
\end{verbatim}

\textbf{Sampled grounding data:}
\begin{verbatim}
# Set saturation to 24%
pyautogui.click(711.86, 830)

# Set saturation to 60%
pyautogui.click(878.90, 830)
...
\end{verbatim}

\subsubsection*{Pipeline 2: Element-Level Action Generation}

\paragraph{Step 1: Element Extraction and Filtering}
We render each augmented component in a browser and traverse the DOM tree to collect element nodes. Two filtering rules are applied:

\begin{itemize}
    \item \textbf{Duplicate boxes}: Only one node is retained if multiple share the same bounding box.
    \item \textbf{Abnormal sizes}: Nodes with very small or very large bounding boxes are discarded.
\end{itemize}

For each valid node, we collect position, text, visibility, interactivity, parent-child relationships, and metadata.

\paragraph{Step 2: Multimodal Context Encoding}
To help GPT-4o understand each element, we provide element box, parent box, cropped screenshot(cropped screenshot with only the element region), context screenshot(cropped screenshot with element region and nearby surroundings, with the element highlighted in red bounding box) and full-page screenshot(full screenshot with the element highlighted in red bounding box) as input. And the model outputs include visual description(a detailed account of the element's appearance), position textual information(spatial relationship relative to the viewport and its parent), element functionality, UI type (e.g., button, slider) and possible actions at element center.

To ensure quality, we also include visibility check and atomicity check, to check whether this element is a single visible UI unit.

\paragraph{Step 3: Action Detail Generation}
For each possible action, we prompt GPT-4o with the action and relevant element information—including visual description, position, text content, functionality, and UI type. The model is asked to generate detailed action information, including the thought process, action description, action parameters, and action code. This is similar to the action detail in Pipeline 1, but limited to the \textbf{unique action space}.

\paragraph{Step 4: Continuous Action Detection}
To identify elements like sliders that support continuous interactions, the model determines whether the element has a continuous action space and generates the corresponding thought process, action description, action parameters, value range(action\_continuous\_interval), and action code. This step parallels the action detail in Pipeline 1, but focuses solely on the \textbf{continuous action space}.

\paragraph{Step 5: Grounding Actions}
We convert each action code into one or more grounding samples, similar to that in Pipeline 1.

\subsubsection{Comparison of Pipelines}

\begin{itemize}
    \item \textbf{Pipeline 1} is simpler. However, it may suffer from inaccurate bounding box targeting, limited action diversity and action vagueness.
    \item \textbf{Pipeline 2} generates data with better localization and diversity. \textbf{In practice, most of our dataset is generated using Pipeline 2.}
\end{itemize}

\subsubsection{Post-Processing and Filtering}

To ensure data quality, we apply multiple filtering stages.

\textbf{1. Visual Filter (via \texttt{GPT-4o})}

Given:
\begin{itemize}
    \item Cropped screenshot
    \item Marked screenshot (click position highlighted with a green dot and circle)
    \item Full screenshot (element highlighted)
\end{itemize}

\texttt{GPT-4o} filters out data that:
\begin{enumerate}
    \item Shows visible errors (e.g., "Compiled with problems" or red overlays)
    \item Targets an incorrect GUI element
    \item Has incorrect click localization (e.g., not centered on button/text)
\end{enumerate}

\textbf{2. Instruction Filter (LLM-Based)}

Using \texttt{GPT-4o-mini}, we filter out ambiguous or low-quality instructions from Pipeline 1:
\begin{enumerate}
    \item Unclear or vague semantics
    \item Multiple interactive targets
    \item References to non-visual identifiers like "index 1"
    \item Multi-step or compound interactions
\end{enumerate}

\textbf{3. Instruction Filter (Rule-Based)}

We filter instructions with high error likelihood based on pattern rules:
\begin{enumerate}
    \item Contains explicit coordinates (e.g., \texttt{(x, y)}): Instructions referencing raw screen coordinates are filtered out, as such positional references are not meaningful in a vision-only context.
    \item Mentions structural terms such as \texttt{child}, \texttt{parent}, \texttt{path}, or \texttt{container}: These terms imply hierarchical relationships derived from accessibility trees, which are not observable in visual input.
    \item Mentions a \texttt{card} component without spatial qualifiers such as \texttt{in}, \texttt{within}, or \texttt{at}: Such instructions typically refer to an entire composite element (e.g., a card) rather than a specific atomic component within it, resulting in ambiguous interaction targets.
    \item Includes directional terms in combination with \texttt{screen}: Phrases like “top-left of the screen” are frequently found to be incorrect or misaligned with actual component layouts, likely due to LLM misinterpretation.
    \item \textbf{Refers to highlights or visual annotations} (e.g., red \texttt{dot}, \texttt{circle}, \texttt{highlight}): These often result from the model misidentifying annotation markers (used to denote interaction points) as intrinsic parts of the interface.
    \item Mentions textual UI elements (e.g., \texttt{text}, \texttt{label}, \texttt{heading}) in combination with interaction verbs (e.g., \texttt{read}, \texttt{hover}, \texttt{click}, \texttt{interact}): If the associated bounding box is visually simple—based on low color variance and edge density—it often indicates that the relevant text is located on the periphery of the box, while its center is visually empty, leading to inaccurate click localization.
    \item Refers to sliders without specifying interaction values: Instructions such as “interact with the slider” without numerical targets are prone to ambiguity and do not provide sufficient grounding for generating actionable behavior.
\end{enumerate}

\subsubsection{Real-world augmentation pipeline}
Office software, including document editors, presentation tools, and spreadsheets, is integral to daily work for many. Automating workflows in these applications can significantly boost productivity. However, a gap exists between synthetic use cases and real-world scenarios, as synthetic datasets often lack sufficient office software-related cases. To bridge this, we propose a targeted approach to designing and generating relevant data.

Our methodology centers on creating two pools: \textbf{a resource pool} and \textbf{an action pool}. 
The resource pool includes a diverse set of office files, such as Excel spreadsheets, Word documents, and PowerPoint slides, sourced from the web, including online tutorials. The action pool enumerates common tasks performed in these applications, such as scrolling through a document, clicking specific cells in a spreadsheet, or auto-filling data in Excel. 
For each action, we manually analyze the associated structural components and develop code to extract relevant coordinate arrangements.
For example, consider the action of "scrolling a document" in Microsoft Word Online. The associated component is the scrollbar. We analyze the webpage structure to identify features that precisely locate the scrollbar, then use code to extract its coordinates, synthesizing a data instance. Similarly, in Excel, for the action "click the center of cell B3," we leverage the accessibility tree and HTML DOM structure to extract cell positions, generating precise instructions like "click the center of cell B3," "auto-fill from the bottom-right corner of cell A1," or "select column D." These rule-based extraction methods ensure accurate component-level interactions across productivity applications.

Additional actions and their components, including spreadsheet-specific tasks, are detailed in Table \ref{tab:office_actions}.
\input{tables/office_data}

\subsection{Additional Data Examples}

\subsubsection{Icon data}

\begin{figure}[htbp]
    \centering
    \begin{subfigure}{0.45\textwidth}
        \includegraphics[width=\linewidth]{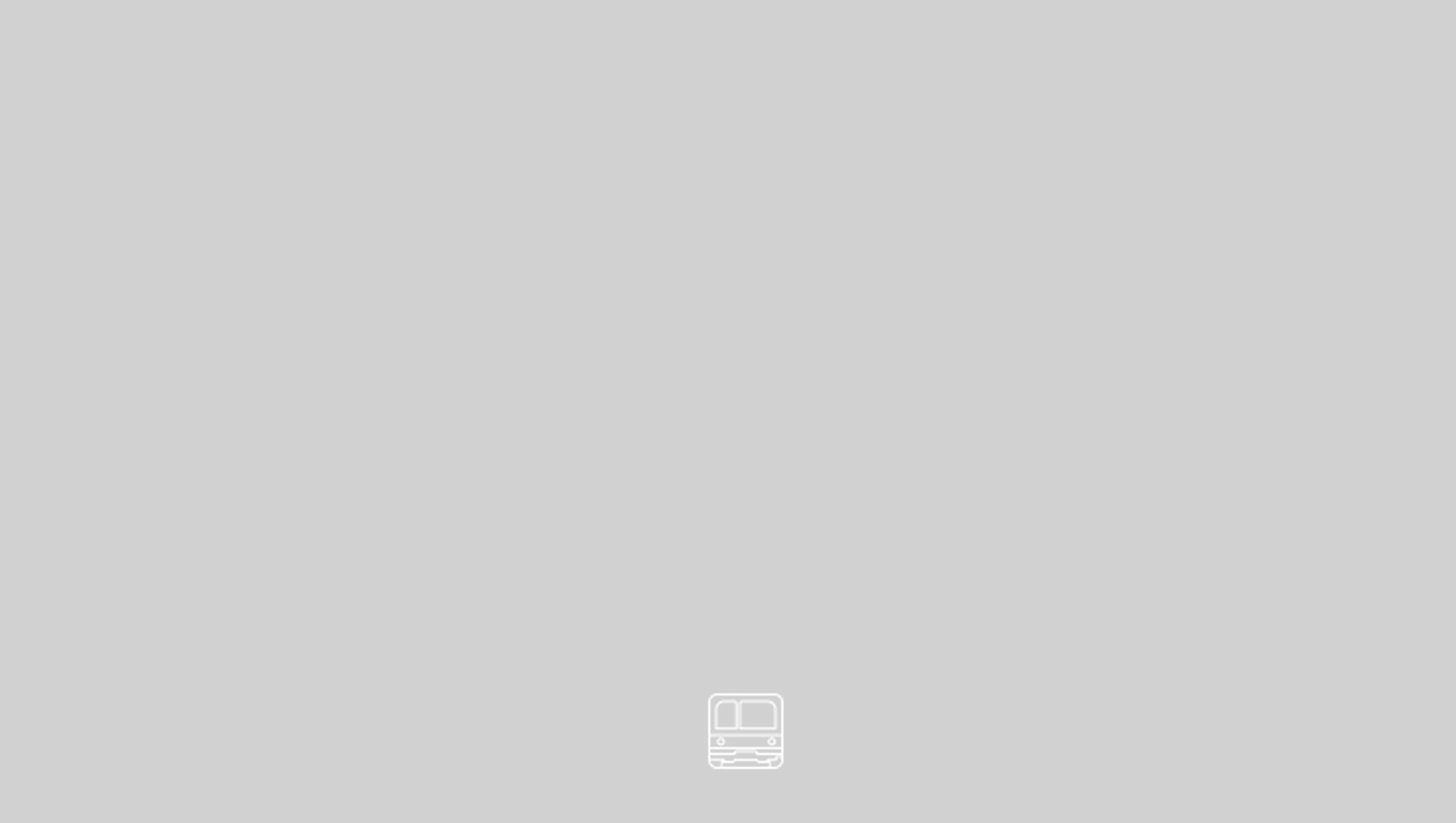}
        \caption{Example of icon description data}
        \label{fig:icon_desc}
    \end{subfigure}
    \hfill
    \begin{subfigure}{0.45\textwidth}
        \includegraphics[width=\linewidth]{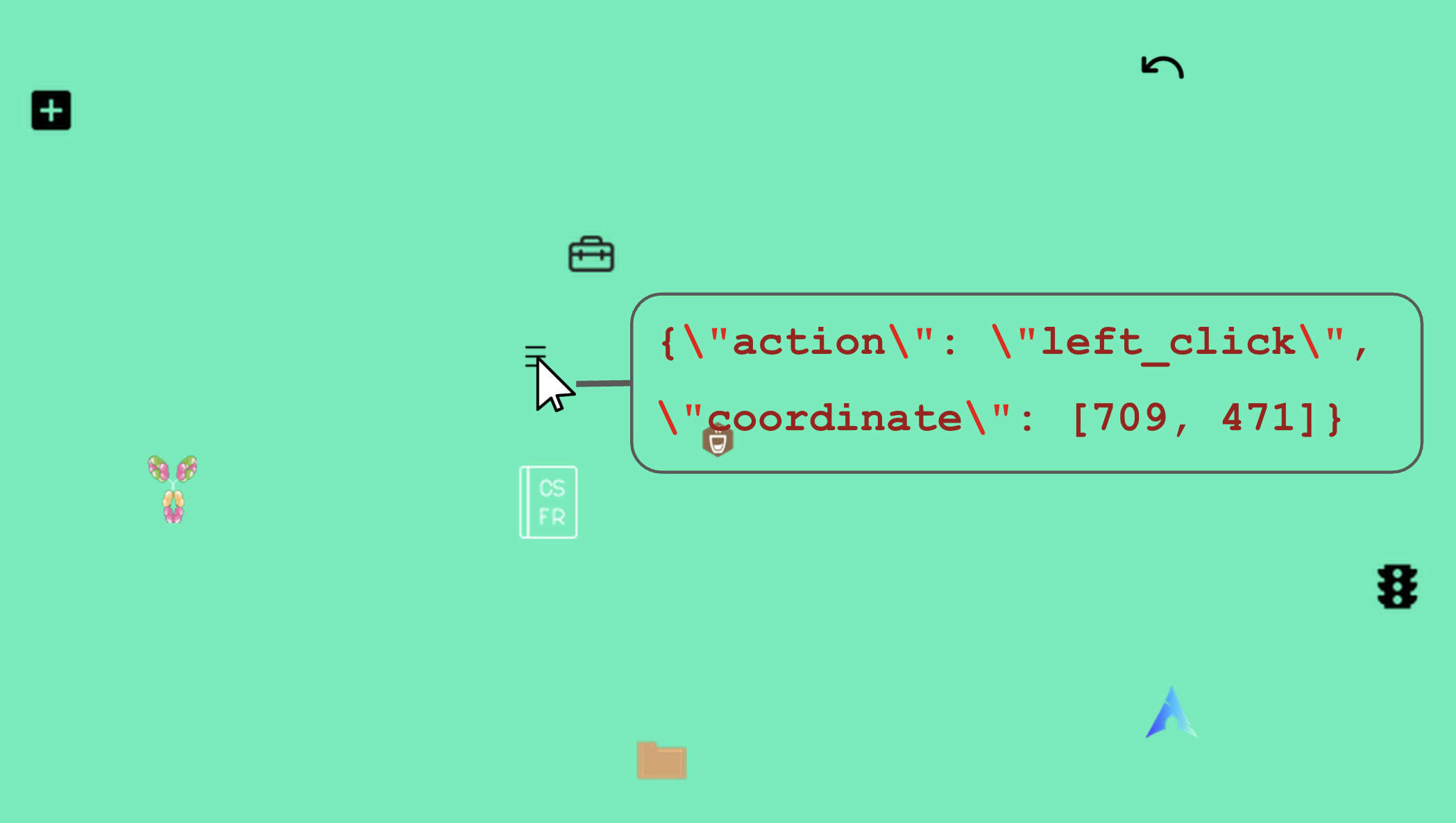}
        \caption{Example of icon grounding data}
        \label{fig:icon_ground}
    \end{subfigure}
    \caption{Examples of icon data}
    \label{fig:icon_data}
\end{figure}

\paragraph{Icon description data}~\par
In the icon description data, each data item includes an image of a single icon as input and produces a description that includes both its visual appearance and functionality as output.

An example in JSON format is shown below. The corresponding image (Figure~\ref{fig:icon_desc}) features an icon labeled krl\_access, which likely represents access to a commuter rail or train service—such as the KRL (Kereta Rel Listrik) system.

\begin{lstlisting}[language=json,breaklines=true]
{
    "conversations": [
        {
            "from": "system",
            "value": {
                "text": "You are a GUI assistant that helps users understand and interact with interface elements. \nWhen describing elements, focus on their visual appearance and functionality in clear, natural language.\nFor each element, describe the visual appearance and functionality:\n1. What it looks like - including shape, color, symbols, and any text\n2. What it does when used - its purpose and effect\n3. When users might want to use it - common use cases and scenarios"
            }
        },
        {
            "from": "user",
            "value": [
                {
                    "image": "images_pure_color_background/categorized_icons/Arcticons-Team/Arcticons/krl_access.png",
                    "width": 1920,
                    "height": 1080,
                    "resized_height": 1092,
                    "resized_width": 1932,
                    "seq_len": 2693
                },
                {
                    "text": "Please describe what this interface element looks like and what it does."
                }
            ]
        },
        {
            "from": "assistant",
            "value": [
                {
                    "text": "Visual Appearance:\nThe icon is a simple, white outline of a train or subway car on a black background. It features a rectangular shape with rounded corners, two windows at the top, and a front bumper at the bottom. The design is minimalistic and uses clean lines to represent the front view of a train.\n\nFunctionality: The icon labeled 'krl_access' likely represents access to a commuter rail or train service, such as a KRL (Kereta Rel Listrik) system. It might be used in applications or websites to indicate train schedules, ticket purchasing, or station information. Clicking this icon could lead to more details about train services or access to a train-related feature."
                }
            ]
        }
    ],
    "image_id": "d4e5596b-c166-4345-a7d0-947014505f0a"
}
\end{lstlisting}

\paragraph{Icon grounding data}~\par
In the icon grounding data, each data item includes an image containing multiple icons and a description of the target icon as input, and produces a grounding action (e.g., a click) in the Qwen2.5-VL tool-use format as output.

An example in JSON format is provided below. The image (Figure~\ref{fig:icon_ground}) contains several icons, and the description corresponds to one commonly used icon which represents “density” or “line spacing” in user interfaces.

\begin{lstlisting}[language=json,breaklines=true]
{
    "conversations": [
        {
            "from": "system",
            "value": {
                "text": "You are a helpful assistant.\n\n# Tools\n\nYou may call one or more functions to assist with the user query.\n\nYou are provided with function signatures within <tools></tools> XML tags:\n<tools>\n{\"type\": \"function\", \"function\": {\"name\": \"computer_use\", \"description\": \"Use a mouse and keyboard to interact with a computer, and take screenshots.\\n* This is an interface to a desktop GUI. You do not have access to a terminal or applications menu. You must click on desktop icons to start applications.\\n* Some applications may take time to start or process actions, so you may need to wait and take successive screenshots to see the results of your actions. E.g. if you click on Firefox and a window doesn't open, try wait and taking another screenshot.\\n* The screen's resolution is 1092x1932.\\n* Whenever you intend to move the cursor to click on an element like an icon, you should consult a screenshot to determine the coordinates of the element before moving the cursor.\\n* If you tried clicking on a program or link but it failed to load, even after waiting, try adjusting your cursor position so that the tip of the cursor visually falls on the element that you want to click.\\n* ..."
            }
        },
        {
            "from": "user",
            "value": [
                {
                    "image": "images_grounded/categorized_icons-
                    /tabler/tabler-icons/baseline-density-medium.png",
                    "width": 1920,
                    "height": 1080,
                    "resized_height": 1092,
                    "resized_width": 1932,
                    "seq_len": 2693
                },
                {
                    "text": "Click on the icon with functionality: This icon is typically used to represent a 'density' or 'line spacing' setting in user interfaces. It suggests adjusting the spacing between lines or items to a medium level. Clicking this icon might change the display settings to a medium density, balancing between compact and spacious layouts."
                }
            ]
        },
        {
            "from": "assistant",
            "value": [
                {
                    "text": "<tool_call>\n{\"name\": \"computer_use\", \"arguments\": {\"action\": \"left_click\", \"coordinate\": [709, 471]}}\n</tool_call>"
                }
            ]
        }
    ],
    "image_id": "cb1cb4ba-61ab-4caf-8c8d-a57ba3a6c310"
}
\end{lstlisting}

\subsubsection{Component data}

\begin{figure}[htbp]
    \centering
    \begin{subfigure}{0.45\textwidth}
        \includegraphics[width=\linewidth]{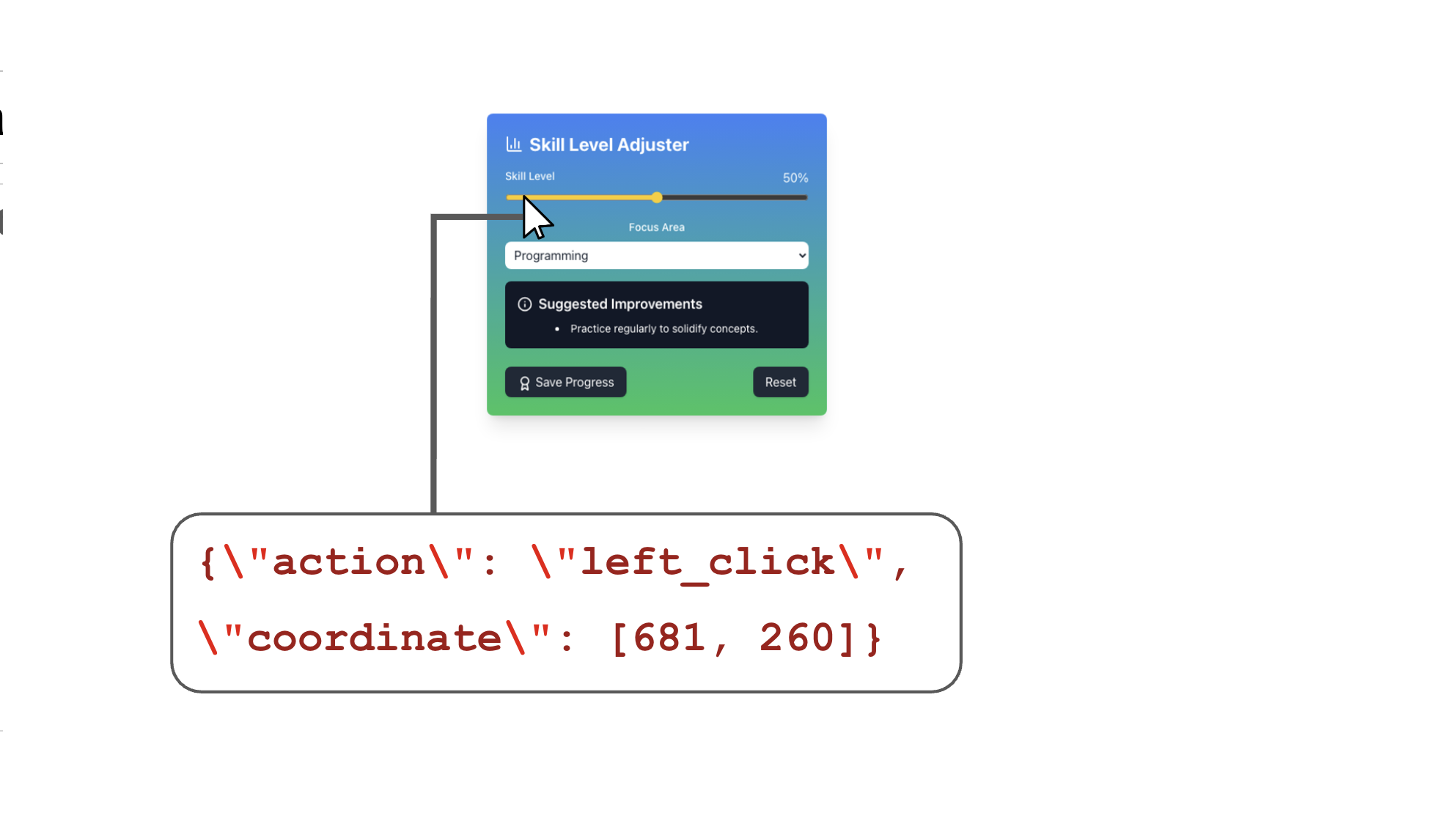}
        \caption{Example of rendered component grounding data}
        \label{fig:component_render}
    \end{subfigure}
    \hfill
    \begin{subfigure}{0.45\textwidth}
        \includegraphics[width=\linewidth]{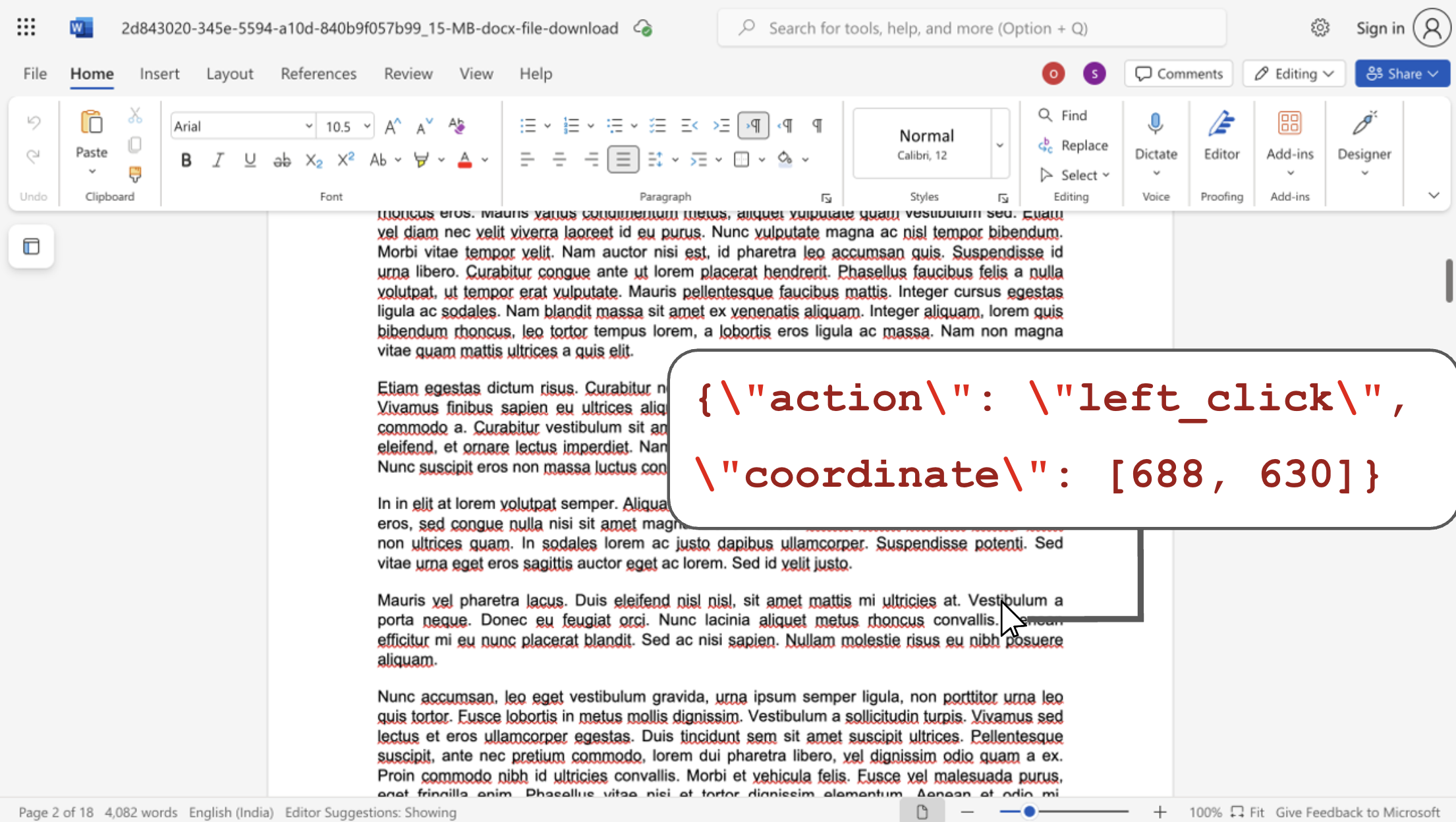}
        \caption{Example of doc grounding data}
        \label{fig:component_doc}
    \end{subfigure}
    \vskip\baselineskip
    \begin{subfigure}{0.45\textwidth}
        \includegraphics[width=\linewidth]{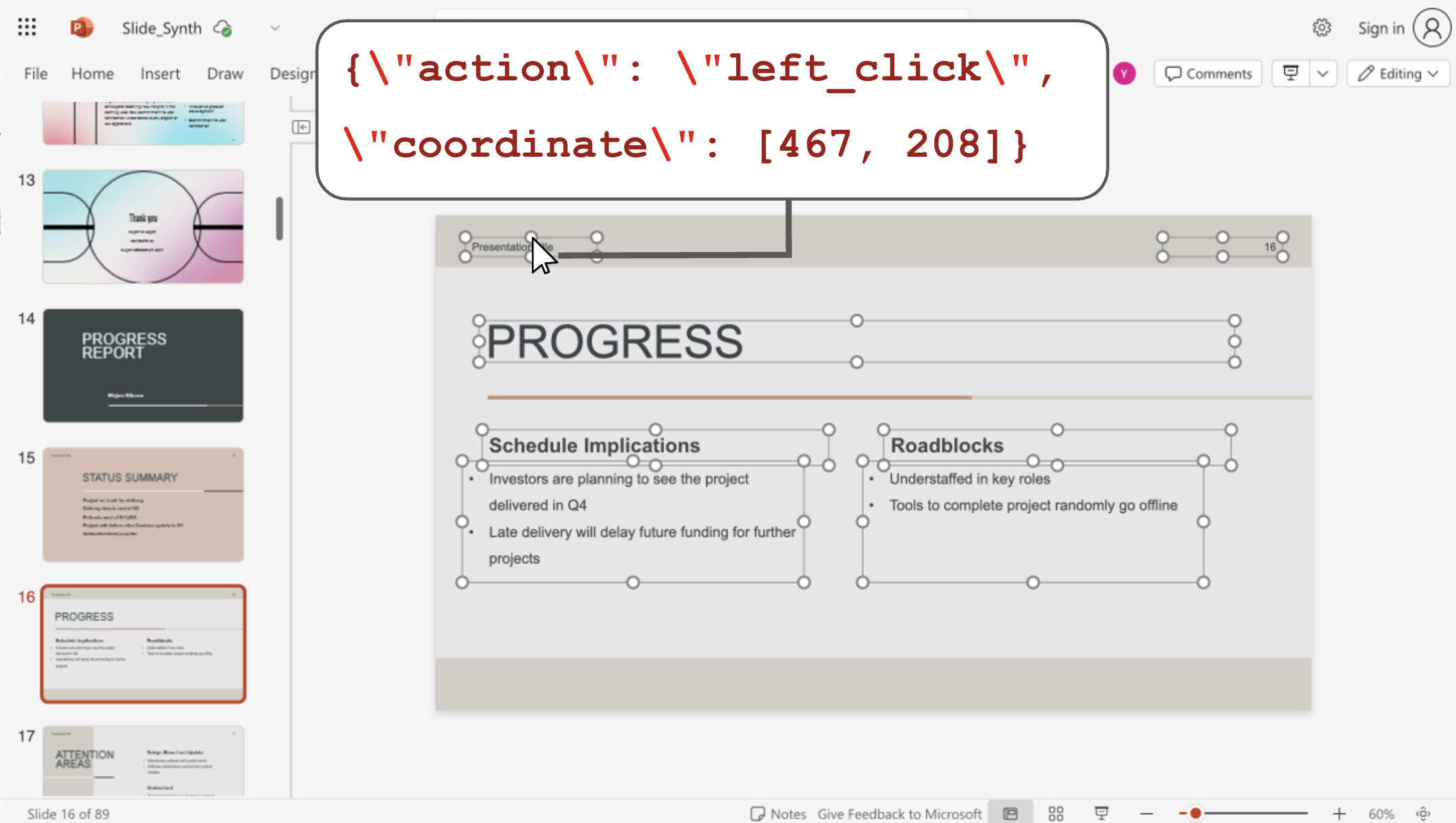}
        \caption{Example of slide grounding data}
        \label{fig:component_slide}
    \end{subfigure}
    \hfill
    \begin{subfigure}{0.45\textwidth}
        \includegraphics[width=\linewidth]{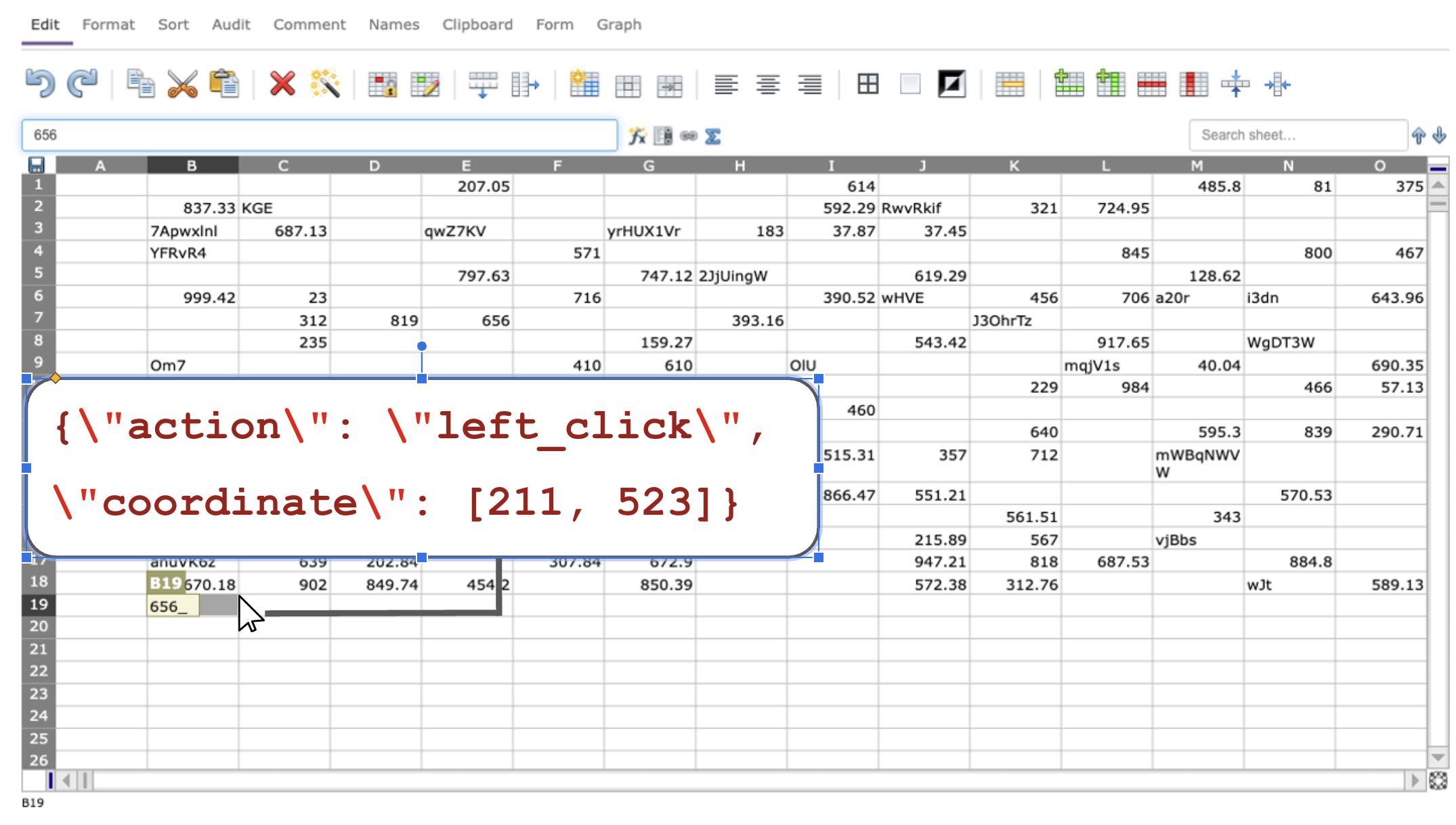}
        \caption{Example of sheet grounding data}
        \label{fig:component_sheet}
    \end{subfigure}
    \caption{Examples of component data}
    \label{fig:component_data}
\end{figure}

\paragraph{Component data from Code-and-rendering pipeline}~\par

In the component data from the Code-and-Rendering pipeline, each data item includes an image containing a rendered UI component and a corresponding user instruction as input, and produces a grounding action (e.g., a click) in the Qwen2.5-VL tool-use format as output.

An example in JSON format is shown below. The associated image (Figure~\ref{fig:component_render}) displays a slider component, and the instruction reads "...Read the text label displaying 'UI/UX Design Basics' located in the lower third of the catalog interface, which is part of a book listing..."

\begin{lstlisting}[language=json,breaklines=true]
{
    "conversations": [
        {
            "from": "system",
            "value": {
                "text": "You are a helpful assistant.\n\n# Tools\n\nYou may call one or more functions to assist with the user query.\n\nYou are provided with function signatures within <tools></tools> XML tags:\n<tools>\n{\"type\": \"function\", \"function\": {\"name\": \"computer_use\", \"description\": \"Use a mouse and keyboard to interact with a computer, and take screenshots.\\n* This is an interface to a desktop GUI. You do not have access to a terminal or applications menu. You must click on desktop icons to start applications.\\n* Some applications may take time to start or process actions, so you may need to wait and take successive screenshots to see the results of your actions. E.g. if you click on Firefox and a window doesn't open, try wait and taking another screenshot.\\n* The screen's resolution is 1092x1932.\\n* Whenever you intend to move the cursor to click on an element like an icon, you should consult a screenshot to determine the coordinates of the element before moving the cursor.\\n* If you tried clicking on a program or link but it failed to load, even after waiting, try adjusting your cursor position so that the tip of the cursor visually falls on the element that you want to click..."
            }
        },
        {
            "from": "user",
            "value": [
                {
                    "image": "data/chakra/slider/other_screenshot/-
                    original/SkillLevelAdjuster_1742005837.475969.png",
                    "width": 1920,
                    "height": 1080,
                    "resized_height": 1092,
                    "resized_width": 1932,
                    "seq_len": 2693
                },
                {
                    "text": "Please generate the next move according to the UI screenshot and instruction.\n\nInstruction: Set the skill level to 4% on the slider."
                }
            ]
        },
        {
            "from": "assistant",
            "value": [
                {
                    "text": "<tool_call>\n{\"name\": \"computer_use\", \"arguments\": {\"action\": \"left_click\", \"coordinate\": [681, 260]}}\n</tool_call>"
                }
            ]
        }
    ],
    "image_id": "568b8930-ec6a-4574-9b75-a18ed2c87cc0"
}
\end{lstlisting}

\paragraph{Component data for real-world augmentation}~\par

In the component data for real-world augmentation, each data item includes an image containing a real-world screenshots from an existing website or application and a corresponding user instruction as input, and produces a grounding action (e.g., a click) in the Qwen2.5-VL tool-use format as output.

We provide one example for each of the three data sources: \textit{doc}, \textit{slide}, and \textit{sheet}. The user instructions for these examples are as follows:

\begin{itemize}
    \item \textbf{Doc:} \textit{Given the following text: ". Vestibulum a", find this text in the document and click the space between the consecutive characters "t" and "i".}
    \item \textbf{Slide:} \textit{Please generate the next move according to the UI screenshot and instruction. Instruction: Select the handle located at the top of the text box that contains the text "Presentation title."}
    \item \textbf{Sheet:} \textit{Navigate to the top-left corner of cell C19.}
\end{itemize}

Corresponding examples in JSON format are shown below. The associated UI screenshots are provided in Figures~\ref{fig:component_doc}, \ref{fig:component_slide}, and \ref{fig:component_sheet}, respectively.

Example of doc grounding data:
\begin{lstlisting}[language=json,breaklines=true]
{
    "conversations": [
        {
            "from": "system",
            "value": {
                "text": "You are a helpful assistant.\n\n# Tools\n\nYou may call one or more functions to assist with the user query.\n\nYou are provided with function signatures within <tools></tools> XML tags:\n<tools>\n{\"type\": \"function\", \"function\": {\"name\": \"computer_use\", \"description\": \"Use a mouse and keyboard to interact with a computer, and take screenshots.\\n* This is an interface to a desktop GUI. You do not have access to a terminal or applications menu. You must click on desktop icons to start applications.\\n* Some applications may take time to start or process actions, so you may need to wait and take successive screenshots to see the results of your actions. E.g. if you click on Firefox and a window doesn't open, try wait and taking another screenshot.\\n* The screen's resolution is 728x1288.\\n* Whenever you intend to move the cursor to click on an element like an icon, you should consult a screenshot to determine the coordinates of the element before moving the cursor.\\n* If you tried clicking on a program or link but it failed to load, even after waiting, try adjusting your cursor position so that the tip of the cursor visually falls on the element that you want to click..."
            }
        },
        {
            "from": "user",
            "value": [
                {
                    "image": "AmHHgw-Nep9dv1S3X9n5gaoKsDxY_1280x720_SPACE_563_5.png",
                    "width": 1280,
                    "height": 720,
                    "resized_height": 728,
                    "resized_width": 1288,
                    "seq_len": 1198
                },
                {
                    "text": "Given the following text:\n\". Vestibulum a \"\n, find the text in the document and click the space between the continuous character \"t\" and \"i\" in the text."
                }
            ]
        },
        {
            "from": "assistant",
            "value": [
                {
                    "text": "<tool_call>\n{\"name\": \"computer_use\", \"arguments\": {\"action\": \"left_click\", \"coordinate\": [688, 630]}}\n</tool_call>"
                }
            ]
        }
    ],
    "image_id": "3d35b0b2-d541-45ea-be23-b668263b5b69"
}
\end{lstlisting}

Example of slide grounding data:
\begin{lstlisting}[language=json,breaklines=true]
{
    "conversations": [
        {
            "from": "system",
            "value": {
                "text": "You are a helpful assistant.\n\n# Tools\n\nYou may call one or more functions to assist with the user query.\n\nYou are provided with function signatures within <tools></tools> XML tags:\n<tools>\n{\"type\": \"function\", \"function\": {\"name\": \"computer_use\", \"description\": \"Use a mouse and keyboard to interact with a computer, and take screenshots.\\n* This is an interface to a desktop GUI. You do not have access to a terminal or applications menu. You must click on desktop icons to start applications.\\n* Some applications may take time to start or process actions, so you may need to wait and take successive screenshots to see the results of your actions. E.g. if you click on Firefox and a window doesn't open, try wait and taking another screenshot.\\n* The screen's resolution is 728x1288.\\n* Whenever you intend to move the cursor to click on an element like an icon, you should consult a screenshot to determine the coordinates of the element before moving the cursor.\\n* If you tried clicking on a program or link but it failed to load, even after waiting, try adjusting your cursor position so that the tip of the cursor visually falls on the element that you want to click..."
            }
        },
        {
            "from": "user",
            "value": [
                {
                    "image": "slides_1280*720/slide_15/original.png",
                    "width": 1280,
                    "height": 720,
                    "resized_height": 728,
                    "resized_width": 1288,
                    "seq_len": 1198
                },
                {
                    "text": "Please generate the next move according to the UI screenshot and instruction.\n\nInstruction: Select the handle located at the top of the text box that contains the text \"Presentation title.\""
                }
            ]
        },
        {
            "from": "assistant",
            "value": [
                {
                    "text": "<tool_call>\n{\"name\": \"computer_use\", \"arguments\": {\"action\": \"left_click\", \"coordinate\": [467, 208]}}\n</tool_call>"
                }
            ]
        }
    ],
    "image_id": "3f2ebbae-dee3-4fea-bbc0-ab93136bedab"
}
\end{lstlisting}

Example of sheet grounding data:
\begin{lstlisting}[language=json,breaklines=true]
{
    "conversations": [
        {
            "from": "system",
            "value": {
                "text": "You are a helpful assistant.\n\n# Tools\n\nYou may call one or more functions to assist with the user query.\n\nYou are provided with function signatures within <tools></tools> XML tags:\n<tools>\n{\"type\": \"function\", \"function\": {\"name\": \"computer_use\", \"description\": \"Use a mouse and keyboard to interact with a computer, and take screenshots.\\n* This is an interface to a desktop GUI. You do not have access to a terminal or applications menu. You must click on desktop icons to start applications.\\n* Some applications may take time to start or process actions, so you may need to wait and take successive screenshots to see the results of your actions. E.g. if you click on Firefox and a window doesn't open, try wait and taking another screenshot.\\n* The screen's resolution is 728x1288.\\n* Whenever you intend to move the cursor to click on an element like an icon, you should consult a screenshot to determine the coordinates of the element before moving the cursor.\\n* If you tried clicking on a program or link but it failed to load, even after waiting, try adjusting your cursor position so that the tip of the cursor visually falls on the element that you want to click..."
            }
        },
        {
            "from": "user",
            "value": [
                {
                    "image": "cell_left_top_corner/images/1280x720_cell_C19.png",
                    "width": 1280,
                    "height": 720,
                    "resized_height": 728,
                    "resized_width": 1288,
                    "seq_len": 1198
                },
                {
                    "text": "Navigate to top left corner of C19"
                }
            ]
        },
        {
            "from": "assistant",
            "value": [
                {
                    "text": "<tool_call>\n{\"name\": \"computer_use\", \"arguments\": {\"action\": \"mouse_move\", \"coordinate\": [211, 523]}}\n</tool_call>"
                }
            ]
        }
    ],
    "image_id": "15987cec-74d4-4624-b539-43a114caac8f"
}
\end{lstlisting}

\subsubsection{Layout data}

\begin{figure}[htbp]
    \centering
    \begin{subfigure}{0.45\textwidth}
        \includegraphics[width=\linewidth]{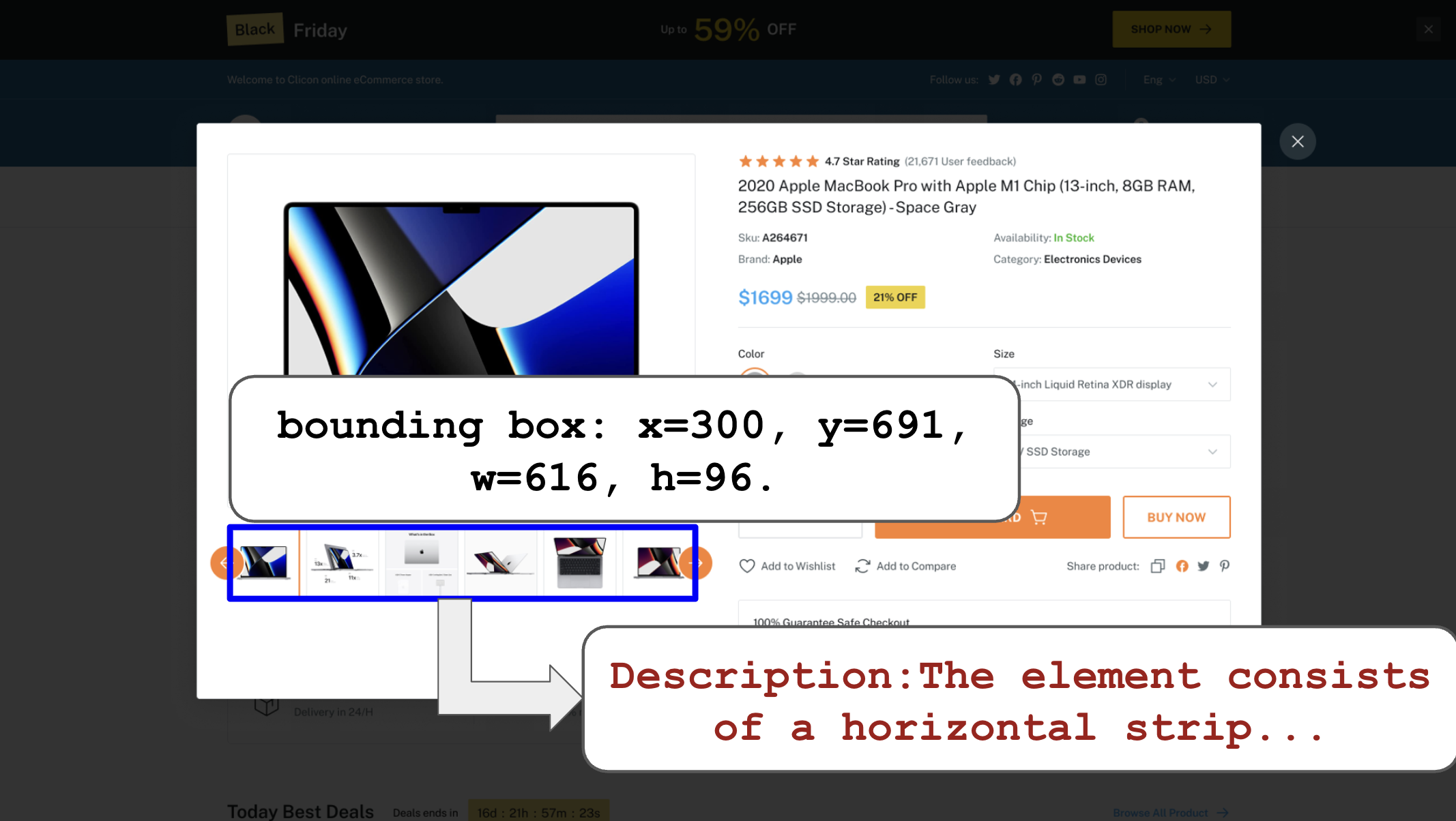}
        \caption{Example of icon description data}
        \label{fig:layout_desc}
    \end{subfigure}
    \hfill
    \begin{subfigure}{0.45\textwidth}
        \includegraphics[width=\linewidth]{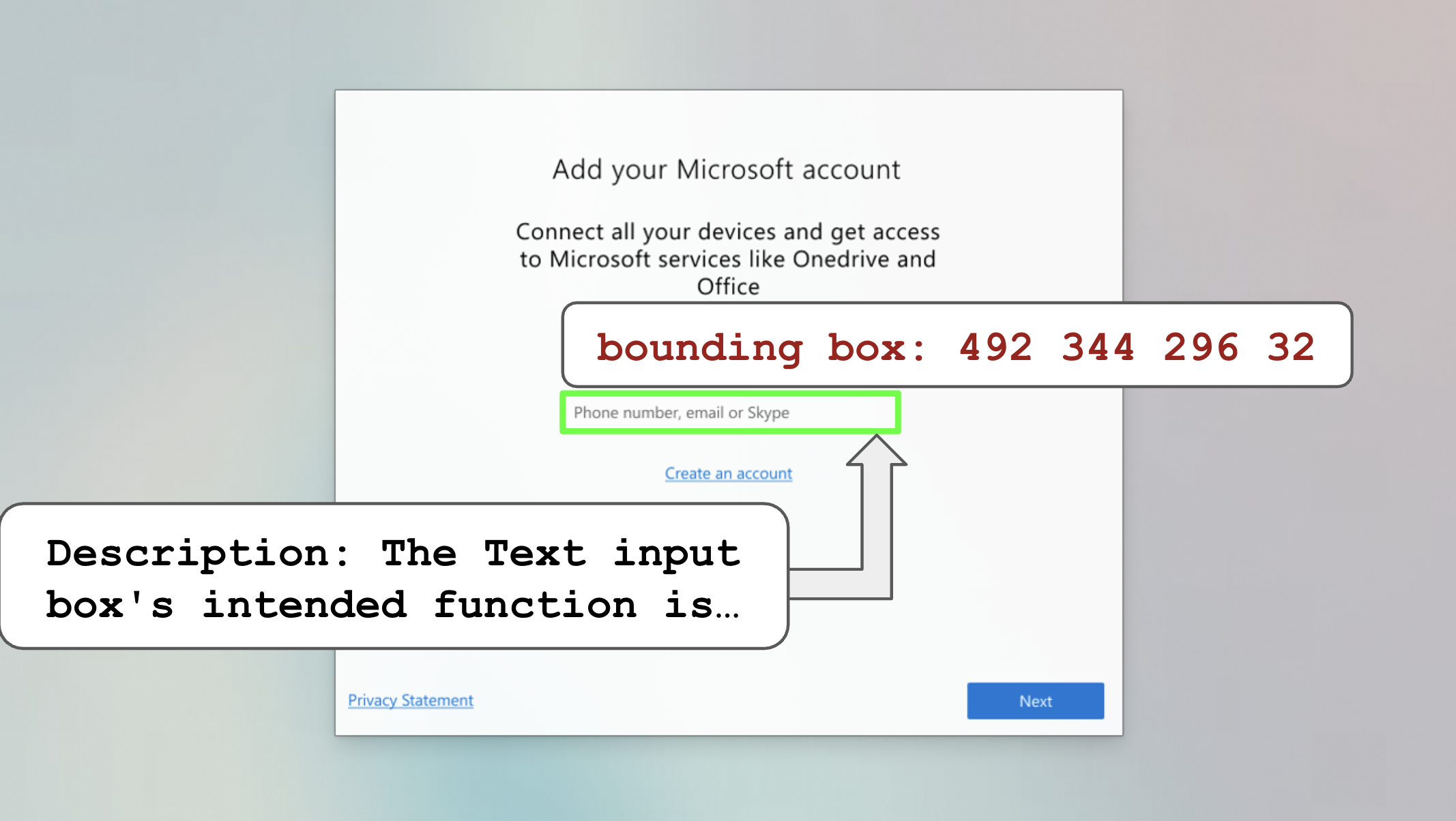}
        \caption{Example of icon grounding data}
        \label{fig:layout_ground}
    \end{subfigure}
    \caption{Examples of component grounding data}
    \label{fig:layout_data}
\end{figure}

\paragraph{Layout description data}~\par
In the layout description data, each item includes a screenshot and the bounding box coordinates of a target element as input, and generates a description covering the element’s visual composition, spatial context, user interaction, and type.

An example in JSON format is provided below. The associated image (Figure~\ref{fig:layout_desc}) shows a product interface from a shopping website, where the target element is an image carousel featuring thumbnail views of a laptop.

\begin{lstlisting}[language=json]
{
    "conversations": [
        {
            "from": "system",
            "value": {
                "text": "Working as a UI navigation guide, you will analyze a screenshot and bounding box coordinates. Your objective is to provide a detailed, user-friendly description of the UI element within the specified region, helping users identify it without relying on coordinates. Address these key points:\n\n1. Visual Composition: Detail the element's visual aspects, including its geometry, color scheme, and styling\n2. Spatial Context: Describe the element's position in relation to surrounding interface components\n3. User Interaction: Explain the element's functionality and how users should interact with it\n4. Element Type: Specify the category of UI component (such as progress bar, tooltip)\n\nMaintain precision and clarity in your descriptions, avoiding vague references. Emphasize distinctive characteristics and contextual relationships that make the element easily identifiable."
            }
        },
        {
            "from": "user",
            "value": [
                {
                    "image": "ce0ff960..."(Truncated due to layout constraints),
                    "width": 1920,
                    "height": 1080,
                    "resized_height": 1092,
                    "resized_width": 1932,
                    "seq_len": 2693
                },
                {
                    "text": "bounding box: x=300, y=691, w=616, h=96. Generate pls."
                }
            ]
        },
        {
            "from": "assistant",
            "value": [
                {
                    "text": "## Visual Composition: The element consists of a horizontal strip of thumbnail images, each depicting different views of a laptop. The thumbnails are bordered by a thin line and are evenly spaced. On either side of the strip, there are orange circular buttons with white arrows, indicating navigation options to scroll through more images. The thumbnails include images of the laptop from various angles and a box content view.\n\n## Spatial Context: The element is located centrally below the main product image within a pop-up product detail view. It is positioned above the product description and purchase options, acting as a secondary image gallery for the product.\n\n## User Interaction: The primary function of this element is to allow users to preview different images of the product. Users can click on the thumbnails to view a larger version of the image in the main display area above. The navigation arrows enable scrolling through additional images.\n\n## Element Type: Image carousel with navigation buttons."
                }
            ]
        }
    ],
    "image_id": "e54778c0-4d45-4997-a382-33244ced2513"
}
\end{lstlisting}

\paragraph{Layout grounding data}~\par
In the layout grounding data, each data item includes a screenshot and a corresponding user description of an element as input, and produces a bounding box of corresponding element as output.

An example in JSON format is shown below. The associated image (Figure~\ref{fig:layout_ground}) displays the layout of adding a Microsoft account, and the description refers to the text input box that allows users to input their phone number, email, or Skype ID.

\begin{lstlisting}[language=json]
{
    "conversations": [
        {
            "from": "system",
            "value": {
                "text": "Operating as a GUI element localization agent, your primary function is to process screenshots and element descriptions to identify the exact position of interface components. You work with coordinates (x, y, width, height)."
            }
        },
        {
            "from": "user",
            "value": [
                {
                    "image": "06c6786b..."(Truncated due to layout constraints),
                    "width": 1280,
                    "height": 720,
                    "resized_height": 728,
                    "resized_width": 1288,
                    "seq_len": 1198
                },
                {
                    "text": "The Text input box's intended function:\nThe primary function of this element is to allow users to input their phone number, email, or Skype ID as part of the Microsoft account sign-in process. Users are expected to click or tap on the box to activate it and then type their credentials."
                }
            ]
        },
        {
            "from": "assistant",
            "value": [
                {
                    "text": "492 344 296 32"
                }
            ]
        }
    ],
    "image_id": "db53ba9b-7416-4a01-bb19-e2a66d3a666d"
}
\end{lstlisting}

For demonstration purposes, we only show one round of dialogue in each "conversations" entry. In practice, to speed up training and save storage space, multiple rounds of dialogue related to a single screenshot are consolidated into a single JSON object's "conversations" list.

\subsection{Additional Case Studies}
In this section, we explain more logic behind the models' actions and analyze the factors that contribute to their improvements or failures.

\subsubsection{Cases of improvement}\label{appendix:case_success}

\paragraph{Case ~\ref{fig:appendix_case_1}} Leveraging training on icon, component, and layout data, \ours places greater emphasis on keywords like \texttt{button} and \texttt{on the left}. In contrast, Qwen concentrates more on literal text matching, resulting in clicking on \texttt{Sheet1}. 

\paragraph{Case ~\ref{fig:appendix_case_2}} \ours effectively identifies the correct icon with the specified function from numerous elements on the screen, showcasing its deep understanding of common icon functionalities. In contrast, traditional models often struggle to learn the association between icons and their functions when trained with coarse-grained data.

\paragraph{Case ~\ref{fig:appendix_case_3}} To execute this example correctly, models must thoroughly understand both the specific component (what constitutes a \texttt{horizontal scroll bar}) and the overall layout (where the scroll bar is located). The Qwen model, however, interacted with an unrelated element.

\paragraph{Case ~\ref{fig:appendix_case_4}} We found that the base model, which has not been trained on components and layouts, may not accurately manage subpages such as pop-ups and message bars. In contrast, \ours successfully identifies clickable text links.

\paragraph{Case ~\ref{fig:appendix_case_5}} This task involves having the model click on a specific mathematical symbol. Although the Qwen model demonstrates strong mathematical skills, these abilities do not improve its GUI grounding capability without fine-tuning on decomposed GUI data.

\paragraph{Case ~\ref{fig:appendix_case_6}} The GUI for this task includes a variety of elements and complex functions. However, \ours successfully identified the area relevant to mode switching through precise text matching.

\input{images/appendix_case_study}

\subsubsection{Cases of failure}\label{appendix:case_failure}

\input{images/appendix_case_failure}

\ours still faces challenges in certain situations. We present a selection of representative examples, with solutions to these challenges reserved for future work.

\paragraph{Case ~\ref{fig:row1_col1}} The task required the model to click the right-pointing arrow to close the right panel. Instead, \ours clicked the \texttt{`x`} button. Although this action was functionally correct, it did not strictly follow the requirement to \texttt{click the right arrow}. Therefore, we classify this as a failure case.

\paragraph{Case ~\ref{fig:row1_col2}} The interface for this task shows the word \texttt{center} twice. One is displayed in white, indicating it is active, while the other is in gray, indicating it is inactive. \ours misunderstood the color difference and mistakenly clicked the lower, inactive \texttt{center} option.

\paragraph{Case ~\ref{fig:row2_col1}} In this task, \ours was instructed to accept the cookie options. However, it mistakenly focused on the \texttt{cookie preferences} option. We believe the wording in the instruction misled the model.

\paragraph{Case ~\ref{fig:row2_col2}} This task required clicking on a letter in an artistic font, highlighting the model's current limitations in handling grounding tasks involving artistic or stylized designs. This shortcoming may arise from the lack of art and design data in the current training corpus.

\subsection{Agentic Benchmark Results of \ours}\label{appendix:detail_bmk_result}

\input{tables/appendix_osworld}

\input{tables/appendix_waa}

%% file: tables/appendix_og_types_full.tex
\begin{table}[htbp]
\centering
\caption{Full table of distribution of examples in the \ourbenchmarkname benchmark categorized by GUI grounding capabilities and their corresponding interface element types.}
\vspace{5px}
\scalebox{0.8}{
\footnotesize
\begin{tabular}{p{3.5cm}p{10cm}c}
\toprule
\textbf{Capabilities} & \textbf{Element Types} & \textbf{\# of Examples} \\
\midrule
Text Matching & Label & 268 \\
\addlinespace
Element Recognition & Icon, Image, Button & 337 \\
\addlinespace
Layout Understanding & Tab, Banner/Notification, Accordion/Collapsible Panel, Pagination Control, Toolbar, Menu Bar, Dropdown Menu, List, Grid, Tree View, Dialog/Modal, Panel/Container, Sidebar, Drawer & 252 \\
\addlinespace
Fine-grained Manipulation & Slider, Stepper, Divider, Toggle/Switch, Accordion/Collapsible Panel, Checkbox, Radio Button, Color Picker, Date Picker, Table, Text Field/Input Box, Search Bar, Text Filed, Input Box & 154 \\
\addlinespace
Refusal & -- & 54 \\
\bottomrule
\end{tabular}
}
\label{tab:app-capabilities-classification}
\end{table}

%% file: tables/bmk_comparison.tex
\begin{table}[htbp]
\centering
\caption{Comparison between \ourbenchmarkname and previous benchmarks.}
\scalebox{0.62}{
\begin{tabular}{p{2.2cm}p{2cm}cp{4cm}ccccc}
\toprule
\textbf{Benchmarks} & \textbf{Platforms} & \textbf{\# of Examples} & \textbf{\# of Annotated UI-Types} & \multicolumn{2}{c}{\textbf{Instruction Annotation}} & \textbf{Fine-grained Actions} & \textbf{Refusal Cases} \\
\cmidrule(lr){5-6}
 & & & & \textbf{Visual} & \textbf{Functional} & & \\
\midrule
ScreenSpot-v2 & Mobile, Desktop, Web & 1272 & 2 (Icon, Text) & \redcross & \greencheck & \redcross & \redcross \\
ScreenSpot Pro & Desktop & 1581 & 2 (Icon, Text) & \redcross & \greencheck & \redcross & \redcross \\
OmniAct & Desktop, Web & 9802 & 3 (Icon, Text, Color) & \redcross & \greencheck & \redcross & \redcross \\
\midrule[\heavyrulewidth]
\rowcolor{gray!10} \textbf{\ourbenchmarkname} & \textbf{Desktop} & \textbf{\numexamples} & \textbf{32} & \textbf{\greencheck} & \textbf{\greencheck} & \textbf{\greencheck} & \textbf{\greencheck} \\
\bottomrule
\end{tabular}
}
\label{tab:benchmark_comparison}
\end{table}

%% file: tables/data_statistics.tex
{\small
\begin{longtable}{lrrrc}
    \caption{Data statistics of our dataset. 
    The `\# Line' indicates the compression of multiple query-answer pairs to improve training efficiency.} \label{tab:data_stats} \\
\textbf{Data Source} & \textbf{\# Image} & \textbf{\# Line} & \textbf{\# Turn} & \textbf{Sampling} \\
\hline
\endfirsthead

\multicolumn{5}{c}%
{{\bfseries \tablename\ \thetable{} -- continued from previous page}} \\
\hline
\textbf{Data Source} & \textbf{\# Image} & \textbf{\# Line} & \textbf{\# Turn} & \textbf{Sampling} \\
\hline
\endhead

\multicolumn{5}{r}{{\textit{Continued on next page}}} \\ 
\endfoot

\hline
\endlastfoot
\\

\multicolumn{5}{l}{\textbf{\ours}} \\
Icon Captioning & 267,102 & 403,584 & 251,837 & All \\
Icon Grounding & 202,399 & 202,419 & 202,419 & All \\
Component Manipulation (Rule-based) & 29,303 & 40,653 & 40,653 & All \\
Component Manipulation (Generated) & 60,085 & 529,749 & 1,192,687 & All \\
Layout Captioning (App) & 5,117 & 17,721 & 366,774 & All \\
Layout Grounding (App) & 5,117 & 25,133 & 916,539 & All \\
Layout Captioning (OS) & 2,901 & 14,351 & 258,334 & All \\
Layout Grounding (OS) & 2,901 & 26,190 & 774,546 & All \\
\hline \\

\multicolumn{5}{l}{\textbf{\ours Refusal}} \\
Refusal Data (Various Sources) & 165,235 & 2,666,124 & 2,666,124 & Random:5\% \\
\hline \\

\multicolumn{5}{l}{\textbf{\textsc{Aguvis++}~\cite{xu2024aguvis}}} \\
SeeClick~\cite{cheng2024seeclickharnessingguigrounding} & 66,426 & 69,634 & 525,442 & All \\
WebUI~\cite{wu2023webui} & 57,389 & 57,389 & 143,187 & All \\
GUIEnv~\cite{chen2024guicourse} & 70,394 & 327,972 & 327,972 & All \\
GUIAct (web single)~\cite{chen2024guicourse} & 17,545 & 17,572 & 17,572 & All \\
Widget Captioning~\cite{li2020widget} & 14,409 & 101,426 & 101,426 & All \\
RicoSCA~\cite{li2020mapping} & 18,146 & 173,212 & 173,212 & All \\
UI RefExp~\cite{bai2021uibert} & 4,646 & 15,624 & 15,624 & All \\
RICO Icon~\cite{deka2017rico} & 16,133 & 16,133 & 32,091 & All \\
OmniACT~\cite{Kapoor2024OmniACTAD} & 6,720 & 6,720 & 6,720 & All \\
DocVQA Grounding~\cite{Mathew2020DocVQAAD, xu2024aguvis} & 9,756 & 34,060 & 34,060 & All \\
MM-Mind2Web~\cite{deng2023mind2web} & 7,351 & 7,351 & 7,351 & All \\
GUIAct (web multi)~\cite{chen2024guicourse} & 13,262 & 65,740 & 65,740 & All \\
AitZ~\cite{zhang2024android} & 12,002 & 11,914 & 11,914 & All \\
AndroidControl~\cite{Li2024OnTE} & 54,678 & 54,678 & 54,678 & All \\
Guide~\cite{superagi2023guide} & 12,422 & 12,422 & 12,422 & All \\
OS-Atlas~\cite{Wu2024OSATLASAF} & 303,472 & 303,472 & 303,472 & All \\
\hline \\

\textbf{In-house Data} \\
Additional In-house Annotated and Augmented Data & 1,392,009 & 1,392,016 & 1,486,289 & All \\
\hline

\end{longtable}
}

%% file: tables/icon_data_srcs.tex
{\small
\begin{longtable}{p{0.25\textwidth}p{0.2\textwidth}p{0.45\textwidth}}
\caption{Data sources for icon collection in the \ours dataset. Due to the diverse and scattered nature of these sources, they are presented here collectively rather than being broken down in the overall dataset overview.} 
\label{tab:icon_data_srcs} \\
\textbf{Data} & \textbf{Source} & \textbf{Link} \\
\hline
\endfirsthead

\multicolumn{3}{c}%
{{\bfseries \tablename\ \thetable{} -- continued from previous page}} \\
\hline
\textbf{Data} & \textbf{Source} & \textbf{Link} \\
\hline
\endhead

\multicolumn{3}{r}{{\textit{Continued on next page}}} \\ 
\endfoot

\hline
\endlastfoot

Ubuntu 2204 & Crawl & \url{https://github.com/ubuntu/yaru/tree/master/icons} \\
Snap Store & Crawl & \url{https://snapcraft.io/store} \\
Windows XP & Reverse engineering & - \\
Windows Vista & Reverse engineering & - \\
Windows 7 & Reverse engineering & - \\
\multirow{2}{*}{Windows 10} & Reverse engineering & - \\
& Crawl & \url{https://learn.microsoft.com/en-us/windows/apps/design/style/segoe-ui-symbol-font} \\
\multirow{2}{*}{Windows 11} & Reverse engineering & - \\
& Crawl & \url{https://github.com/microsoft/fluentui-system-icons/tree/main/assets} \\
Miscrosoft App Store & Crawl & \url{https://apps.microsoft.com/apps?hl=en-gb&gl=US} \\
macOS Ventura  & Reverse engineering & -  \\
macOS Sonoma & Reverse engineering & -  \\
macOS Sequoia & Reverse engineering & - \\
macOS icon Collection & Crawl & \url{https://macosicons.com/} \\
Apple App store & Crawl & \url{}  \\
iOS App store & Crawl & \url{}  \\
Calculator & Crawl & \url{https://github.com/microsoft/calculator/tree/main/src/Calculator/Assets} \\
Audacity & Crawl & \url{https://github.com/audacity/audacity/tree/master/au3/libraries/lib-theme-resources} \\
Google & Crawl & \url{https://fonts.google.com/icons}  \\
VSCode & Crawl & \url{https://github.com/microsoft/vscode-icons}  \\
LibreOffice & Crawl & \url{https://github.com/LibreOffice/core/tree/master/icon-themes}  \\
GitHub & Crawl & \url{https://github.com/primer/octicons/tree/main/icons}  \\
GIMP & Crawl & \url{https://github.com/GNOME/gimp}  \\
VLC & Crawl & \url{https://github.com/videolan/vlc}  \\
PhotoShop & Reverse engineering & -  \\

\hline
\end{longtable}
}

%% file: tables/component_data_srcs.tex
\small
\begin{longtable}{p{0.40\textwidth}p{0.25\textwidth}p{0.25\textwidth}}
\caption{Statistics of Material UI Components} 
\label{tab:component_data_srcs} \\
\textbf{Component Type} & \textbf{Conversations} & \textbf{Images} \\
\hline
\endfirsthead

\multicolumn{3}{c}%
{{\bfseries \tablename\ \thetable{} -- continued from previous page}} \\
\hline
\textbf{Component Type} & \textbf{Conversations} & \textbf{Images} \\
\hline
\endhead

\multicolumn{3}{r}{{\textit{Continued on next page}}} \\ 
\endfoot

\hline
\endlastfoot

material (Total) & 385,493 & 31,309 \\
\quad no-\-ssr & 321 & 24 \\
\quad box & 560 & 47 \\
\quad textarea-\-autosize & 445 & 37 \\
\quad click-\-away-\-listener & 764 & 45 \\
\quad links & 886 & 35 \\
\quad floating-\-action-\-button & 689 & 51 \\
\quad bottom-\-navigation & 6,709 & 535 \\
\quad popper & 3,258 & 169 \\
\quad modal & 1,699 & 71 \\
\quad speed-\-dial & 3,974 & 630 \\
\quad accordion & 1,840 & 82 \\
\quad rating & 9,409 & 869 \\
\quad use-\-media-\-query & 7,285 & 113 \\
\quad dividers & 2,318 & 83 \\
\quad skeleton & 2,103 & 85 \\
\quad alert & 7,290 & 1,378 \\
\quad typography & 511 & 38 \\
\quad button-\-group & 2,474 & 102 \\
\quad radio-\-buttons & 3,020 & 115 \\
\quad steppers & 5,252 & 869 \\
\quad container & 625 & 37 \\
\quad badges & 2,991 & 108 \\
\quad cards & 3,881 & 160 \\
\quad progress & 4,448 & 231 \\
\quad icons & 4,663 & 173 \\
\quad image-\-list & 2,389 & 96 \\
\quad popover & 658 & 41 \\
\quad toggle-\-button & 7,919 & 1,183 \\
\quad checkboxes & 8,447 & 1,148 \\
\quad buttons & 4,545 & 206 \\
\quad selects & 5,122 & 194 \\
\quad backdrop & 214 & 16 \\
\quad menus & 15,498 & 1,839 \\
\quad transitions & 1,794 & 92 \\
\quad masonry & 7,932 & 106 \\
\quad text-\-fields & 3,964 & 285 \\
\quad portal & 134 & 26 \\
\quad dialogs & 9,478 & 1,445 \\
\quad breadcrumbs & 3,693 & 110 \\
\quad switches & 7,050 & 1,050 \\
\quad stack & 2,371 & 82 \\
\quad paper & 5,993 & 97 \\
\quad tooltips & 5,648 & 266 \\
\quad timeline & 7,893 & 219 \\
\quad chips & 13,440 & 1,951 \\
\quad transfer-\-list & 2,100 & 295 \\
\quad tabs & 52,425 & 2,917 \\
\quad snackbars & 6,891 & 1,477 \\
\quad app-\-bar & 17,474 & 2,096 \\
\quad table & 11,536 & 839 \\
\quad lists & 17,377 & 2,094 \\
\quad drawers & 15,942 & 1,846 \\
\quad grid-\-legacy & 5,979 & 149 \\
\quad pagination & 12,497 & 197 \\
\quad slider & 27,843 & 2,210 \\
\quad autocomplete & 10,356 & 322 \\
\quad avatars & 5,634 & 154 \\
\quad grid & 7,842 & 174 \\
\hline
mantine (Total) & 27,814 & 762 \\
\quad InputValidation & 577 & 14 \\
\quad DndTable & 118 & 6 \\
\quad ButtonProgress & 45 & 6 \\
\quad ActionToggle & 852 & 17 \\
\quad HeaderMenu & 62 & 3 \\
\quad AutocompleteLoading & 17 & 3 \\
\quad AuthenticationImage & 34 & 3 \\
\quad NavbarMinimalColored & 97 & 4 \\
\quad PasswordStrength & 98 & 4 \\
\quad HeaderTabs & 75 & 3 \\
\quad NavbarLinksGroup & 5,322 & 56 \\
\quad ArticleCard & 185 & 7 \\
\quad HeroBullets & 111 & 4 \\
\quad InputWithButton & 73 & 4 \\
\quad FeaturesGrid & 121 & 4 \\
\quad CardsCarousel & 79 & 4 \\
\quad UsersRolesTable & 94 & 4 \\
\quad ContainedInputs & 89 & 6 \\
\quad FeaturesImages & 140 & 5 \\
\quad NavbarMinimal & 64 & 3 \\
\quad HeroImageBackground & 102 & 7 \\
\quad TableSelection & 245 & 6 \\
\quad CardGradient & 718 & 15 \\
\quad HeroContentLeft & 95 & 6 \\
\quad ButtonCopy & 52 & 5 \\
\quad FeaturesCards & 128 & 4 \\
\quad TableReviews & 140 & 3 \\
\quad UserCardImage & 202 & 7 \\
\quad StatsGrid & 214 & 7 \\
\quad NavbarSearch & 141 & 5 \\
\quad ArticlesCardsGrid & 144 & 6 \\
\quad ProgressCard & 60 & 4 \\
\quad NotFoundImage & 22 & 3 \\
\quad ProgressCardColored & 1,852 & 28 \\
\quad UserInfoAction & 138 & 8 \\
\quad ImageCheckboxes & 263 & 9 \\
\quad StatsCard & 111 & 5 \\
\quad ImageActionBanner & 60 & 4 \\
\quad HeaderSearch & 81 & 4 \\
\quad CustomSwitch & 32 & 3 \\
\quad FaqSimple & 97 & 4 \\
\quad HeaderSimple & 63 & 4 \\
\quad ForgotPasswordInput & 37 & 4 \\
\quad DndList & 167 & 6 \\
\quad ArticleCardFooter & 118 & 4 \\
\quad CarouselCard & 94 & 4 \\
\quad CommentSimple & 107 & 5 \\
\quad StatsGroup & 78 & 3 \\
\quad StatsControls & 124 & 5 \\
\quad DoubleHeader & 70 & 5 \\
\quad TableOfContentsFloating & 74 & 4 \\
\quad FaqWithImage & 71 & 4 \\
\quad CardWithStats & 250 & 8 \\
\quad EmailBanner & 146 & 6 \\
\quad LeadGrid & 145 & 7 \\
\quad Subgrid & 73 & 4 \\
\quad SliderIcon & 132 & 3 \\
\quad UserButton & 72 & 4 \\
\quad NavbarSegmented & 54 & 4 \\
\quad NavbarSimple & 103 & 4 \\
\quad NothingFoundBackground & 116 & 11 \\
\quad FeaturesTitle & 181 & 5 \\
\quad HeroImageRight & 132 & 5 \\
\quad UsersStack & 240 & 4 \\
\quad FooterLinks & 208 & 5 \\
\quad NotFoundTitle & 19 & 3 \\
\quad ContactUs & 398 & 12 \\
\quad ButtonMenu & 261 & 17 \\
\quad GradientSegmentedControl & 102 & 5 \\
\quad ArticleCardVertical & 99 & 7 \\
\quad NavbarSimpleColored & 99 & 4 \\
\quad CurrencyInput & 43 & 5 \\
\quad SliderLabel & 196 & 3 \\
\quad ArticleCardImage & 48 & 4 \\
\quad FeaturesAsymmetrical & 76 & 3 \\
\quad FooterSocial & 157 & 8 \\
\quad HeaderMegaMenu & 91 & 4 \\
\quad StatsRingCard & 74 & 5 \\
\quad TableSort & 108 & 4 \\
\quad AuthenticationTitle & 77 & 3 \\
\quad TableScrollArea & 92 & 3 \\
\quad CommentHtml & 147 & 7 \\
\quad AuthenticationForm & 195 & 15 \\
\quad GetInTouch & 305 & 8 \\
\quad HeroTitle & 57 & 3 \\
\quad DropzoneButton & 24 & 4 \\
\quad ServerOverload & 124 & 8 \\
\quad SliderMarks & 32 & 4 \\
\quad GetInTouchSimple & 109 & 4 \\
\quad SliderWhite & 64 & 4 \\
\quad StatsRing & 138 & 4 \\
\quad StatsSegments & 149 & 5 \\
\quad HeroText & 117 & 8 \\
\quad FloatingLabelInput & 19 & 4 \\
\quad CookiesBanner & 48 & 4 \\
\quad TaskCard & 1,383 & 19 \\
\quad ForgotPassword & 49 & 3 \\
\quad InputTooltip & 40 & 4 \\
\quad TableOfContents & 119 & 4 \\
\quad CheckboxCard & 15 & 4 \\
\quad ServerError & 35 & 5 \\
\quad FaqWithBg & 74 & 4 \\
\quad SplitButton & 76 & 3 \\
\quad LanguagePicker & 100 & 5 \\
\quad BadgeCard & 38 & 3 \\
\quad SwitchesCard & 1,339 & 16 \\
\quad FeaturesCard & 91 & 4 \\
\quad ImageCard & 115 & 8 \\
\quad DoubleNavbar & 110 & 4 \\
\quad FaqWithHeader & 151 & 8 \\
\quad UserMenu & 136 & 4 \\
\quad UserInfoIcons & 103 & 4 \\
\quad NavbarNested & 167 & 7 \\
\quad SliderInput & 113 & 4 \\
\quad StatsGridIcons & 186 & 5 \\
\quad FooterSimple & 137 & 4 \\
\quad UsersTable & 2,927 & 23 \\
\quad SocialButtons & 325 & 10 \\
\quad SliderHover & 84 & 4 \\
\quad FooterCentered & 58 & 2 \\
\quad DndListHandle & 76 & 2 \\
\quad ActionsGrid & 122 & 4 \\
\quad GridAsymmetrical & 172 & 2 \\
\hline
ant-design (Total) & 473,723 & 16,837 \\
\quad switch & 1,484 & 94 \\
\quad watermark & 1,849 & 83 \\
\quad skeleton & 1,913 & 99 \\
\quad divider & 2,038 & 98 \\
\quad tooltip & 3,525 & 194 \\
\quad rate & 5,492 & 135 \\
\quad auto-\-complete & 4,359 & 203 \\
\quad tour & 2,174 & 114 \\
\quad checkbox & 6,531 & 255 \\
\quad splitter & 4,878 & 254 \\
\quad time-\-picker & 6,642 & 276 \\
\quad collapse & 5,191 & 225 \\
\quad qr-\-code & 2,711 & 160 \\
\quad menu & 6,182 & 215 \\
\quad segmented & 6,088 & 238 \\
\quad flex & 3,386 & 113 \\
\quad notification & 6,090 & 208 \\
\quad alert & 6,124 & 199 \\
\quad list & 7,470 & 190 \\
\quad button & 8,060 & 329 \\
\quad timeline & 5,727 & 163 \\
\quad carousel & 1,997 & 120 \\
\quad modal & 9,508 & 379 \\
\quad drawer & 7,818 & 275 \\
\quad steps & 9,873 & 338 \\
\quad affix & 1,105 & 66 \\
\quad card & 7,331 & 345 \\
\quad progress & 9,485 & 274 \\
\quad mentions & 3,829 & 190 \\
\quad typography & 3,405 & 203 \\
\quad tree-\-select & 5,515 & 248 \\
\quad descriptions & 6,732 & 236 \\
\quad message & 3,626 & 156 \\
\quad transfer & 7,308 & 193 \\
\quad popover & 3,009 & 163 \\
\quad empty & 1,105 & 89 \\
\quad badge & 8,339 & 292 \\
\quad radio & 6,888 & 239 \\
\quad spin & 1,825 & 126 \\
\quad float-\-button & 3,457 & 215 \\
\quad image & 3,765 & 217 \\
\quad cascader & 6,992 & 372 \\
\quad popconfirm & 2,156 & 153 \\
\quad calendar & 10,389 & 141 \\
\quad form & 10,818 & 679 \\
\quad config-\-provider & 3,167 & 143 \\
\quad app & 663 & 35 \\
\quad statistic & 2,345 & 96 \\
\quad back-\-top & 454 & 29 \\
\quad breadcrumb & 3,964 & 130 \\
\quad input-\-number & 5,420 & 286 \\
\quad space & 6,426 & 240 \\
\quad avatar & 5,120 & 150 \\
\quad icon & 2,898 & 135 \\
\quad slider & 9,588 & 231 \\
\quad tabs & 36,465 & 609 \\
\quad upload & 3,270 & 373 \\
\quad anchor & 4,524 & 165 \\
\quad tag & 5,118 & 214 \\
\quad tree & 16,354 & 433 \\
\quad input & 7,012 & 478 \\
\quad select & 15,055 & 550 \\
\quad color-\-picker & 6,342 & 410 \\
\quad pagination & 9,673 & 221 \\
\quad layout & 6,604 & 188 \\
\quad dropdown & 9,035 & 301 \\
\quad grid & 13,957 & 241 \\
\quad date-\-picker & 20,445 & 739 \\
\quad table & 44,891 & 844 \\
\quad result & 744 & 42 \\
\hline
chakra (Total) & 330,074 & 11,784 \\
\quad mark & 288 & 24 \\
\quad loader & 488 & 35 \\
\quad bleed & 420 & 35 \\
\quad aspect & 1,045 & 74 \\
\quad center & 964 & 60 \\
\quad skeleton & 1,097 & 80 \\
\quad fieldset & 834 & 45 \\
\quad locale & 424 & 26 \\
\quad list & 1,758 & 64 \\
\quad theme & 787 & 41 \\
\quad separator & 1,751 & 83 \\
\quad editable & 1,684 & 111 \\
\quad code & 1,630 & 72 \\
\quad float & 1,669 & 83 \\
\quad visually & 1,725 & 50 \\
\quad box & 2,337 & 118 \\
\quad segmented & 2,405 & 110 \\
\quad spinner & 1,921 & 110 \\
\quad simple & 2,256 & 58 \\
\quad link & 1,910 & 92 \\
\quad for & 1,057 & 33 \\
\quad hover & 1,231 & 63 \\
\quad blockquote & 1,770 & 109 \\
\quad flex & 1,653 & 74 \\
\quad alert & 2,702 & 143 \\
\quad accordion & 3,855 & 160 \\
\quad steps & 3,763 & 168 \\
\quad timeline & 2,784 & 85 \\
\quad stat & 3,082 & 123 \\
\quad switch & 3,285 & 173 \\
\quad radiomark & 680 & 28 \\
\quad text & 1,250 & 69 \\
\quad highlight & 2,075 & 97 \\
\quad drawer & 13,309 & 272 \\
\quad menu & 5,395 & 193 \\
\quad tooltip & 11,407 & 448 \\
\quad toggle & 846 & 56 \\
\quad collapsible & 325 & 24 \\
\quad button & 5,358 & 252 \\
\quad container & 672 & 27 \\
\quad checkmark & 624 & 29 \\
\quad badge & 2,590 & 91 \\
\quad close & 446 & 34 \\
\quad show & 636 & 47 \\
\quad field & 1,850 & 124 \\
\quad card & 3,155 & 108 \\
\quad empty & 987 & 70 \\
\quad textarea & 2,550 & 183 \\
\quad action & 673 & 30 \\
\quad image & 1,173 & 83 \\
\quad password & 2,104 & 68 \\
\quad toaster & 9,295 & 231 \\
\quad rating & 9,917 & 195 \\
\quad pin & 2,384 & 147 \\
\quad qr & 2,394 & 141 \\
\quad status & 1,879 & 60 \\
\quad group & 1,453 & 66 \\
\quad popover & 5,154 & 200 \\
\quad file & 1,644 & 168 \\
\quad prose & 2,275 & 90 \\
\quad tabs & 23,088 & 397 \\
\quad native & 1,429 & 67 \\
\quad em & 215 & 23 \\
\quad kbd & 8,119 & 197 \\
\quad portal & 218 & 26 \\
\quad dialog & 5,663 & 220 \\
\quad select & 6,850 & 232 \\
\quad tag & 4,416 & 150 \\
\quad clipboard & 1,444 & 93 \\
\quad grid & 1,784 & 45 \\
\quad table & 10,317 & 210 \\
\quad heading & 1,719 & 86 \\
\quad presence & 1,450 & 77 \\
\quad stack & 2,259 & 86 \\
\quad breadcrumb & 3,537 & 116 \\
\quad radio & 13,935 & 496 \\
\quad progress & 11,892 & 359 \\
\quad format & 2,572 & 134 \\
\quad pagination & 16,881 & 254 \\
\quad icon & 3,051 & 139 \\
\quad checkbox & 13,930 & 464 \\
\quad input & 3,568 & 224 \\
\quad avatar & 11,467 & 403 \\
\quad wrap & 3,186 & 56 \\
\quad number & 3,941 & 206 \\
\quad slider & 14,841 & 302 \\
\quad color & 10,339 & 560 \\
\quad data & 888 & 29 \\
\hline
\end{longtable}

%% file: tables/appendix_layout_os_stats.tex
\begin{longtable}{p{0.5\textwidth}cc}
\caption{Statistics of OS Layout Data\label{tab:rollout_stats}} \\
\toprule
\textbf{Rollout Environments} & \textbf{Screenshots} & \textbf{Elements} \\
\midrule
\endfirsthead

\multicolumn{3}{c}{{\bfseries Table \thetable{} -- continued from previous page}} \\
\toprule
\textbf{Rollout Environments} & \textbf{Screenshots} & \textbf{Number of Elements} \\
\midrule
\endhead

\midrule
\multicolumn{3}{r}{{\textit{Continued on next page}}} \\
\endfoot

\bottomrule
\endlastfoot

OSWorld (Ubuntu) & 2000 & 183889 \\
WindowsAgentArena (Windows) & 903 & 74445 \\
\midrule
\textbf{Total} & \textbf{2903} & \textbf{258334} \\

\end{longtable}

%% file: tables/appendix_layout_figma_stats.tex
\begin{longtable}{p{0.7\textwidth}cc}
\caption{Statistics of Layout Data Collected from Figma Commnuity Design Templates\label{tab:figma_stats}} \\
\toprule
\textbf{Design Templates} & \textbf{Images} & \textbf{Elements} \\
\midrule
\endfirsthead

\multicolumn{3}{c}{{\bfseries Table \thetable{} -- countined from previous page}} \\
\toprule
\textbf{Design Templates} & \textbf{Images} & \textbf{Elements} \\
\midrule
\endhead

\midrule
\multicolumn{3}{r}{{\textit{Continued on next page}}} \\
\endfoot

\bottomrule
\endlastfoot

{[}Freebie{]}-Home-Rent-App-UI-Design-(Community) & 3 & 59 \\
(Variants)-macOS-Big-Sur-UI-Kit-for-Figma-(Community) & 10 & 685 \\
10-Real-Chat/Messaging-Pages---Facebook, Reddit, Snapchat-\&-more-(Community) & 10 & 1269 \\
10-Real-Dashboard-Pages---AirBnB, Basecamp, Github, \&-more-(Community) & 10 & 2214 \\
10-Real-Homepages---AirBnb, Github, and-more-(Community) & 10 & 2162 \\
10-Real-Notification-Pages---AirBnB, Dropbox, Notion, \&-more-(Community) & 10 & 1103 \\
10-Real-Pricing-Pages---Basecamp, Dribble, \&-more-(Community) & 10 & 3554 \\
10-Real-Search-Results-Pages---Github, Loom, Notion-\&-more-(Community) & 11 & 1714 \\
10-Real-Sign-Up-Pages---Calendly, Dribbble, \&-more-(Community) & 11 & 570 \\
10-Real-User-Settings-Pages---Calendly, Github, Behance, \&-more-(Community) & 13 & 1006 \\
11-Real-Sign-In-Pages---AirBnB, Calendly, \&-more-(Community) & 12 & 557 \\
20-Modals, Popups, Alerts-(Community) & 13 & 135 \\
AWS-Admin-Redesign-by-FluentUI-(Community) & 21 & 1456 \\
AWS-Amplify-UI-Kit-(Community) & 28 & 713 \\
AWS-Platform-(Community) & 5 & 690 \\
Ai-Design-Templates-(Community) & 10 & 1071 \\
Airbnb---Home, Search, and-Listing-Pages-(Community) & 5 & 905 \\
Airbnb-UI-Kit-(Community) & 10 & 49 \\
Amazon-UI-Design-(Community) & 18 & 3522 \\
Android-UI-Kit-(Community) & 28 & 2364 \\
App-Clips-(Community) & 4 & 77 \\
App-Store-Template---See-how-your-App-looks-like-in-App-Store-(Community) & 3 & 756 \\
Apple Design Resources - macOS (Community) & 29 & 1164 \\
Apple-Mail-(Community) & 1 & 283 \\
Apple-Mail-Design-(Community)-(Community) & 3 & 72 \\
Apple-Maps-iOS-(Community) & 7 & 351 \\
Apple-Messages-Templates-(Community) & 8 & 567 \\
Apple-Pay-(Community) & 18 & 442 \\
Apple-TV+-UI-Kit-(Community) & 16 & 1290 \\
Apple-Website-UI-2023-(apple.com)-(Community) & 7 & 2618 \\
Apple-Widgets-UI-Kit-(Community) & 78 & 264 \\
Apple-and-Google-Play-store-UI-(Community) & 6 & 130 \\
Apple-iCloud-Login-(Community) & 2 & 2 \\
Apps-Paywalls-and-Subscription-Screens-(Community) & 5 & 48 \\
Assets-Kit-UI-Mobile, Tablet-\&-Desktop-(Community) & 45 & 986 \\
Audiobooks-by-Booksbury-(Community) & 3 & 76 \\
Betting-Mobile-app-(Community) & 6 & 162 \\
Binance-Market-Trade-Dashboard-UI-Design-(Community) & 1 & 584 \\
Booking.com-Mobile-App-Redesign---UX/UI-Case-Study-(Community) & 4 & 85 \\
Budddy-Chatbot-Freebie-(Community) & 8 & 241 \\
CAPTCHA-UI-Kit-(Community) & 16 & 124 \\
CAR-RENTAL-WEBSITE-(RESOONSIVE-DESIGN)-(Community) & 3 & 60 \\
Calendar-Interactive-UI-Kit-(Community) & 6 & 578 \\
Call-Center-Desktop-App-(Community) & 4 & 80 \\
Car-Rent-Website-Design---Pickolab-Studio-(Community) & 10 & 1232 \\
Car-Rental-Mobile-App-(Community) & 3 & 83 \\
Casino-Web-Site-(Community) & 12 & 3831 \\
Chat-for-desktop/mobile-|-Free-to-use-(Community) & 3 & 169 \\
ChatGPT-UI-Kit, AI-Chat-(Community) & 2 & 182 \\
Cinema-4D-GUI-Redesign-(Community) & 2 & 250 \\
Clicon---eCommerce-Marketplace-Website-Figma-Template-(Community) & 48 & 7178 \\
Club-Website-Design-|-WEB-UI-(Community) & 4 & 66 \\
Code-block, Syntax-highlighting & 2 & 20 \\
Coding-Website---UI-Kit-(Community) & 11 & 416 \\
Coinbase-Clone---Website-Prices-Page-(Community) & 1 & 448 \\
Components-library---Light-\&-Dark-mode-(Community) & 13 & 256 \\
Concept-~-Mailbox-Design-(Community) & 1 & 78 \\
Coursera-UI-KIT-(Community) & 0 & 0 \\
Crypto-App-Ui-Kit-(Community) & 61 & 2430 \\
Customer-onboarding-designs-\&-components---by-Bento-(Community) & 10 & 1315 \\
Dark-UI-Elements, Dropdowns-\&-Calendar-(Community) & 4 & 254 \\
Dashboard---Online-Learning-Profile-(Community) & 3 & 201 \\
Dashboard-UI-Kit---Dashboard, Free-Admin-Dashboard-(Community) & 6 & 2224 \\
Data-table-design-components.-Free-UI-Kit-(Community) & 13 & 31059 \\
Dating-Mobile-App-(Community) & 42 & 722 \\
Delivery-App-Ui-Kit-(Community) & 54 & 2393 \\
Desktop-Messaging-App-Concept-(Community) & 1 & 26 \\
Deupload---Decentralized-Cloud-Storage-Landing-Pages-(Community) & 45 & 2188 \\
Discord-(Community) & 2 & 116 \\
Discord-Redesign-(Community) & 16 & 3236 \\
Discord-UI-Mockup-(Community) & 11 & 672 \\
Disney+-App-Redesign-(Community) & 2 & 6 \\
DocketHub-(Community) & 10 & 1121 \\
Doordash-FREE-UI-Kit---By-Marvilo-(Community) & 5 & 569 \\
Dota-2-UI-Redesign-(Community) & 12 & 2287 \\
Duolingo-Pages-Collection-by-DesignDrops.io-(Community) & 13 & 714 \\
Duolingo-Workflows---Onboarding, Learning-a-language, Upgrading, \&-Cancelling-(Community) & 145 & 4974 \\
E-Store---Mobile/web-(Community) & 15 & 1502 \\
E-Tutor---Learning-Management-System-(Community) & 69 & 8475 \\
E-commerce-UI---Figma-Ecommerce-UI-Kit-(Demo-Version)-(Community) & 160 & 6522 \\
E-commerce-Website-Template-(Freebie)-(Community) & 9 & 858 \\
Ebay-New-Design-Concept-(Community) & 1 & 34 \\
Ecommerce-Website-Design-(Community) & 1 & 71 \\
Element-UI-Kit-2.15.7-(Community) & 42 & 2078 \\
Elite---Food-Restaurant-\&-Coffee-Free-Figma-Template-(Community) & 16 & 769 \\
Email-Message-Modal-(Community) & 2 & 165 \\
Embed-Media-Components-(Community) & 6 & 38 \\
Eonify---Mobile-App-Authentication-Page-(Community) & 7 & 91 \\
FREE-Gmail-Mockup-2024-template!-(Community) & 4 & 98 \\
FREEBIES-Landingpage-LaslesVPN-(Community) & 1 & 28 \\
Facebook-Page-Mockup-(2022)-(Community) & 1 & 53 \\
Facebook-ReDesign-2023-(Community) & 7 & 449 \\
Fantastical-Calendar-(Community) & 1 & 655 \\
FigmaSharp-Toolkit: macOS-Big-Sur-2.0.0-(Community) & 5 & 573 \\
Finance-Market-Trading-Terminal-(Community) & 14 & 766 \\
Fitness-App-UI-Kit-for-Gym-Workout-App-Fitness-Tracker-Mobile-App-Gym-Fitness-Mobile-App-UI-Kit-(Community) & 87 & 1723 \\
Fiverr--UI-Redesigned---Freelance-Marketplace-Website-Design-(Community) & 3 & 681 \\
Flight-Booking-App-UI-Kits-(Community) & 20 & 249 \\
Food-Catering-Service-App-With-Landing-Page---Figma-Freebies-|-Doradesign-(Community) & 13 & 696 \\
Food-Delivery-Website-\&-App-Design-UI-Kit-(Community) & 17 & 37 \\
Food-delivery-app-Ui-kit-(Community) & 18 & 117 \\
FoodWagon-Food-Delivery-Landing-Template-by-ThemeWagon-(Community) & 1 & 304 \\
Forms--/--Desktop-\&-Mobile-(Community) & 12 & 341 \\
Forum-Concept-for-Alem.school-(Community) & 6 & 403 \\
Free-Fitness-App-Ui-Kit-(Community) & 48 & 841 \\
Free-Instagram-UI-Mockups-2023-(Community) & 12 & 364 \\
Free-Modal-Upload-Files-Kit-for-Web-and-Mobile---Include-4-modes-(Community) & 27 & 1150 \\
Free-Trading-UI-Kit-(Community) & 43 & 725 \\
Free-YouTube-Shorts-Mockups-(Community) & 18 & 634 \\
Free-YouTube-Video-Player-Mockups-(Community) & 6 & 506 \\
Freebies---Apps-Tracking-Truck-Cargo-Courier-Delivery-(Community) & 2 & 152 \\
Freebies---Scooter-Renting-App-(Community) & 4 & 77 \\
Full-Apple-Music-Classical-App-(Community) & 160 & 18706 \\
Full-E-Commerce-Website-UI-UX-Design-(Community) & 15 & 2318 \\
GitHub-UI-(Community) & 2 & 1101 \\
Github-UI---Free-UI-Kit-(Recreated)-(Community) & 18 & 6568 \\
Gmail-UI-Mobile-Design-Template-2024!-(Community) & 4 & 38 \\
Gmail-UI-Part-1: Inbox-(Community) & 4 & 1940 \\
Gmail-UI-Part-2: Reading-\&-Composing-Emails-(Community) & 9 & 2505 \\
Google-Anlytics-Dashobard-(Community) & 1 & 28 \\
Google-Calendar---Web-version-revamp-(Community) & 4 & 392 \\
Google-Chrome-Browser-UI-Kit-2025-(Community) & 14 & 695 \\
Google-Chrome-UI-Kit-2022-(Community) & 1 & 39 \\
Google-Drive-Reverse-Engineer-(Community) & 14 & 1620 \\
Google-Gemini---Built-with-Material-3-Design-Kit-(Community) & 4 & 646 \\
Google-Maps---Bus-ticket-booking-(Community) & 19 & 34 \\
Google-Maps-Parking-Prototype-Testing-(Community) & 5 & 68 \\
Google-Meet-UI-(Community) & 1 & 2 \\
Google-Scholar-re-designed-(Community) & 3 & 33 \\
Google-Search-Result-Page-(SERP)-(Community) & 2 & 167 \\
Google-Sheet---Template-(Unofficial)-(Community) & 23 & 3152 \\
Google-Sign-in-GIS---Google-Identity-Services-(Community) & 9 & 136 \\
Google-Translate-Redesign-(Community) & 9 & 140 \\
Google-Weather-App-Redesign-(Community) & 3 & 47 \\
Google-search-(Community) & 8 & 313 \\
Health-Fitness-Workout-App-(FREEBIE---Prototype)-(Community) & 8 & 332 \\
HealthRise-Health-Tech-Dashboard-(Community) & 4 & 4513 \\
Hero-Giveaway---Redesigns-(Community) & 7 & 1391 \\
Hotel-booking-website-UI-(Community) & 1 & 5 \\
Hoteliq---Booking-Hotel-App-Design-(Community) & 3 & 222 \\
IKEA-/-eCommerce-Concept-Design-(Community) & 4 & 26 \\
IMDb-Redesign-(Community) & 15 & 2293 \\
InTouch---Messaging-App-UI-Kit-(Community) & 4 & 87 \\
Instagram-UI-Screens-(Community) & 36 & 441 \\
IntelliJ-Platform-UI-Kit-(Community) & 48 & 4598 \\
Invoice/Payment-Components---Dipa-Inhouse-(Community) & 17 & 828 \\
Job-Finder-App-UI-Kit-(Community) & 1 & 43 \\
Job-Finder-Ui-App-Kit-(Community) & 83 & 85 \\
Jobpilot---Job-Portal-Figma-UI-Template-(Community) & 3 & 913 \\
LOGIFY---WEB-LOGIN-UI-KIT-(Community) & 40 & 273 \\
Leetcode-Homepage-(Community) & 1 & 1 \\
Lenskart-Redesigned---HiFi-Wireframes-(Community) & 5 & 85 \\
LinkedIn-Business-Page-Mockup-(2024)-(Community) & 1 & 111 \\
LinkedIn-Redesign-UI-Kit-(Community) & 8 & 223 \\
Linkedin-Page-Mockup-(2022)-(Community) & 1 & 30 \\
Linkedin-UI-Screens-(Community) & 28 & 948 \\
Liquipedia-Web-Redesign-(Community) & 4 & 1606 \\
Live-Score-UI-KIT-(FREEBIES)-(Community) & 12 & 349 \\
Login-\&-Register-Web-UI-Kit-(Freebie)-(Community) & 5 & 170 \\
Loom-UI---Free-UI-Kit-(Recreated)-(Community) & 28 & 5405 \\
MEDDICAL---Hospital-website-template-(Community) & 10 & 189 \\
MacOS-file-upload-\&-download-(Community) & 3 & 363 \\
Map-Navigation-Mobile-App-UI-Kit-Template-(Community) & 2 & 15 \\
Market-Stock-Exchange-(Community) & 6 & 601 \\
Medical-Clinic-Booking-(Doctor-Appointment)-App-UI-Concept-(Community) & 3 & 20 \\
Mercedes-Benz-App-(Community) & 8 & 9 \\
Messager-Dashboard-design.-(Community) & 9 & 582 \\
Metroway---Train-Ticket-booking-website-(Community) & 5 & 301 \\
Microsoft-365-UI-Kit-(Community) & 358 & 95741 \\
Microsoft-Excel-+-Word-2024-(Community) & 4 & 1733 \\
Mobile-Chat-Figma-UI-Kits-|-BRIX-Templates-(Community) & 70 & 10979 \\
Mobile-eCommerce-Clothing-Store-App-Design-(Community) & 6 & 322 \\
Modern-Profile-UI-Kit---Freebies-UI-(Community) & 4 & 6 \\
Money-transfer-Ui-App-Kit-(Community) & 55 & 236 \\
Movie-App-Redesigned-HULU-(Community) & 23 & 400 \\
Movie-Ticket-Booking-Application---Coursera-UX-Specialization-(Community) & 22 & 40 \\
Movie-Ticket-Booking-Apps-(Community) & 5 & 49 \\
MyCourses.io---Course-Website-|-Course-Online-|-Course-details-|-Course-landing-page-|-Untitled-UI-(Community) & 50 & 3731 \\
Native-Web-Components---Browser-Default's-UI-Kit-(Community) & 12 & 214 \\
Navigation-App-Design-(Waze-App-Redesign)-(Community) & 2 & 17 \\
Neomorphism-music-player-for-desktop-(Community) & 6 & 19 \\
Netflix-Home-Page-desktop-\&-TV-(Community) & 1 & 8 \\
Netflix-home-page---Mobile-\&-TV/Desktop-(Community) & 2 & 24 \\
News-\&-Blog-App-UI-Kit-By-Al-Ferdous-(Community) & 6 & 46 \\
News-Website-UI-and-Presentation-for-Opportunists-(Community) & 2 & 2 \\
Nike-UI---Free-UI-Kit-(Recreated)-(Community) & 18 & 2769 \\
Nowted-–-A-Note-taking-App-(Community) & 6 & 362 \\
Officevibe-UI---Free-UI-Kit-(Recreated)-(Community) & 18 & 3577 \\
On-Demand-Medicine-Delivery-App-(My-Orders-Flow)-(Community) & 9 & 379 \\
Onboarding-Appointment-booking-(Community) & 1 & 183 \\
Onest---Classified-Ads-Listing-Figma-Template-(Commnity) & 44 & 6139 \\
PDF-Viewer-(Community) & 1 & 18 \\
Papery---News-Magazine-Mobile-App-(Community) & 21 & 1145 \\
Parking-App-Design-UI-|-Figma-(Community) & 30 & 270 \\
Patterns \& Layouts UI Kit (Community) & 108 & 12275 \\
Payment-Page-(Desktop)-(Community) & 2 & 184 \\
Picto---Personal-Portfolio-Free-Template-(Community) & 1 & 202 \\
Pinterest-Redesign-(Community) & 5 & 363 \\
Pinterest-UI---Free-UI-Kit-(Recreated)-(Community) & 4 & 716 \\
Plant-App-Freebies-(Community) & 13 & 69 \\
Print-dialog–Firefox-macOS-(Community) & 2 & 72 \\
Quiz-Game-(Community) & 12 & 206 \\
QuizGrad-webapp-(Community) & 6 & 132 \\
Quora-Redesign-(Community) & 2 & 10 \\
REIS---Real-State-Listing-Figma-Template-(Community) & 3 & 572 \\
Real-Estate-App-UI-Kit-(Community) & 79 & 3255 \\
Recreating-Google-Drive-Using-Lexicon-(Community) & 2 & 1002 \\
Reddit-Design-System-(Community) & 16 & 762 \\
Reddit-Material-Design-Redesign-(Community) & 14 & 142 \\
Redesign---ChatGPT-(Community) & 1 & 53 \\
Registration-Form-for-a-Medical-Laboratory-|-Medical-Analyzes-(Community) & 5 & 502 \\
Restaurant-Booking-Uikit-(Community) & 20 & 122 \\
Roommates-Apartments-Booking-(Community) & 2 & 54 \\
Sass Plat form Layouts - Wireframe Kit (Community) & 11 & 1073 \\
Scheddo---Bookings-\&-Reservations-UI/UX---Freebie-(Community) & 7 & 255 \\
Shell-Template---Windows-11-(Community) & 87 & 6881 \\
Shopcart---Online-Ecommerce-website-(Community) & 1 & 67 \\
Shopery---Organic-eCommerce-Shop-Website-Figma-Template-(Community) & 37 & 6916 \\
Simple-Chat-Widget-for-Desktop-(Community) & 4 & 80 \\
Siri-\&-App-Shortcuts-(Community) & 57 & 5077 \\
Slack-Desktop-App-Clone-(Community) & 5 & 238 \\
Slack-UI---Desktop-(Community) & 1 & 118 \\
Snow-Dashboard-UI-Kit-(Community) & 6 & 2608 \\
Soccer-Score-App-(Community) & 5 & 328 \\
Social-Login-Auth-Modals-(Community) & 5 & 49 \\
Sportify---Sports-streaming-app-(Community) & 34 & 1832 \\
Spotify---Mobile-UI-Kit-(Community) & 22 & 468 \\
Spotify-Redesign-(Community) & 32 & 6253 \\
Spotify-UI---Free-UI-Kit-(Recreated)-(Community) & 10 & 1329 \\
Spotify-UI-Design-(Search/Artist-Profile)-(Community) & 2 & 580 \\
Starbucks-Redesign-Mobil-App-(Community) & 8 & 330 \\
Steam-Redesign-(Community) & 36 & 6855 \\
Stock-Trading-App---UI-Concept-(Community) & 6 & 302 \\
Stripe-Apps-UI-toolkit-(Community) & 35 & 2625 \\
Stripe-Connect-Embedded-Components--UI-Toolkit-(Community) & 63 & 6002 \\
Subscription-Paywall-Modal-(Community) & 1 & 56 \\
Table-Booking-Restaurant-Application-(Web-+-Mobile-+-Admin-Panels)-(Community) & 99 & 2524 \\
Table-UI-3.0-|-Variants-Update-(Community) & 1 & 223 \\
Tap-to-Pay-on-iPhone-(Community) & 2 & 20 \\
Tasky---Task-and-Time-Management-Dashboard-(Community) & 1 & 5 \\
Taxi-Booking-App-(Community) & 8 & 46 \\
Technical-Support-Applications-Page-(Community) & 4 & 266 \\
Telegram-Design-System-(Community) & 46 & 8535 \\
Terminal-app-UI-(Community) & 6 & 26 \\
Tesla-Mobile-App-Redesign-(Community) & 3 & 185 \\
The-Unofficial-Spotify-Design-System-(Community) & 5 & 854 \\
Ticketing-App-Freebies-(Community) & 8 & 549 \\
TikTok-UI-Screens-(Community) & 14 & 100 \\
Tinder-Mobile-App-(Community) & 23 & 192 \\
TipKit-(Community) & 8 & 67 \\
To-do-list-dashboard-(Freebie)-(Community) & 2 & 86 \\
ToDoHQ--Activity-management-website-design-(Community) & 18 & 329 \\
Todoist-Free-UI-Kit---By-Marvilo-(Community) & 8 & 868 \\
Todoist-for-macOS-app-concept-(Community) & 3 & 1196 \\
Tour-Guide---travel-agency/travel-booking-website-(Community) & 4 & 155 \\
Travel-\&-Hotel-Booking-Light-Mobile-App-(Community) & 4 & 171 \\
Trello-Concept-(Community) & 2 & 2 \\
Twitch-UI---Autolayout-Interface-(Community) & 1 & 182 \\
Twitch-UI---Free-UI-Kit-(Recreated)-(Community) & 3 & 1670 \\
Twitter-UI-Clone-Design-(Community) & 9 & 992 \\
Twitter-UI-Screens-(Community) & 22 & 315 \\
Twitter-desktop-pages-(feed, sigup, login, profile)-(Community) & 5 & 175 \\
UF-File-Manager-(Community) & 15 & 280 \\
UI-DESIGN-FOR-MOCK-INTERVIEW-PLATFORM-(Company-side)-(Community) & 50 & 528 \\
Uber-App-UI---Free-UI-Kit-(Recreated)-(Community) & 3 & 306 \\
Uber-Redesign-(Community) & 3 & 19 \\
Ubuntu-Shiro-(Community) & 10 & 606 \\
VPN-App---UI-Kit-(Community) & 9 & 200 \\
Video-Player-For-Web-\&-Mobile-(Community) & 9 & 186 \\
Video-Streaming-Website---Responsive-web-app-prototype-(Community) & 3 & 50 \\
Visual-Studio-Code-Toolkit-(Community) & 45 & 8887 \\
Wallet-(Community) & 12 & 230 \\
WeChat-(Community) & 11 & 318 \\
WeUI-kit(Wechat)-(Community) & 33 & 391 \\
Web-Browser-Mockups-(Community) & 4 & 60 \\
Web-Dashboard-UI---Task-\&-Project-Management-(Community) & 1 & 7 \\
Website-FAQ-Accordions-Figma-Template-|-BRIX-Templates-(Community) & 3 & 3 \\
Website-Wireframes-UI-Kit-|-BRIX-Templates-(Community) & 108 & 1800 \\
WhatsApp-Pay-\&-Split-(Community) & 24 & 294 \\
Wikipedia-(Community) & 44 & 11773 \\
Windows-11-Chat-UI-Kit-(Community) & 15 & 857 \\
Windows-File-Explorer-—-Ego's-Take-(Community) & 18 & 3384 \\
Windows-Install-Redesigned-(Concept)-(Community) & 47 & 22947 \\
Windows-Outlook-Template-(Community) & 17 & 2269 \\
WordPress-Design-System-(Community) & 12 & 5289 \\
YouTube-Music-App-Redesign: Elevating-the-Music-Experience-(Community) & 3 & 50 \\
YouTube-Redesign-(Community) & 34 & 9176 \\
YouTube-UI-Clone-Design-(Community) & 10 & 1801 \\
Zoom-Apps-UI-Overview-(Community) & 19 & 4507 \\
aeroSpeed-Bus-Booking-Application-UI-Kit-[User-+-Driver]-(Community) & 14 & 330 \\
chat-app-UI-kit-(Community) & 5 & 7 \\
eDex---Online-Course-E-Learning-Website-(Comunity) & 2 & 447 \\
iBank---Banking-\&-E-Money-Management-App-|-FinPay-|-Digital-|-Finance-Mobile-Banking-App-Ui-Kit-(Community) & 89 & 2027 \\
iMessage-Apps-and-Stickers-(Community) & 22 & 1761 \\
iOS-17-Apple-music-Now-Playing-interface-(Community) & 4 & 4 \\
iOS-18-and-iPadOS-18-(Community) & 91 & 7171 \\
lark-(Community) & 28 & 45169 \\
macOS-Big-Sur-UI-Kit-(Community) & 20 & 3214 \\
macOS-Browser-UI-Kit-(Big-Sur-Update)-(Community) & 6 & 51 \\
telegram-app-(Community) & 15 & 297 \\
ui---Design-System-(Community) & 17 & 343 \\
\midrule
\textbf{Total} & \textbf{5273} & \textbf{563721} \\
\end{longtable}

%% file: tables/office_data.tex
\begin{table}[htbp]
    \centering
    \caption{Actions and Associated Components in Office Software}
    \scalebox{0.9}{
    \begin{tabular}{p{3.5cm}p{4cm}c}
    \toprule
    \textbf{Office Software} & \textbf{Action Type} & \textbf{Associated Component} \\
    \midrule
    Doc & Scroll & Scrollbar \\
    \addlinespace
    Doc & Select & NormalTextRun \\
    \addlinespace
    Doc & Click & NormalTextRun \\
    \addlinespace
    Slide & Drag & Text Box \\
    \addlinespace
    Slide & Click & Text Box \\
    \addlinespace
    Slide & Click & Slide Thumbnail \\
    \addlinespace
    Sheet & Click & Cell \\
    Sheet & Click & Edge \\
    Sheet & Click & Cell Corner \\
    Sheet & Click & Column Header \\
    Sheet & Click & Row Header \\
    \bottomrule
    \end{tabular}
    }
    \label{tab:office_actions}
\end{table}

%% file: images/appendix_case_study.tex
\begin{figure}[H]
    \centering

    \begin{subfigure}[t]{0.48\textwidth}
        \centering
        \includegraphics[width=\linewidth,height=0.2\textheight,keepaspectratio]{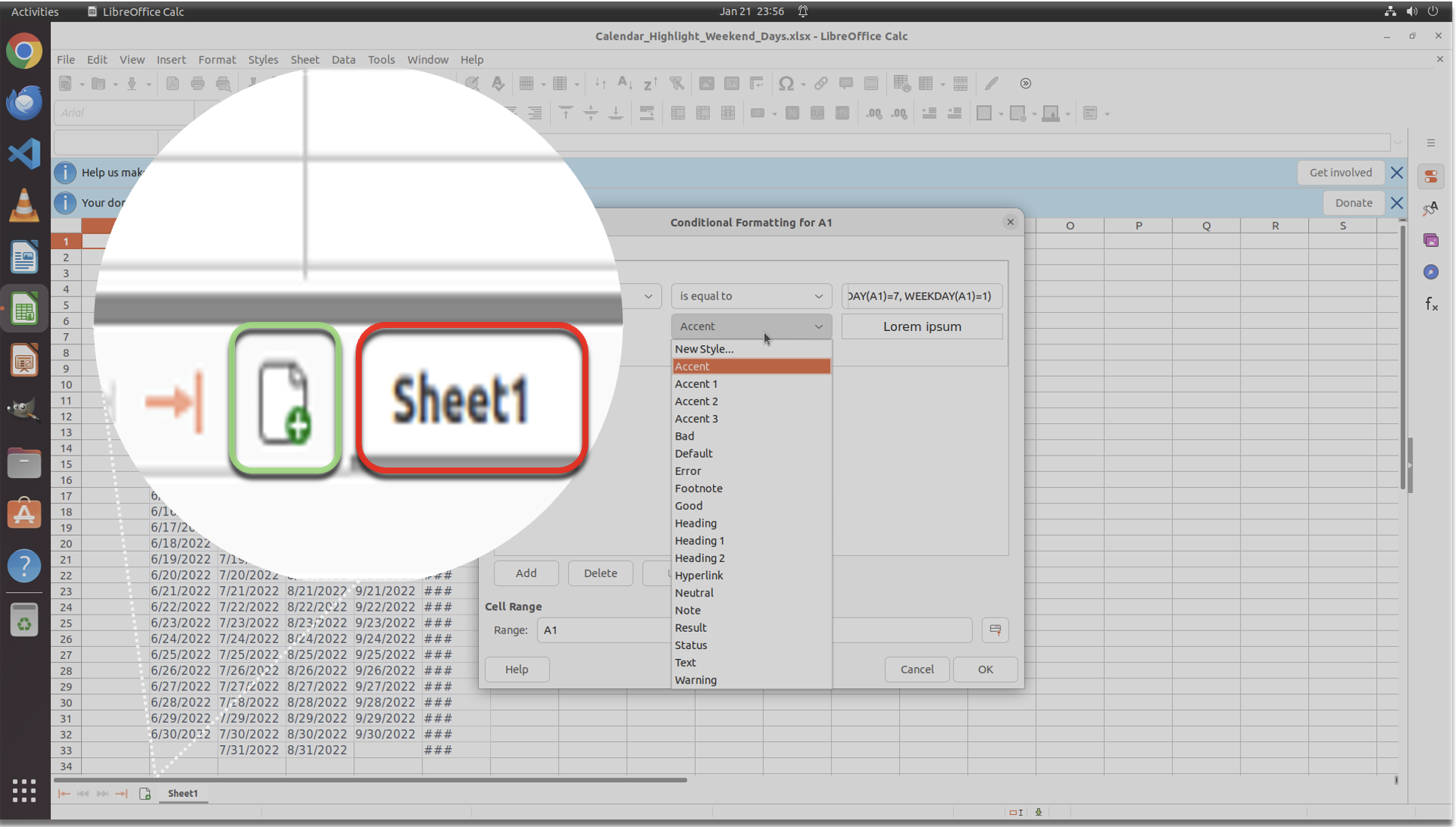}
        \subcaption{Instruction: Add a sheet by clicking the button on the left of "Sheet1".}
        \label{fig:appendix_case_1}
    \end{subfigure}
    \hfill
    \begin{subfigure}[t]{0.48\textwidth}
        \centering
        \includegraphics[width=\linewidth,height=0.2\textheight,keepaspectratio]{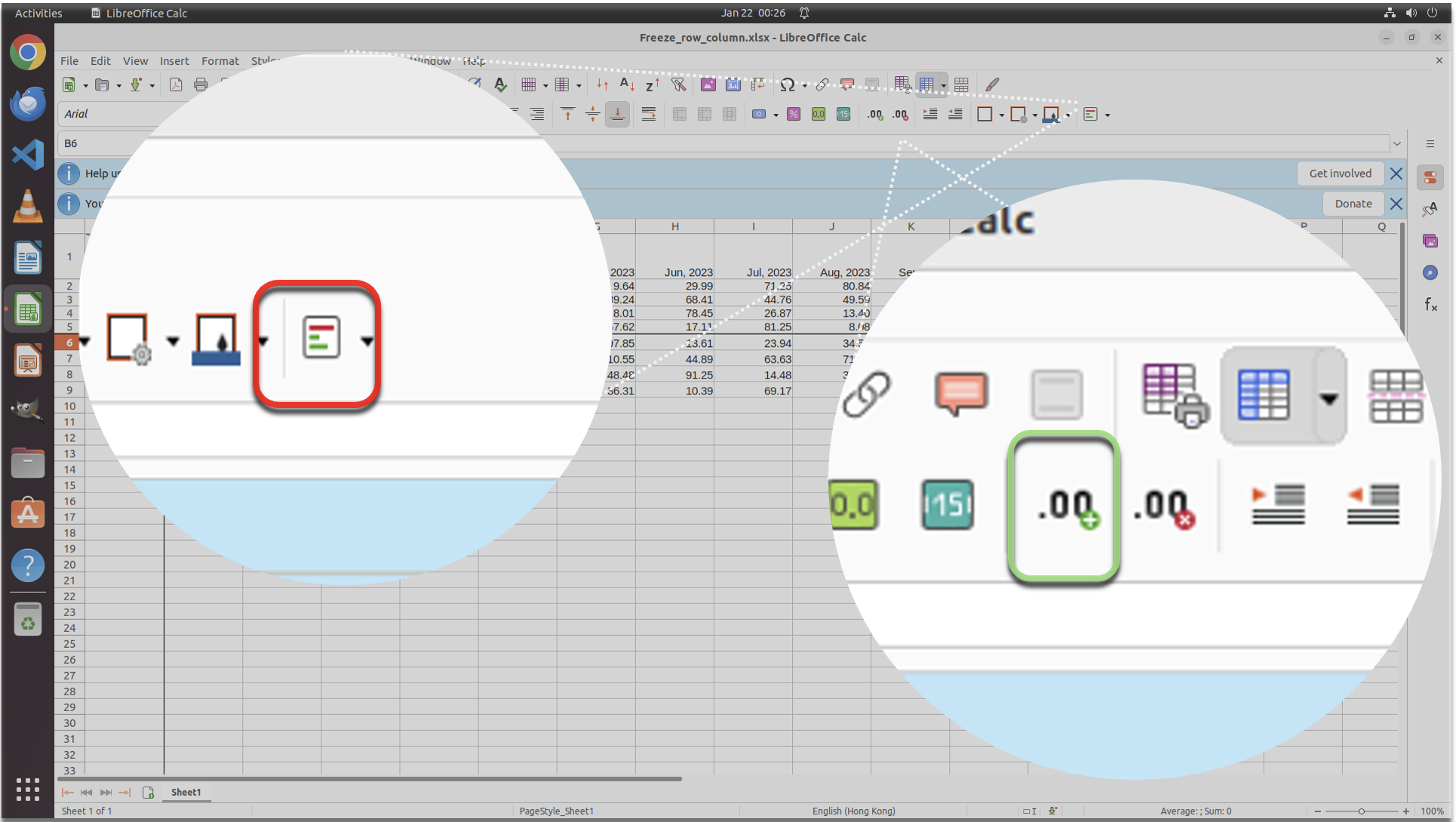}
        \subcaption{Instruction: Add Decimal Place for the current cell.}
        \label{fig:appendix_case_2}
    \end{subfigure}

    \vspace{0.5\baselineskip} 

    \begin{subfigure}[t]{0.48\textwidth}
        \centering
        \includegraphics[width=\linewidth,height=0.2\textheight,keepaspectratio]{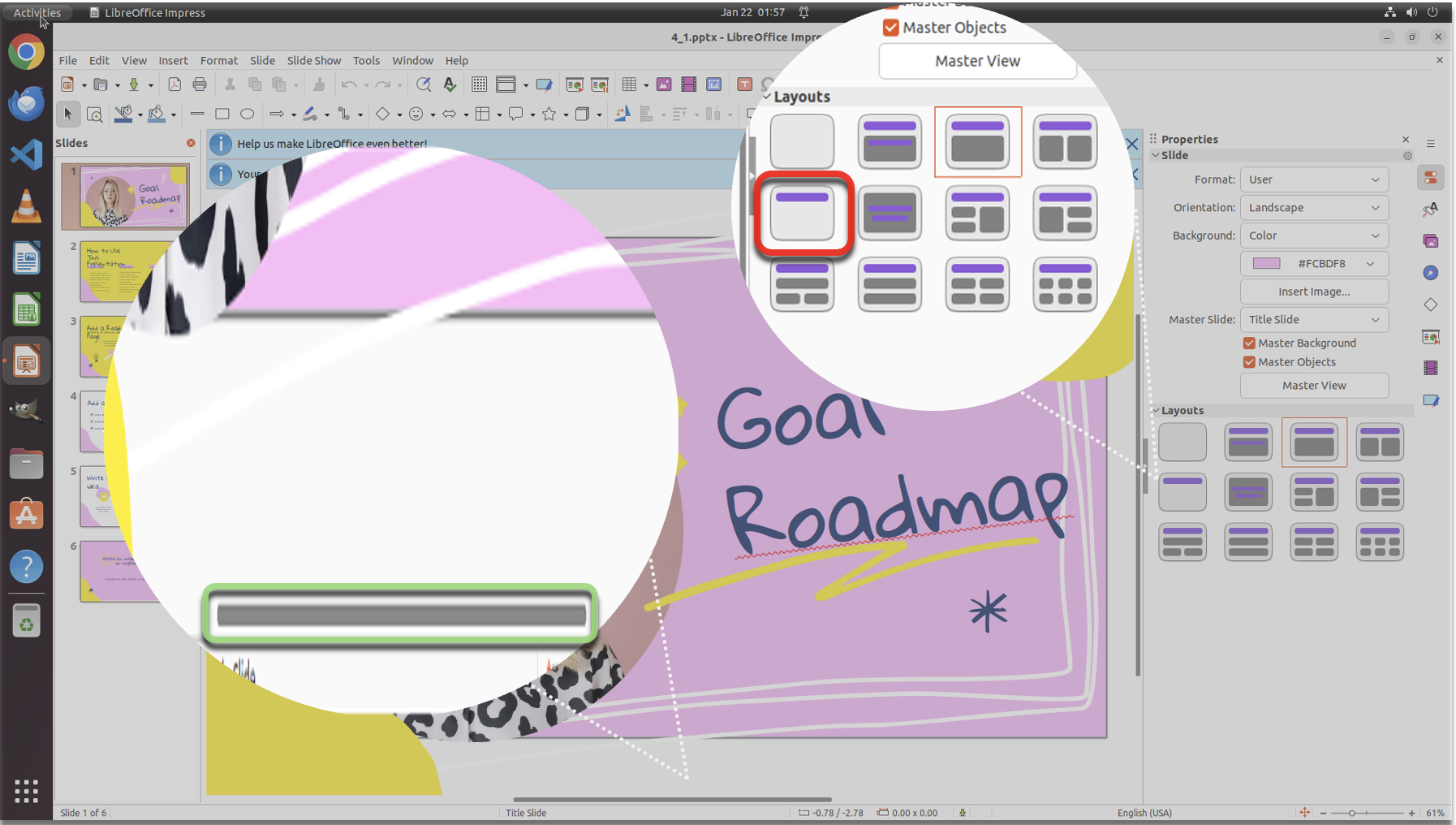}
        \subcaption{Instruction: Drag the horizontal scroll bar to center the image in the viewing area.}
        \label{fig:appendix_case_3}
    \end{subfigure}
    \hfill
    \begin{subfigure}[t]{0.48\textwidth}
        \centering
        \includegraphics[width=\linewidth,height=0.2\textheight,keepaspectratio]{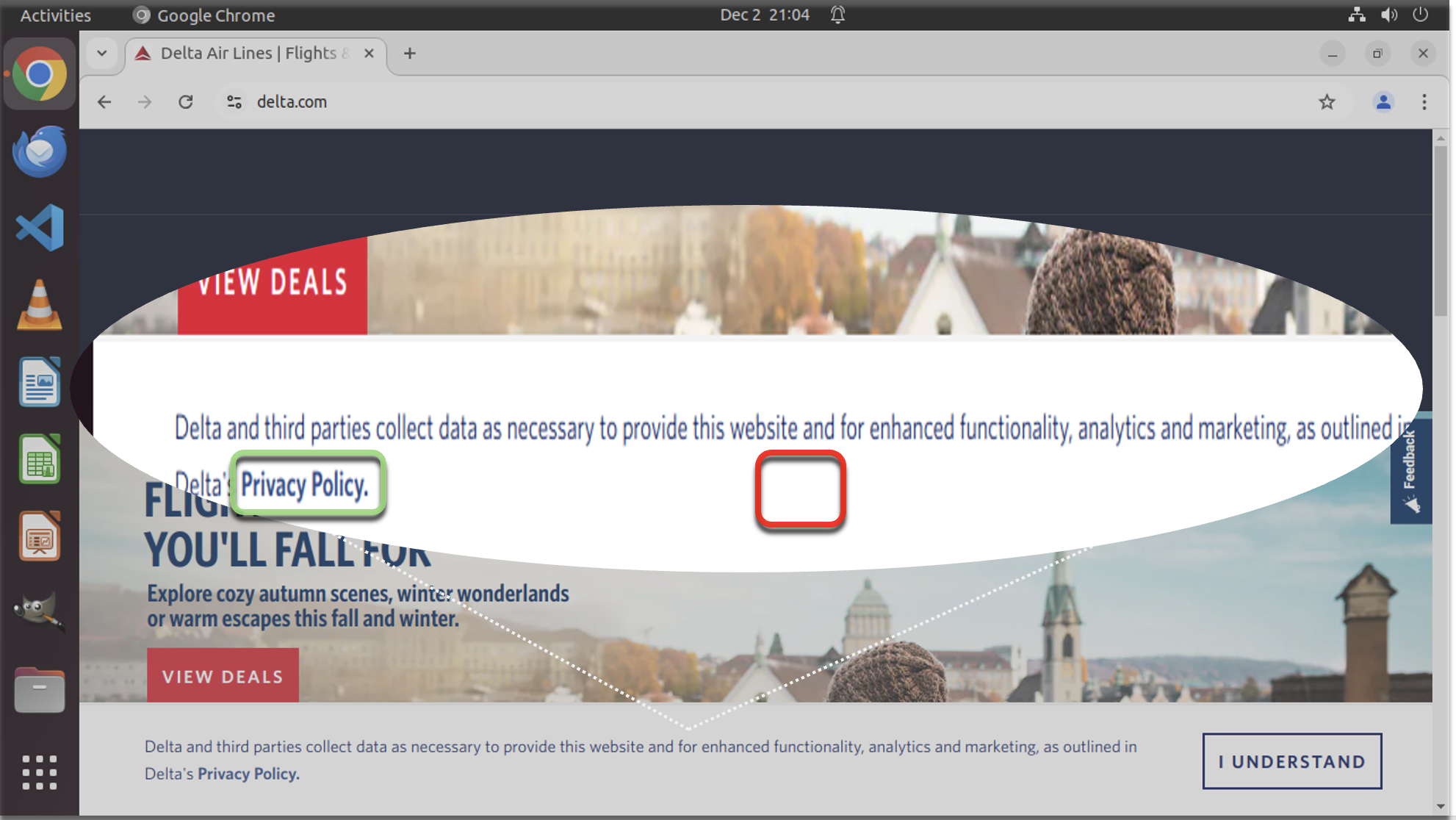}
        \subcaption{Instruction: Check the privacy policy of delta.com.}
        \label{fig:appendix_case_4}
    \end{subfigure}

    \begin{subfigure}[t]{0.48\textwidth}
        \centering
        \includegraphics[width=\linewidth,height=0.2\textheight,keepaspectratio]{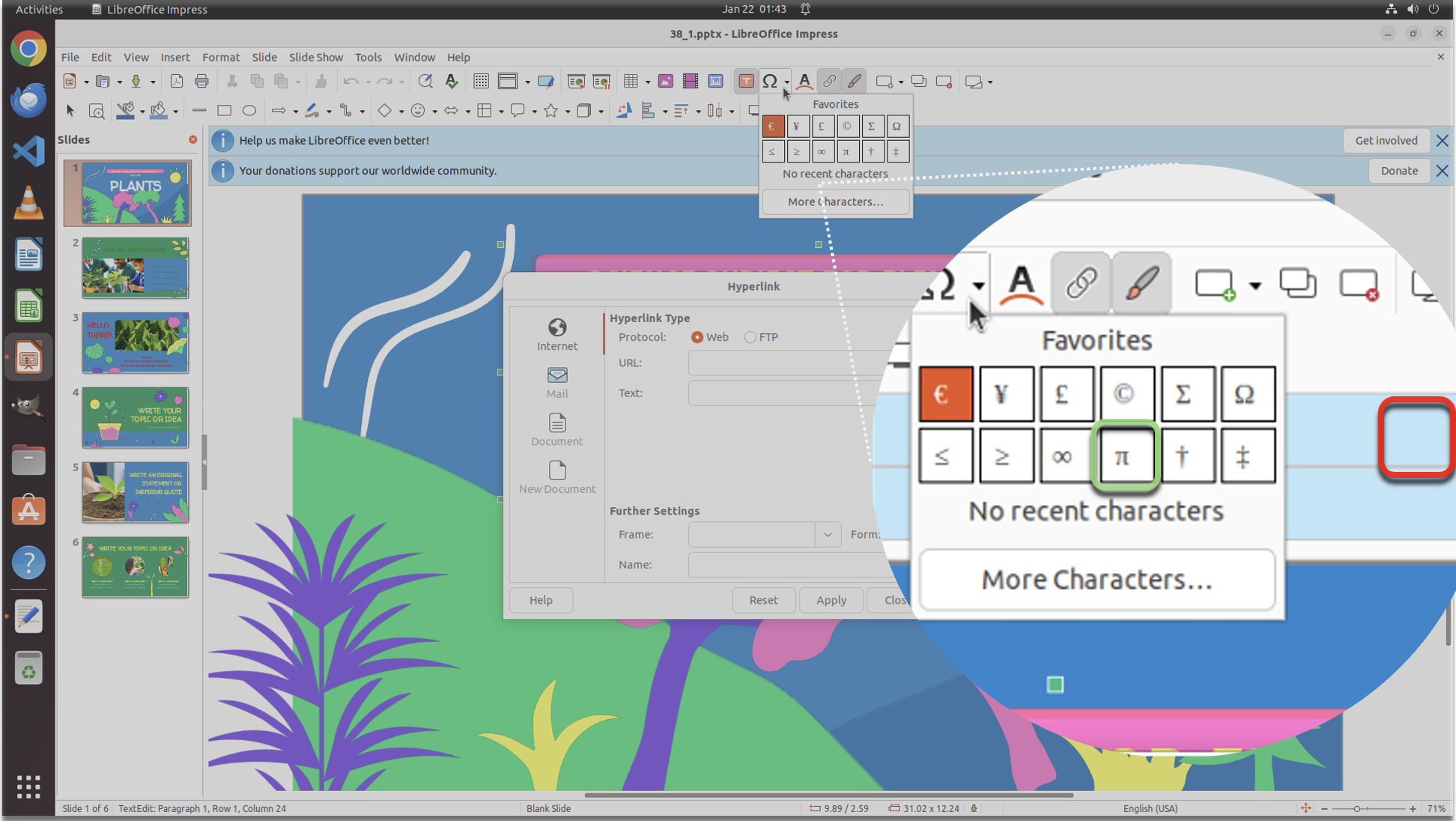}
        \subcaption{Instruction: Click on the character of PI.}
        \label{fig:appendix_case_5}
    \end{subfigure}
    \hfill
    \begin{subfigure}[t]{0.48\textwidth}
        \centering
        \includegraphics[width=\linewidth,height=0.2\textheight,keepaspectratio]{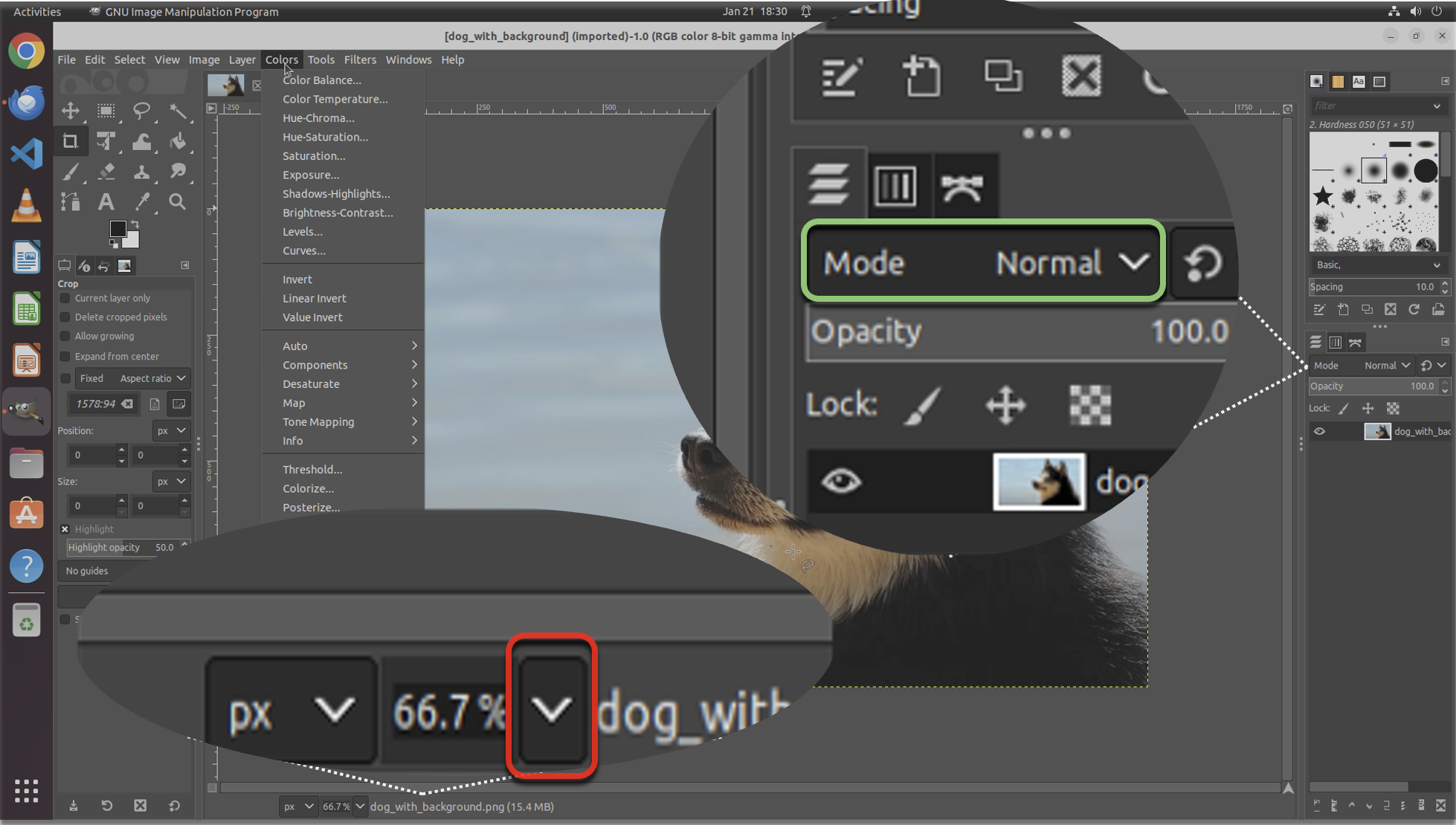}
        \subcaption{Instruction: Change the mode of this image.}
        \label{fig:appendix_case_6}
    \end{subfigure}

    \vspace{0.5\baselineskip} 

    \caption{Additional cases demonstrate \ours's improvement compared to Qwen2.5-VL-7B-Instruct. The green square represents the click position of \ours, while the red square indicates the click position of Qwen.}
    \label{fig:appendix_cases}
\end{figure}

%% file: images/appendix_case_failure.tex
\begin{figure}[H]
    \centering

    \begin{subfigure}[t]{0.48\textwidth}
        \centering
        \includegraphics[width=\linewidth,height=0.2\textheight,keepaspectratio]{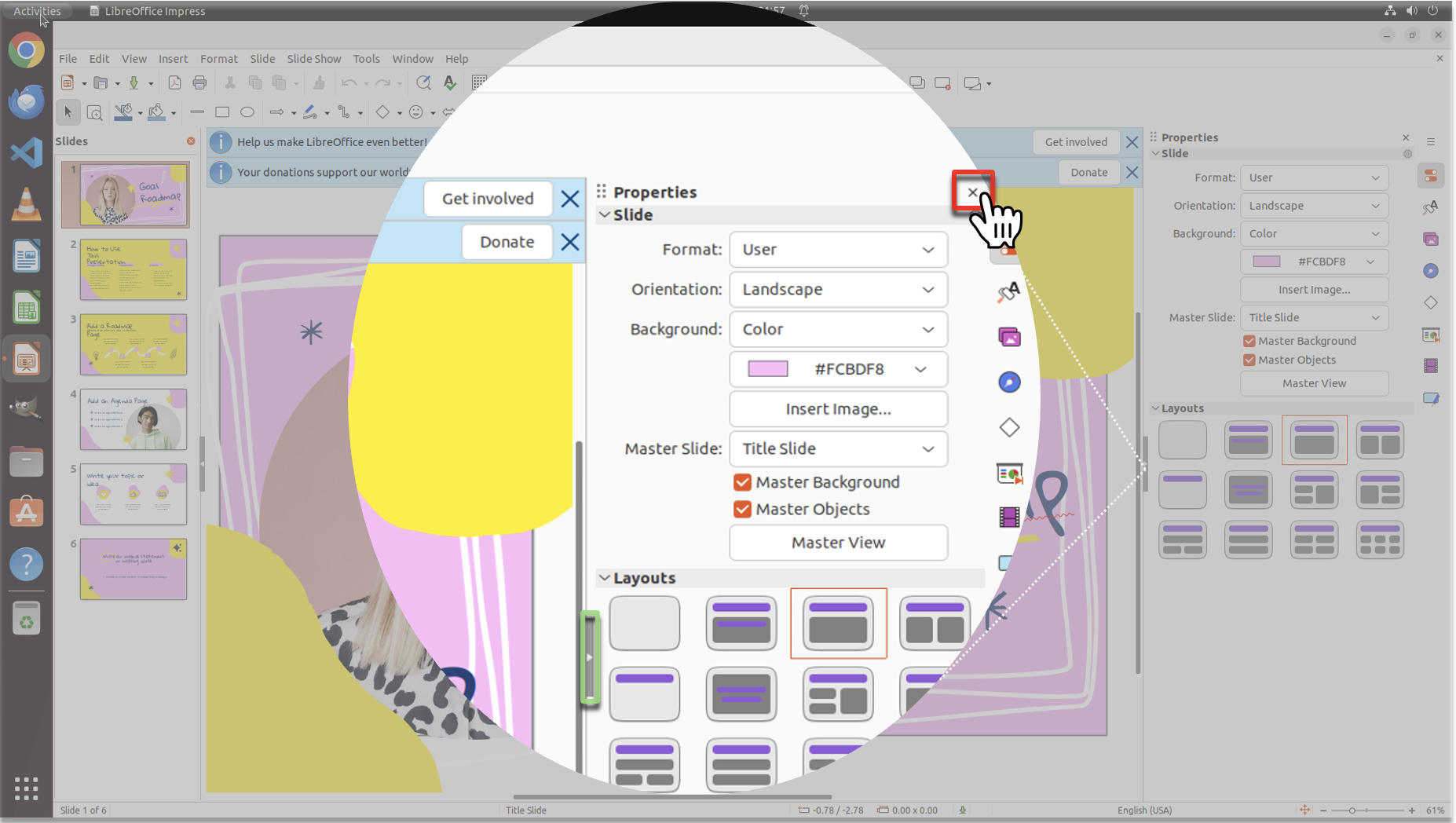}
        \subcaption{Instruction: Collapse the Properties panel by clicking on the right arrow.}
        \label{fig:row1_col1}
    \end{subfigure}
    \hfill
    \begin{subfigure}[t]{0.48\textwidth}
        \centering
        \includegraphics[width=\linewidth,height=0.2\textheight,keepaspectratio]{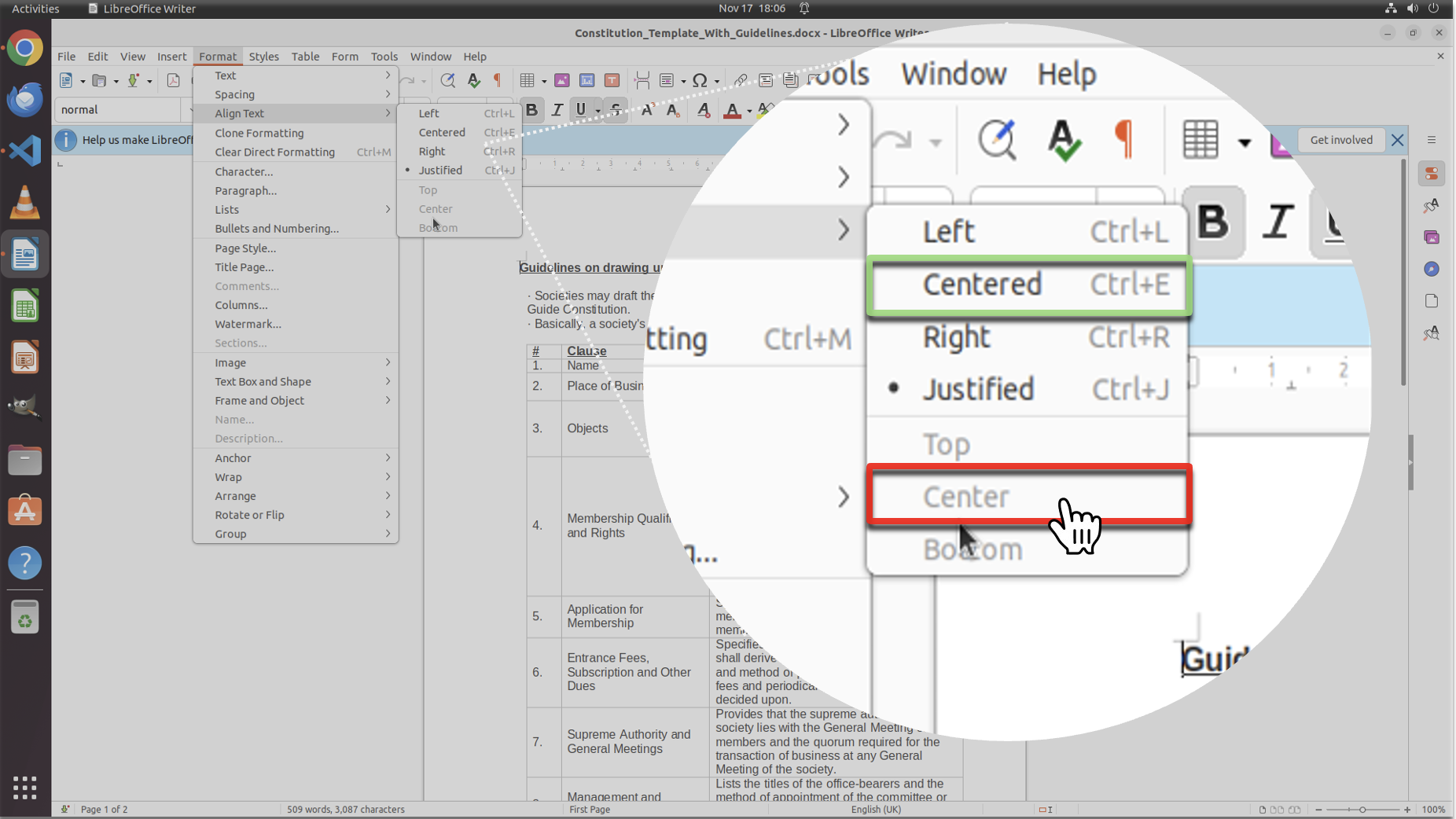}
        \subcaption{Instruction: Align the text to the center.}
        \label{fig:row1_col2}
    \end{subfigure}

    \vspace{0.5\baselineskip} 

    \begin{subfigure}[t]{0.48\textwidth}
        \centering
        \includegraphics[width=\linewidth,height=0.2\textheight,keepaspectratio]{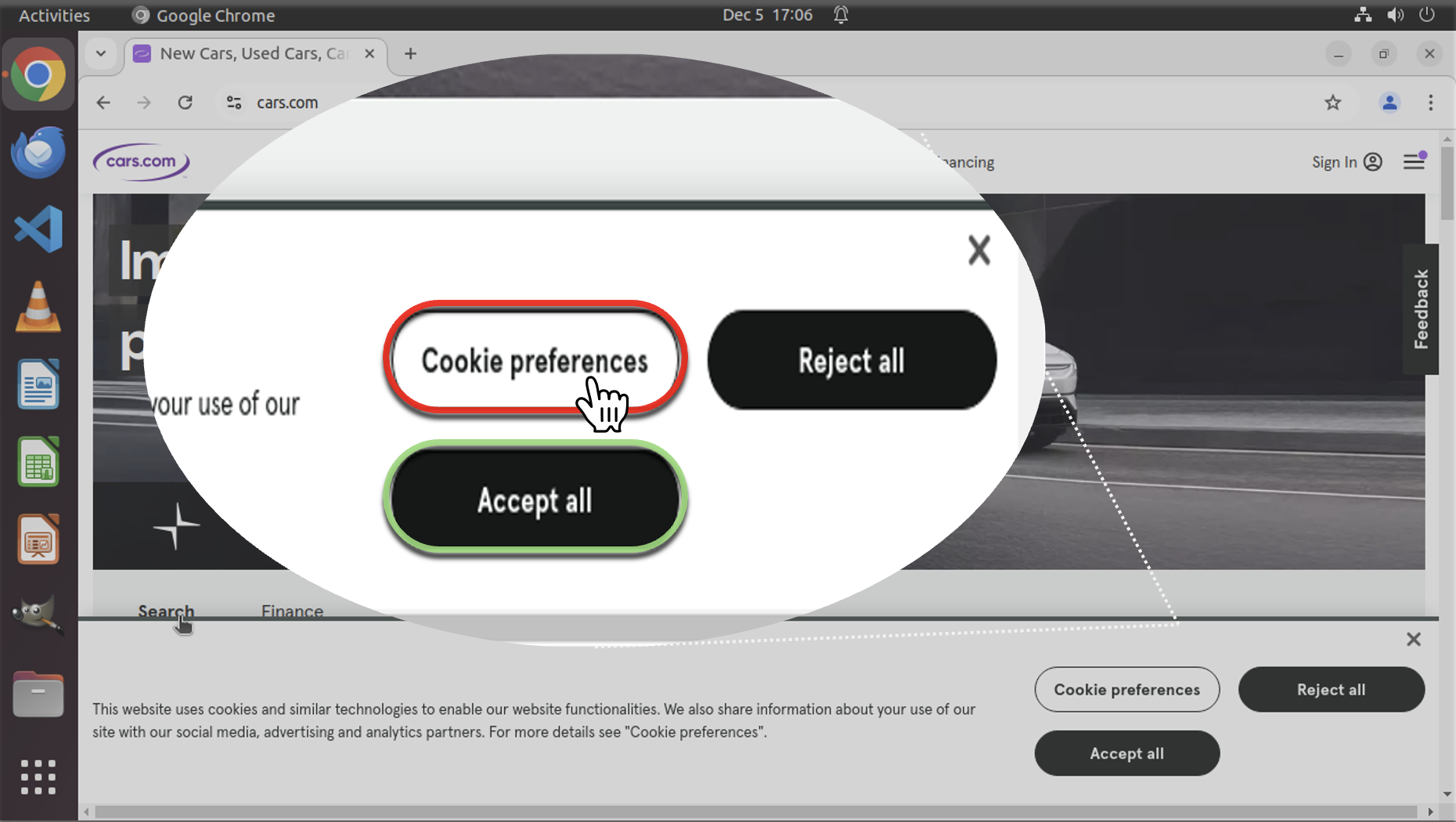}
        \subcaption{Instruction: Accept the cookie preferences.}
        \label{fig:row2_col1}
    \end{subfigure}
    \hfill
    \begin{subfigure}[t]{0.48\textwidth}
        \centering
        \includegraphics[width=\linewidth,height=0.2\textheight,keepaspectratio]{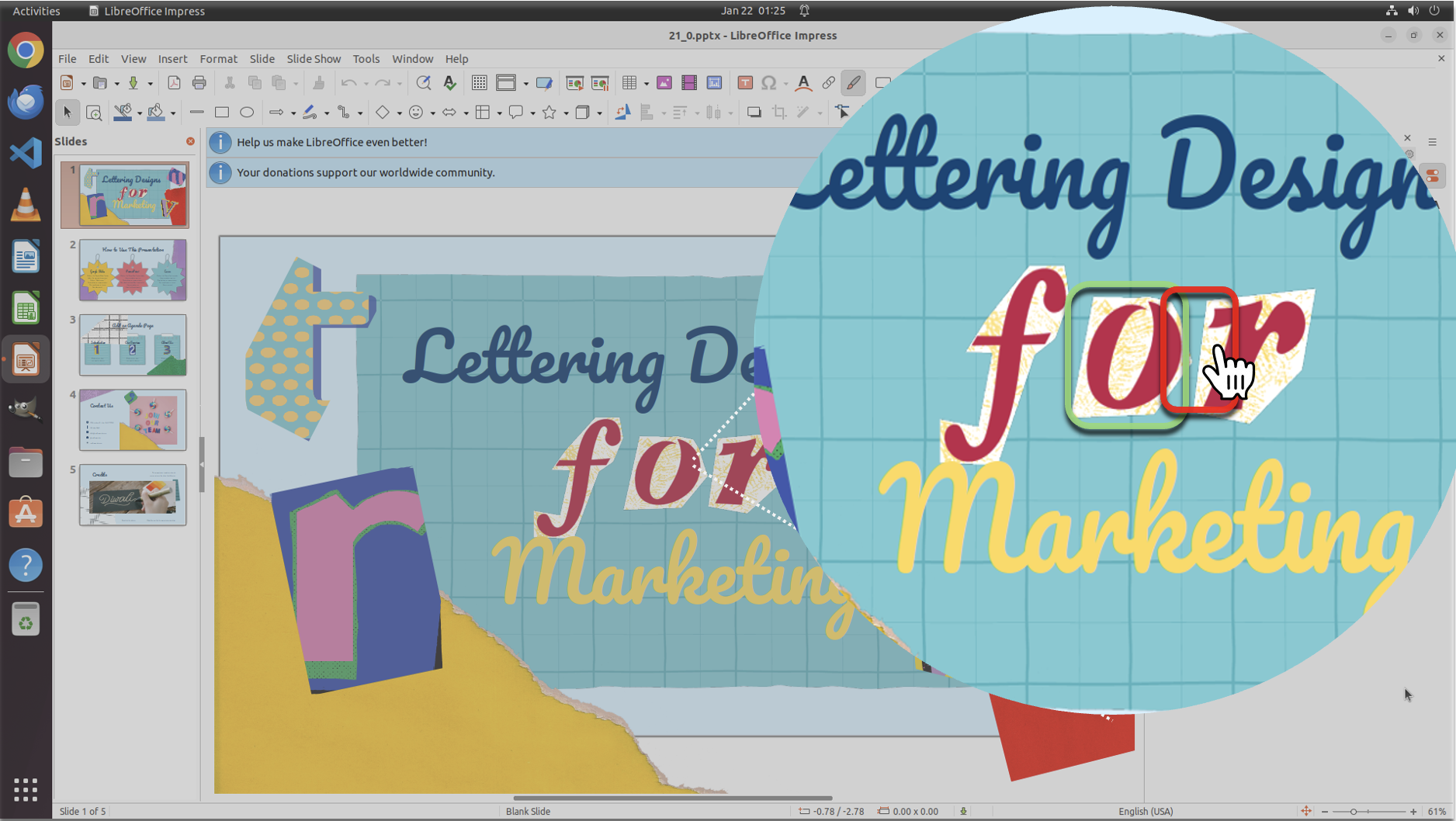}
        \subcaption{Instruction: Click on the letter "o" of the word "for" in the slide.}
        \label{fig:row2_col2}
    \end{subfigure}

    \caption{Four \ourbenchmarkname failure cases for \ours.}
    \label{fig:2x2_comparison}
\end{figure}

%% file: tables/appendix_osworld.tex
\begin{table}[htbp]
\centering
\caption{Detailed performance of \ours on OSWorld with four runs for each configuration}
\renewcommand{\arraystretch}{1.2}
\setlength{\extrarowheight}{2pt}
\resizebox{\textwidth}{!}{%
\begin{tabular}{cl|c|cccccccccc|ccccc}
\hline
\textbf{Model} & \textbf{Configuration} & \textbf{Overall} & \textbf{OS} & \textbf{Calc} & \textbf{Impress} & \textbf{Writer} & \textbf{VLC} & \textbf{TB} & \textbf{Chrome} & \textbf{VSC} & \textbf{GIMP} & \textbf{Workflow} & \textbf{OS} & \textbf{Office} & \textbf{Daily} & \textbf{Pro} & \textbf{Workflow} \\
\hline
\multirow{18}{*}{\ourmodelthreebillion} & \multicolumn{17}{l}{\textbf{15 Steps}} \\[2pt]
\cline{2-18}
& Run 1 & 21.95 & 39.13 & 6.38 & 8.57 & 26.08 & 29.41 & 20.00 & 26.09 & 56.52 & 50.00 & 11.83 & 39.13 & 11.13 & 25.64 & 53.06 & 11.83 \\
& Run 2 & 22.76 & 39.13 & 6.38 & 12.83 & 17.38 & 29.41 & 33.33 & 28.06 & 60.87 & 53.85 & 9.68 & 39.13 & 11.13 & 29.37 & 57.14 & 9.68 \\
& Run 3 & 22.37 & 43.48 & 6.38 & 12.77 & 17.38 & 17.65 & 33.33 & 25.94 & 56.52 & 53.85 & 11.39 & 43.48 & 11.11 & 25.55 & 55.10 & 11.39 \\
& Run 4 & 22.36 & 39.13 & 8.51 & 10.70 & 34.77 & 32.72 & 26.67 & 21.54 & 60.87 & 50.00 & 8.60 & 39.13 & 14.55 & 24.96 & 55.10 & 8.60 \\
& \textit{Pass@4} & 32.50 & 60.87 & 12.77 & 14.89 & 43.48 & 47.06 & 33.33 & 34.78 & 78.26 & 73.08 & 15.05 & 60.87 & 19.66 & 37.18 & 75.51 & 15.05 \\
& \textit{Avg} & 22.36 & - & - & - & - & - & - & - & - & - & - & - & - & - & - & - \\[2pt]
\cline{2-18}
& \multicolumn{17}{l}{\textbf{50 Steps}} \\[2pt]
\cline{2-18}
& Run 1 & 23.83 & 47.83 & 6.38 & 10.70 & 26.08 & 29.41 & 33.33 & 32.61 & 52.17 & 53.85 & 10.48 & 47.83 & 11.99 & 32.05 & 53.06 & 10.48 \\
& Run 2 & 24.73 & 43.48 & 8.70 & 12.77 & 39.12 & 23.53 & 40.00 & 30.43 & 47.83 & 64.00 & 9.78 & 43.48 & 16.54 & 30.77 & 56.25 & 9.78 \\
& Run 3 & 23.61 & 45.45 & 6.38 & 10.64 & 43.47 & 29.41 & 33.33 & 26.09 & 47.83 & 53.85 & 10.75 & 45.45 & 15.38 & 28.21 & 51.02 & 10.75 \\
& Run 4 & 22.36 & 39.13 & 8.51 & 12.77 & 21.73 & 23.53 & 33.33 & 23.91 & 56.52 & 53.85 & 9.68 & 39.13 & 13.46 & 25.64 & 55.10 & 9.68 \\
& \textit{Pass@4} & 33.33 & 52.17 & 10.64 & 12.77 & 56.52 & 47.06 & 46.67 & 36.96 & 78.26 & 76.92 & 15.05 & 52.17 & 20.51 & 41.03 & 77.55 & 15.05 \\
& \textit{Avg} & 23.63 & - & - & - & - & - & - & - & - & - & - & - & - & - & - & - \\[2pt]
\cline{2-18}
& \multicolumn{17}{l}{\textbf{100 Steps}} \\[2pt]
\cline{2-18}
& Run 1 & 24.43 & 38.10 & 8.51 & 13.11 & 26.08 & 23.53 & 46.67 & 32.40 & 43.48 & 68.00 & 9.68 & 38.10 & 13.82 & 33.21 & 56.25 & 9.68 \\
& Run 2 & 25.19 & 40.91 & 8.51 & 14.95 & 39.12 & 29.41 & 40.00 & 34.58 & 39.13 & 56.00 & 12.62 & 40.91 & 17.12 & 34.49 & 47.92 & 12.62 \\
& Run 3 & 23.66 & 45.45 & 10.64 & 17.47 & 30.42 & 23.53 & 20.00 & 25.88 & 56.52 & 53.85 & 9.78 & 45.45 & 17.27 & 24.24 & 55.10 & 9.78 \\
& Run 4 & 22.74 & 39.13 & 6.38 & 13.11 & 21.73 & 31.32 & 20.00 & 32.40 & 56.52 & 46.15 & 11.55 & 39.13 & 12.09 & 29.78 & 51.02 & 11.55 \\
& \textit{Pass@4} & 34.44 & 56.52 & 14.89 & 21.28 & 43.48 & 41.18 & 46.67 & 39.13 & 69.57 & 73.08 & 18.28 & 56.52 & 23.08 & 41.03 & 71.43 & 18.28 \\
& \textit{Avg} & 24.00 & - & - & - & - & - & - & - & - & - & - & - & - & - & - & - \\[2pt]
\hline
\multirow{18}{*}{\ourmodelsevenbillion} & \multicolumn{17}{l}{\textbf{15 Steps}} \\[2pt]
\cline{2-18}
& Run 1 & 22.20 & 43.48 & 8.51 & 10.70 & 30.42 & 35.29 & 26.67 & 23.71 & 56.52 & 56.00 & 5.38 & 43.48 & 13.70 & 24.00 & 56.25 & 5.38 \\
& Run 2 & 23.04 & 43.48 & 10.64 & 10.70 & 30.42 & 17.65 & 33.33 & 28.05 & 65.22 & 50.00 & 7.53 & 43.48 & 14.55 & 23.68 & 57.14 & 7.53 \\
& Run 3 & 22.42 & 34.78 & 10.64 & 10.70 & 21.73 & 29.41 & 33.33 & 25.88 & 65.22 & 42.31 & 10.75 & 34.78 & 12.84 & 26.20 & 53.06 & 10.75 \\
& Run 4 & 23.31 & 34.78 & 17.02 & 8.51 & 43.47 & 23.53 & 20.00 & 28.05 & 60.87 & 50.00 & 7.61 & 34.78 & 18.80 & 19.74 & 55.10 & 7.61 \\
& \textit{Pass@4} & 31.86 & 52.17 & 21.28 & 10.70 & 43.48 & 41.18 & 40.00 & 39.13 & 82.61 & 65.38 & 11.83 & 52.17 & 21.37 & 31.58 & 73.47 & 11.83 \\
& \textit{Avg} & 22.74 & - & - & - & - & - & - & - & - & - & - & - & - & - & - & - \\[2pt]
\cline{2-18}
& \multicolumn{17}{l}{\textbf{50 Steps}} \\[2pt]
\cline{2-18}
& Run 1 & 26.06 & 30.43 & 17.02 & 8.59 & 30.42 & 40.40 & 46.67 & 34.58 & 60.87 & 61.54 & 8.60 & 30.43 & 16.27 & 38.17 & 61.22 & 8.60 \\
& Run 2 & 26.27 & 47.83 & 19.15 & 10.71 & 43.47 & 38.96 & 40.00 & 25.88 & 56.52 & 46.15 & 10.75 & 47.83 & 20.54 & 31.45 & 51.02 & 10.75 \\
& Run 3 & 23.87 & 39.13 & 14.89 & 14.95 & 43.47 & 17.65 & 33.33 & 32.40 & 56.52 & 46.15 & 5.38 & 39.13 & 20.54 & 29.36 & 51.02 & 5.38 \\
& Run 4 & 23.87 & 34.78 & 12.77 & 10.70 & 34.77 & 23.53 & 33.33 & 25.88 & 69.57 & 53.85 & 8.60 & 34.78 & 16.26 & 26.8 & 61.22 & 8.60 \\
& \textit{Pass@4} & 35.56 & 52.17 & 25.53 & 14.89 & 56.52 & 47.06 & 53.33 & 39.13 & 86.96 & 65.38 & 13.98 & 52.17 & 27.35 & 43.59 & 75.51 & 13.98 \\
& \textit{Avg} & 25.02 & - & - & - & - & - & - & - & - & - & - & - & - & - & - & - \\[2pt]
\cline{2-18}
& \multicolumn{17}{l}{\textbf{100 Steps}} \\[2pt]
\cline{2-18}
& Run 1 & 25.94 & 39.13 & 14.89 & 16.30 & 34.77 & 29.41 & 26.67 & 32.40 & 60.87 & 46.15 & 12.90 & 39.13 & 19.39 & 30.65 & 53.06 & 12.90 \\
& Run 2 & 29.40 & 52.17 & 12.77 & 14.95 & 43.47 & 29.41 & 46.67 & 36.75 & 73.91 & 57.69 & 11.83 & 52.17 & 19.68 & 37.06 & 65.31 & 11.83 \\
& Run 3 & 25.64 & 43.48 & 6.38 & 14.95 & 36.35 & 20.72 & 53.33 & 28.05 & 73.91 & 46.15 & 11.68 & 43.48 & 15.54 & 31.32 & 59.18 & 11.68 \\
& Run 4 & 26.86 & 34.78 & 10.64 & 10.71 & 39.12 & 29.41 & 46.67 & 32.40 & 78.26 & 53.85 & 11.56 & 34.78 & 16.27 & 34.49 & 65.31 & 11.56 \\
& \textit{Pass@4} & 38.89 & 65.22 & 21.28 & 21.28 & 60.87 & 41.18 & 53.33 & 45.65 & 95.65 & 65.38 & 17.2 & 65.22 & 29.06 & 46.15 & 79.59 & 17.2 \\
& \textit{Avg} & 27.04 & - & - & - & - & - & - & - & - & - & - & - & - & - & - & - \\
\hline
\end{tabular}
}
\label{tab:performance_comparison}
\end{table}

%% file: tables/appendix_waa.tex
\begin{table}[htbp]
\centering
\caption{Detailed performance of \ours on WindowsAgentArena with four runs for each configuration}
\renewcommand{\arraystretch}{1.2}
\setlength{\extrarowheight}{2pt}
\resizebox{\textwidth}{!}{%
\begin{tabular}{cl|c|ccccccccccc}
\hline
\textbf{Model} & \textbf{Configuration} & \textbf{Overall} & \textbf{Chrome} & \textbf{File Explorer} & \textbf{Notepad} & \textbf{Edge} & \textbf{OS Settings} & \textbf{VLC} & \textbf{VS Code} & \textbf{Calculator} & \textbf{Libre Calc} & \textbf{Libre Writer} & \textbf{Paint} \\
\hline
\multirow{18}{*}{\ourmodelthreebillion} & \multicolumn{13}{l}{\textbf{15 Steps}} \\[2pt]
\cline{2-14}
& Run 1 & 28.86 & 0.00 & 47.37 & 50.00 & 30.77 & 60.00 & 38.10 & 45.83 & 0.00 & 4.17 & 26.30 & 33.33 \\
& Run 2 & 28.72 & 5.88 & 36.84 & 50.00 & 23.08 & 80.00 & 52.75 & 45.83 & 0.00 & 8.33 & 10.53 & 33.33 \\
& Run 3 & 29.92 & 5.88 & 47.37 & 50.00 & 30.77 & 40.00 & 42.32 & 50.00 & 0.00 & 4.17 & 26.30 & 33.33 \\
& Run 4 & 28.72 & 5.88 & 42.11 & 50.00 & 38.46 & 60.00 & 43.23 & 37.50 & 0.00 & 8.33 & 21.04 & 33.33 \\
& \textit{Pass@4} & 41.33 & 5.88 & 57.89 & 50.00 & 38.46 & 80.00 & 57.14 & 70.83 & 0.00 & 8.33 & 42.11 & 33.33 \\
& \textit{Avg} & 29.06 & - & - & - & - & - & - & - & - & - & - & - \\[2pt]
\cline{2-14}
& \multicolumn{13}{l}{\textbf{50 Steps}} \\[2pt]
\cline{2-14}
& Run 1 & 32.05 & 5.88 & 47.37 & 50.00 & 30.77 & 60.00 & 43.23 & 45.83 & 0.00 & 8.33 & 31.57 & 66.67 \\
& Run 2 & 32.48 & 5.88 & 44.44 & 50.00 & 38.46 & 40.00 & 47.99 & 52.17 & 0.00 & 4.17 & 36.83 & 33.33 \\
& Run 3 & 32.05 & 0.00 & 57.89 & 0.00 & 23.08 & 60.00 & 52.75 & 50.00 & 0.00 & 8.33 & 26.30 & 33.33 \\
& Run 4 & 28.72 & 5.88 & 42.11 & 50.00 & 38.46 & 60.00 & 42.23 & 37.50 & 0.00 & 8.33 & 21.04 & 33.33 \\
& \textit{Pass@4} & 44.00 & 5.88 & 63.16 & 50.00 & 53.85 & 60.00 & 52.38 & 75.00 & 0.00 & 8.33 & 47.37 & 66.67 \\
& \textit{Avg} & 31.33 & - & - & - & - & - & - & - & - & - & - & - \\[2pt]
\cline{2-14}
& \multicolumn{13}{l}{\textbf{100 Steps}} \\[2pt]
\cline{2-14}
& Run 1 & 34.57 & 5.88 & 52.63 & 50.00 & 38.46 & 80.00 & 46.91 & 45.83 & 0.00 & 12.50 & 31.57 & 33.33 \\
& Run 2 & 30.72 & 5.88 & 57.89 & 50.00 & 30.77 & 80.00 & 33.70 & 33.33 & 0.00 & 8.33 & 36.83 & 33.33 \\
& Run 3 & 33.23 & 5.88 & 63.16 & 50.00 & 7.69 & 60.00 & 42.15 & 58.33 & 0.00 & 8.33 & 31.58 & 33.33 \\
& Run 4 & 33.61 & 6.25 & 47.37 & 50.00 & 30.77 & 80.00 & 43.23 & 58.33 & 0.00 & 8.33 & 21.04 & 66.67 \\
& \textit{Pass@4} & 46.67 & 11.76 & 63.16 & 50.00 & 53.85 & 80.00 & 57.14 & 70.83 & 0.00 & 12.50 & 52.63 & 66.67 \\
& \textit{Avg} & 33.03 & - & - & - & - & - & - & - & - & - & - & - \\[2pt]
\hline
\multirow{18}{*}{\ourmodelsevenbillion} & \multicolumn{13}{l}{\textbf{15 Steps}} \\[2pt]
\cline{2-14}
& Run 1 & 30.00 & 5.88 & 31.58 & 50.00 & 23.08 & 40.00 & 52.38 & 41.67 & 0.00 & 8.33 & 36.83 & 66.67 \\
& Run 2 & 29.38 & 0.00 & 31.58 & 50.00 & 23.08 & 60.00 & 43.23 & 50.00 & 0.00 & 8.33 & 31.57 & 66.67 \\
& Run 3 & 31.90 & 0.00 & 42.11 & 50.00 & 38.46 & 60.00 & 42.15 & 50.00 & 0.00 & 4.17 & 36.83 & 66.67 \\
& Run 4 & 29.38 & 0.00 & 42.11 & 50.00 & 30.77 & 60.00 & 42.23 & 41.67 & 0.00 & 8.33 & 26.30 & 66.67 \\
& \textit{Pass@4} & 42.67 & 5.88 & 52.63 & 50.00 & 46.15 & 60.00 & 57.14 & 70.83 & 0.00 & 8.33 & 52.63 & 66.67 \\
& \textit{Avg} & 30.17 & - & - & - & - & - & - & - & - & - & - & - \\[2pt]
\cline{2-14}
& \multicolumn{13}{l}{\textbf{50 Steps}} \\[2pt]
\cline{2-14}
& Run 1 & 32.57 & 0.00 & 52.63 & 50.00 & 30.77 & 80.00 & 46.91 & 50.00 & 0.00 & 4.17 & 26.30 & 66.67 \\
& Run 2 & 32.57 & 11.76 & 47.37 & 50.00 & 46.15 & 60.00 & 51.67 & 41.67 & 0.00 & 4.17 & 26.30 & 33.33 \\
& Run 3 & 34.05 & 0.00 & 47.37 & 50.00 & 46.15 & 80.00 & 43.23 & 50.00 & 33.33 & 4.17 & 31.57 & 66.67 \\
& Run 4 & 32.00 & 0.00 & 42.11 & 50.00 & 46.15 & 60.00 & 52.38 & 45.83 & 0.00 & 8.33 & 26.30 & 33.33 \\
& \textit{Pass@4} & 46.00 & 11.76 & 52.63 & 50.00 & 61.54 & 80.00 & 61.90 & 70.83 & 33.33 & 8.33 & 47.37 & 66.67 \\
& \textit{Avg} & 32.80 & - & - & - & - & - & - & - & - & - & - & - \\[2pt]
\cline{2-14}
& \multicolumn{13}{l}{\textbf{100 Steps}} \\[2pt]
\cline{2-14}
& Run 1 & 33.90 & 0.00 & 52.63 & 50.00 & 30.77 & 80.00 & 46.91 & 54.17 & 0.00 & 8.33 & 31.57 & 33.33 \\
& Run 2 & 34.67 & 5.88 & 47.37 & 50.00 & 38.46 & 60.00 & 52.38 & 45.83 & 0.00 & 8.33 & 36.83 & 66.67 \\
& Run 3 & 33.46 & 0.00 & 47.37 & 50.00 & 38.46 & 80.00 & 43.76 & 45.83 & 0.00 & 8.33 & 42.09 & 33.33 \\
& Run 4 & 32.67 & 5.88 & 52.63 & 50.00 & 38.46 & 40.00 & 42.86 & 45.83 & 33.33 & 8.33 & 31.57 & 33.33 \\
& \textit{Pass@4} & 47.33 & 5.88 & 63.16 & 50.00 & 53.85 & 80.00 & 61.90 & 75.00 & 33.33 & 8.33 & 52.63 & 66.67 \\
& \textit{Avg} & 33.68 & - & - & - & - & - & - & - & - & - & - & - \\
\hline
\end{tabular}
}
\label{tab:performance_comparison_windows}
\end{table}

%% file: main.bbl
\begin{thebibliography}{56}
\providecommand{\natexlab}[1]{#1}
\providecommand{\url}[1]{\texttt{#1}}
\expandafter\ifx\csname urlstyle\endcsname\relax
  \providecommand{\doi}[1]{doi: #1}\else
  \providecommand{\doi}{doi: \begingroup \urlstyle{rm}\Url}\fi

\bibitem[The()]{TheC3}
The claude 3 model family: Opus, sonnet, haiku.
\newblock URL \url{https://api.semanticscholar.org/CorpusID:268232499}.

\bibitem[Agashe et~al.(2024)Agashe, Han, Gan, Yang, Li, and Wang]{agashe2024agentsopenagentic}
Saaket Agashe, Jiuzhou Han, Shuyu Gan, Jiachen Yang, Ang Li, and Xin~Eric Wang.
\newblock Agent s: An open agentic framework that uses computers like a human, 2024.
\newblock URL \url{https://arxiv.org/abs/2410.08164}.

\bibitem[Anthropic()]{AnthropicModelCA}
Sonnet Anthropic.
\newblock Model card addendum: Claude 3.5 haiku and upgraded claude 3.5 sonnet.
\newblock URL \url{https://api.semanticscholar.org/CorpusID:273639283}.

\bibitem[Bai et~al.(2021)Bai, Zang, Xu, Sunkara, Rastogi, Chen, et~al.]{bai2021uibert}
Chongyang Bai, Xiaoxue Zang, Ying Xu, Srinivas Sunkara, Abhinav Rastogi, Jindong Chen, et~al.
\newblock Uibert: Learning generic multimodal representations for ui understanding.
\newblock \emph{arXiv preprint arXiv:2107.13731}, 2021.

\bibitem[Bai et~al.(2025{\natexlab{a}})Bai, Zhou, Pan, Cemri, Suhr, Levine, and Kumar]{bai2025digirl}
Hao Bai, Yifei Zhou, Jiayi Pan, Mert Cemri, Alane Suhr, Sergey Levine, and Aviral Kumar.
\newblock Digirl: Training in-the-wild device-control agents with autonomous reinforcement learning.
\newblock \emph{Advances in Neural Information Processing Systems}, 37:\penalty0 12461--12495, 2025{\natexlab{a}}.

\bibitem[Bai et~al.(2025{\natexlab{b}})Bai, Chen, Liu, Wang, Ge, Song, Dang, Wang, Wang, Tang, Zhong, Zhu, Yang, Li, Wan, Wang, Ding, Fu, Xu, Ye, Zhang, Xie, Cheng, Zhang, Yang, Xu, and Lin]{bai2025qwen25vltechnicalreport}
Shuai Bai, Keqin Chen, Xuejing Liu, Jialin Wang, Wenbin Ge, Sibo Song, Kai Dang, Peng Wang, Shijie Wang, Jun Tang, Humen Zhong, Yuanzhi Zhu, Mingkun Yang, Zhaohai Li, Jianqiang Wan, Pengfei Wang, Wei Ding, Zheren Fu, Yiheng Xu, Jiabo Ye, Xi~Zhang, Tianbao Xie, Zesen Cheng, Hang Zhang, Zhibo Yang, Haiyang Xu, and Junyang Lin.
\newblock Qwen2.5-vl technical report, 2025{\natexlab{b}}.
\newblock URL \url{https://arxiv.org/abs/2502.13923}.

\bibitem[Bonatti et~al.(2024)Bonatti, Zhao, Bonacci, Dupont, Abdali, Li, Lu, Wagle, Koishida, Bucker, Jang, and Hui]{Bonatti2024WindowsAA}
Rogerio Bonatti, Dan Zhao, Francesco Bonacci, Dillon Dupont, Sara Abdali, Yinheng Li, Yadong Lu, Justin Wagle, Kazuhito Koishida, Arthur Fender~C. Bucker, Lawrence Jang, and Zack Hui.
\newblock Windows agent arena: Evaluating multi-modal os agents at scale.
\newblock \emph{ArXiv}, abs/2409.08264, 2024.
\newblock URL \url{https://api.semanticscholar.org/CorpusID:272600411}.

\bibitem[Chen et~al.(2024{\natexlab{a}})Chen, Cui, Hu, Qin, Fang, Zhao, Wang, Liu, Chen, Huo, et~al.]{chen2024guicourse}
Wentong Chen, Junbo Cui, Jinyi Hu, Yujia Qin, Junjie Fang, Yue Zhao, Chongyi Wang, Jun Liu, Guirong Chen, Yupeng Huo, et~al.
\newblock Guicourse: From general vision language models to versatile gui agents.
\newblock \emph{arXiv preprint arXiv:2406.11317}, 2024{\natexlab{a}}.

\bibitem[Chen et~al.(2024{\natexlab{b}})Chen, Wang, Tian, Ye, Gao, Cui, Tong, Hu, Luo, Ma, Ma, Wang, Dong, Yan, Guo, He, Shi, Jin, Xu, Wang, Wei, Li, Zhang, Zhang, Cai, Wen, Yan, Dou, Lu, Zhu, Lu, Lin, Qiao, Dai, and Wang]{chen2024fargpt4vclosinggap}
Zhe Chen, Weiyun Wang, Hao Tian, Shenglong Ye, Zhangwei Gao, Erfei Cui, Wenwen Tong, Kongzhi Hu, Jiapeng Luo, Zheng Ma, Ji~Ma, Jiaqi Wang, Xiaoyi Dong, Hang Yan, Hewei Guo, Conghui He, Botian Shi, Zhenjiang Jin, Chao Xu, Bin Wang, Xingjian Wei, Wei Li, Wenjian Zhang, Bo~Zhang, Pinlong Cai, Licheng Wen, Xiangchao Yan, Min Dou, Lewei Lu, Xizhou Zhu, Tong Lu, Dahua Lin, Yu~Qiao, Jifeng Dai, and Wenhai Wang.
\newblock How far are we to gpt-4v? closing the gap to commercial multimodal models with open-source suites, 2024{\natexlab{b}}.
\newblock URL \url{https://arxiv.org/abs/2404.16821}.

\bibitem[Cheng et~al.(2024)Cheng, Sun, Chu, Xu, Li, Zhang, and Wu]{cheng2024seeclickharnessingguigrounding}
Kanzhi Cheng, Qiushi Sun, Yougang Chu, Fangzhi Xu, Yantao Li, Jianbing Zhang, and Zhiyong Wu.
\newblock Seeclick: Harnessing gui grounding for advanced visual gui agents, 2024.
\newblock URL \url{https://arxiv.org/abs/2401.10935}.

\bibitem[Deka et~al.(2017)Deka, Huang, Franzen, Hibschman, Afergan, Li, Nichols, and Kumar]{deka2017rico}
Biplab Deka, Zifeng Huang, Chad Franzen, Joshua Hibschman, Daniel Afergan, Yang Li, Jeffrey Nichols, and Ranjitha Kumar.
\newblock Rico: A mobile app dataset for building data-driven design applications.
\newblock In \emph{Proceedings of the 30th annual ACM symposium on user interface software and technology}, pages 845--854, 2017.

\bibitem[Deng et~al.(2023)Deng, Gu, Zheng, Chen, Stevens, Wang, Sun, and Su]{deng2023mind2web}
Xiang Deng, Yu~Gu, Boyuan Zheng, Shijie Chen, Samuel Stevens, Boshi Wang, Huan Sun, and Yu~Su.
\newblock Mind2web: Towards a generalist agent for the web.
\newblock \emph{arXiv preprint arXiv:2306.06070}, 2023.

\bibitem[Drouin et~al.(2024)Drouin, Gasse, Caccia, Laradji, Del~Verme, Marty, Boisvert, Thakkar, Cappart, Vazquez, et~al.]{drouin2024workarena}
Alexandre Drouin, Maxime Gasse, Massimo Caccia, Issam~H Laradji, Manuel Del~Verme, Tom Marty, L{\'e}o Boisvert, Megh Thakkar, Quentin Cappart, David Vazquez, et~al.
\newblock Workarena: How capable are web agents at solving common knowledge work tasks?
\newblock \emph{arXiv preprint arXiv:2403.07718}, 2024.

\bibitem[Fan et~al.(2022)Fan, Wang, Jiang, Mandlekar, Yang, Zhu, Tang, Huang, Zhu, and Anandkumar]{fan2022minedojo}
Linxi Fan, Guanzhi Wang, Yunfan Jiang, Ajay Mandlekar, Yuncong Yang, Haoyi Zhu, Andrew Tang, De-An Huang, Yuke Zhu, and Anima Anandkumar.
\newblock Minedojo: Building open-ended embodied agents with internet-scale knowledge.
\newblock \emph{Advances in Neural Information Processing Systems}, 35:\penalty0 18343--18362, 2022.

\bibitem[Gou et~al.(2024)Gou, Wang, Zheng, Xie, Chang, Shu, Sun, and Su]{gou2024navigatingdigitalworldhumans}
Boyu Gou, Ruohan Wang, Boyuan Zheng, Yanan Xie, Cheng Chang, Yiheng Shu, Huan Sun, and Yu~Su.
\newblock Navigating the digital world as humans do: Universal visual grounding for gui agents, 2024.
\newblock URL \url{https://arxiv.org/abs/2410.05243}.

\bibitem[Hong et~al.(2023)Hong, Wang, Lv, Xu, Yu, Ji, Wang, Wang, Dong, Ding, et~al.]{hong2023cogagent}
Wenyi Hong, Weihan Wang, Qingsong Lv, Jiazheng Xu, Wenmeng Yu, Junhui Ji, Yan Wang, Zihan Wang, Yuxiao Dong, Ming Ding, et~al.
\newblock Cogagent: A visual language model for gui agents.
\newblock \emph{arXiv preprint arXiv:2312.08914}, 2023.

\bibitem[Hu et~al.(2024)Hu, Shi, Fu, Roth, Ostendorf, Zettlemoyer, Smith, and Krishna]{hu2024visual}
Yushi Hu, Weijia Shi, Xingyu Fu, Dan Roth, Mari Ostendorf, Luke Zettlemoyer, Noah~A Smith, and Ranjay Krishna.
\newblock Visual sketchpad: Sketching as a visual chain of thought for multimodal language models.
\newblock \emph{arXiv preprint arXiv:2406.09403}, 2024.

\bibitem[Kapoor et~al.(2024)Kapoor, Butala, Russak, Koh, Kamble, Alshikh, and Salakhutdinov]{Kapoor2024OmniACTAD}
Raghav Kapoor, Yash~Parag Butala, Melisa Russak, Jing~Yu Koh, Kiran Kamble, Waseem Alshikh, and Ruslan Salakhutdinov.
\newblock Omniact: A dataset and benchmark for enabling multimodal generalist autonomous agents for desktop and web.
\newblock \emph{arXiv preprint arXiv:2402.17553}, 2024.

\bibitem[Koh et~al.(2024)Koh, Lo, Jang, Duvvur, Lim, Huang, Neubig, Zhou, Salakhutdinov, and Fried]{Koh2024VisualWebArenaEM}
Jing~Yu Koh, Robert Lo, Lawrence Jang, Vikram Duvvur, Ming~Chong Lim, Po-Yu Huang, Graham Neubig, Shuyan Zhou, Ruslan Salakhutdinov, and Daniel Fried.
\newblock Visualwebarena: Evaluating multimodal agents on realistic visual web tasks.
\newblock \emph{arXiv preprint arXiv:2401.13649}, 2024.

\bibitem[Li et~al.(2024)Li, Bishop, Li, Rawles, Campbell-Ajala, Tyamagundlu, and Riva]{Li2024OnTE}
Wei Li, Will Bishop, Alice Li, Christopher Rawles, Folawiyo Campbell-Ajala, Divya Tyamagundlu, and Oriana Riva.
\newblock On the effects of data scale on ui control agents.
\newblock In \emph{Neural Information Processing Systems}, 2024.
\newblock URL \url{https://api.semanticscholar.org/CorpusID:270285816}.

\bibitem[Li et~al.(2020{\natexlab{a}})Li, He, Zhou, Zhang, and Baldridge]{li2020mapping}
Yang Li, Jiacong He, Xin Zhou, Yuan Zhang, and Jason Baldridge.
\newblock Mapping natural language instructions to mobile ui action sequences.
\newblock \emph{arXiv preprint arXiv:2005.03776}, 2020{\natexlab{a}}.

\bibitem[Li et~al.(2020{\natexlab{b}})Li, Li, He, Zheng, Li, and Guan]{li2020widget}
Yang Li, Gang Li, Luheng He, Jingjie Zheng, Hong Li, and Zhiwei Guan.
\newblock Widget captioning: Generating natural language description for mobile user interface elements.
\newblock \emph{arXiv preprint arXiv:2010.04295}, 2020{\natexlab{b}}.

\bibitem[Lin et~al.(2024)Lin, Li, Gao, Yang, Bai, Lei, Wang, and Shou]{lin2024showui}
Kevin~Qinghong Lin, Linjie Li, Difei Gao, Zhengyuan Yang, Zechen Bai, Weixian Lei, Lijuan Wang, and Mike~Zheng Shou.
\newblock Showui: One vision-language-action model for generalist gui agent.
\newblock In \emph{NeurIPS 2024 Workshop on Open-World Agents}, 2024.

\bibitem[Liu et~al.(2018)Liu, Guu, Pasupat, Shi, and Liang]{liu2018reinforcement}
Evan~Zheran Liu, Kelvin Guu, Panupong Pasupat, Tianlin Shi, and Percy Liang.
\newblock Reinforcement learning on web interfaces using workflow-guided exploration.
\newblock \emph{arXiv preprint arXiv:1802.08802}, 2018.

\bibitem[Liu et~al.(2025)Liu, Zhang, Xu, Wanyan, Wang, Yan, Zhang, Yuan, Xu, Hu, et~al.]{liu2025pc}
Haowei Liu, Xi~Zhang, Haiyang Xu, Yuyang Wanyan, Junyang Wang, Ming Yan, Ji~Zhang, Chunfeng Yuan, Changsheng Xu, Weiming Hu, et~al.
\newblock Pc-agent: A hierarchical multi-agent collaboration framework for complex task automation on pc.
\newblock \emph{arXiv preprint arXiv:2502.14282}, 2025.

\bibitem[Lu et~al.(2024)Lu, Yang, Shen, and Awadallah]{lu2024omniparser}
Yadong Lu, Jianwei Yang, Yelong Shen, and Ahmed Awadallah.
\newblock Omniparser for pure vision based gui agent.
\newblock \emph{arXiv preprint arXiv:2408.00203}, 2024.

\bibitem[Mathew et~al.(2020)Mathew, Karatzas, Manmatha, and Jawahar]{Mathew2020DocVQAAD}
Minesh Mathew, Dimosthenis Karatzas, R.~Manmatha, and C.~V. Jawahar.
\newblock Docvqa: A dataset for vqa on document images.
\newblock \emph{2021 IEEE Winter Conference on Applications of Computer Vision (WACV)}, pages 2199--2208, 2020.
\newblock URL \url{https://api.semanticscholar.org/CorpusID:220280200}.

\bibitem[Nakano et~al.(2021)Nakano, Hilton, Balaji, Wu, Ouyang, Kim, Hesse, Jain, Kosaraju, Saunders, et~al.]{nakano2021webgpt}
Reiichiro Nakano, Jacob Hilton, Suchir Balaji, Jeff Wu, Long Ouyang, Christina Kim, Christopher Hesse, Shantanu Jain, Vineet Kosaraju, William Saunders, et~al.
\newblock Webgpt: Browser-assisted question-answering with human feedback.
\newblock \emph{arXiv preprint arXiv:2112.09332}, 2021.

\bibitem[Nayak et~al.(2025)Nayak, Jian, Lin, Rodriguez, Kalsi, Awal, Chapados, Özsu, Agrawal, Vazquez, Pal, Taslakian, Gella, and Rajeswar]{nayak2025uivisiondesktopcentricguibenchmark}
Shravan Nayak, Xiangru Jian, Kevin~Qinghong Lin, Juan~A. Rodriguez, Montek Kalsi, Rabiul Awal, Nicolas Chapados, M.~Tamer Özsu, Aishwarya Agrawal, David Vazquez, Christopher Pal, Perouz Taslakian, Spandana Gella, and Sai Rajeswar.
\newblock Ui-vision: A desktop-centric gui benchmark for visual perception and interaction, 2025.
\newblock URL \url{https://arxiv.org/abs/2503.15661}.

\bibitem[OpenAI(2025)]{cua2025}
OpenAI.
\newblock Computer-using agent: Introducing a universal interface for ai to interact with the digital world.
\newblock 2025.
\newblock URL \url{https://openai.com/index/computer-using-agent}.

\bibitem[Qi et~al.(2024)Qi, Liu, Iong, Lai, Sun, Zhao, Yang, Yang, Sun, Yao, et~al.]{qi2024webrl}
Zehan Qi, Xiao Liu, Iat~Long Iong, Hanyu Lai, Xueqiao Sun, Wenyi Zhao, Yu~Yang, Xinyue Yang, Jiadai Sun, Shuntian Yao, et~al.
\newblock Webrl: Training llm web agents via self-evolving online curriculum reinforcement learning.
\newblock \emph{arXiv preprint arXiv:2411.02337}, 2024.

\bibitem[Qin et~al.(2025)Qin, Ye, Fang, Wang, Liang, Tian, Zhang, Li, Li, Huang, Zhong, Li, Yang, Miao, Lin, Liu, Jiang, Ma, Li, Xiao, Cai, Li, Zheng, Jin, Li, Zhou, Wang, Chen, Li, Yang, Liu, Lin, Peng, Liu, and Shi]{qin2025uitarspioneeringautomatedgui}
Yujia Qin, Yining Ye, Junjie Fang, Haoming Wang, Shihao Liang, Shizuo Tian, Junda Zhang, Jiahao Li, Yunxin Li, Shijue Huang, Wanjun Zhong, Kuanye Li, Jiale Yang, Yu~Miao, Woyu Lin, Longxiang Liu, Xu~Jiang, Qianli Ma, Jingyu Li, Xiaojun Xiao, Kai Cai, Chuang Li, Yaowei Zheng, Chaolin Jin, Chen Li, Xiao Zhou, Minchao Wang, Haoli Chen, Zhaojian Li, Haihua Yang, Haifeng Liu, Feng Lin, Tao Peng, Xin Liu, and Guang Shi.
\newblock Ui-tars: Pioneering automated gui interaction with native agents, 2025.
\newblock URL \url{https://arxiv.org/abs/2501.12326}.

\bibitem[Rawles et~al.(2023)Rawles, Li, Rodriguez, Riva, and Lillicrap]{rawles2023android}
Christopher Rawles, Alice Li, Daniel Rodriguez, Oriana Riva, and Timothy Lillicrap.
\newblock Android in the wild: A large-scale dataset for android device control.
\newblock \emph{arXiv preprint arXiv:2307.10088}, 2023.

\bibitem[Shi et~al.(2017)Shi, Karpathy, Fan, Hernandez, and Liang]{shi2017world}
Tianlin Shi, Andrej Karpathy, Linxi Fan, Jonathan Hernandez, and Percy Liang.
\newblock World of bits: An open-domain platform for web-based agents.
\newblock In \emph{International Conference on Machine Learning}, pages 3135--3144. PMLR, 2017.

\bibitem[Shridhar et~al.(2020)Shridhar, Thomason, Gordon, Bisk, Han, Mottaghi, Zettlemoyer, and Fox]{shridhar2020alfred}
Mohit Shridhar, Jesse Thomason, Daniel Gordon, Yonatan Bisk, Winson Han, Roozbeh Mottaghi, Luke Zettlemoyer, and Dieter Fox.
\newblock Alfred: A benchmark for interpreting grounded instructions for everyday tasks.
\newblock In \emph{Proceedings of the IEEE/CVF conference on computer vision and pattern recognition}, pages 10740--10749, 2020.

\bibitem[Sun et~al.(2024)Sun, Cheng, Ding, Jin, Wang, Xu, Wu, Jia, Chen, Liu, et~al.]{sun2024genesis}
Qiushi Sun, Kanzhi Cheng, Zichen Ding, Chuanyang Jin, Yian Wang, Fangzhi Xu, Zhenyu Wu, Chengyou Jia, Liheng Chen, Zhoumianze Liu, et~al.
\newblock Os-genesis: Automating gui agent trajectory construction via reverse task synthesis.
\newblock \emph{arXiv preprint arXiv:2412.19723}, 2024.

\bibitem[{SuperAGI}(2023)]{superagi2023guide}
{SuperAGI}.
\newblock Guide.
\newblock Hugging Face Datasets, 2023.
\newblock URL \url{https://huggingface.co/datasets/SuperAGI/GUIDE}.
\newblock Apache 2.0 License.

\bibitem[Team(2025)]{seed2025seed1_5vl}
ByteDance~Seed Team.
\newblock Seed1.5-vl technical report.
\newblock \emph{arXiv preprint arXiv:2505.07062}, 2025.

\bibitem[Team et~al.(2025)Team, Du, Yin, Xing, Qu, Wang, Chen, Zhang, Du, Wei, Wang, Zhang, Du, Wang, Yuan, Lu, Li, Sung, Wei, Lai, Zhu, Ding, Hu, Yang, Zhang, Wu, Yao, Lu, Wang, Gao, Zheng, Li, Su, Wang, Deng, Qiu, Xie, Wang, Liu, Yan, Ouyang, Chen, Sui, Yu, Dong, Dong, Xu, Cheng, Gu, Zhou, Liu, Cao, Yu, Song, Bai, Song, He, Huang, Xu, Yuan, Yao, Wu, Zu, Zhou, Wang, Charles, Zhong, Li, Hu, Chen, Wang, Liu, Miao, Qin, Chen, Bao, Wang, Kang, Liu, Du, Wu, Wang, Yan, Zhou, Li, Jiang, Zhang, Yang, Huang, Huang, Zhao, and Chen]{kimiteam2025kimivltechnicalreport}
Kimi Team, Angang Du, Bohong Yin, Bowei Xing, Bowen Qu, Bowen Wang, Cheng Chen, Chenlin Zhang, Chenzhuang Du, Chu Wei, Congcong Wang, Dehao Zhang, Dikang Du, Dongliang Wang, Enming Yuan, Enzhe Lu, Fang Li, Flood Sung, Guangda Wei, Guokun Lai, Han Zhu, Hao Ding, Hao Hu, Hao Yang, Hao Zhang, Haoning Wu, Haotian Yao, Haoyu Lu, Heng Wang, Hongcheng Gao, Huabin Zheng, Jiaming Li, Jianlin Su, Jianzhou Wang, Jiaqi Deng, Jiezhong Qiu, Jin Xie, Jinhong Wang, Jingyuan Liu, Junjie Yan, Kun Ouyang, Liang Chen, Lin Sui, Longhui Yu, Mengfan Dong, Mengnan Dong, Nuo Xu, Pengyu Cheng, Qizheng Gu, Runjie Zhou, Shaowei Liu, Sihan Cao, Tao Yu, Tianhui Song, Tongtong Bai, Wei Song, Weiran He, Weixiao Huang, Weixin Xu, Xiaokun Yuan, Xingcheng Yao, Xingzhe Wu, Xinxing Zu, Xinyu Zhou, Xinyuan Wang, Y.~Charles, Yan Zhong, Yang Li, Yangyang Hu, Yanru Chen, Yejie Wang, Yibo Liu, Yibo Miao, Yidao Qin, Yimin Chen, Yiping Bao, Yiqin Wang, Yongsheng Kang, Yuanxin Liu, Yulun Du, Yuxin Wu, Yuzhi Wang, Yuzi Yan, Zaida Zhou, Zhaowei Li, Zhejun
  Jiang, Zheng Zhang, Zhilin Yang, Zhiqi Huang, Zihao Huang, Zijia Zhao, and Ziwei Chen.
\newblock {Kimi-VL} technical report, 2025.
\newblock URL \url{https://arxiv.org/abs/2504.07491}.

\bibitem[Toyama et~al.(2021)Toyama, Hamel, Gergely, Comanici, Glaese, Ahmed, Jackson, Mourad, and Precup]{toyama2021androidenv}
Daniel Toyama, Philippe Hamel, Anita Gergely, Gheorghe Comanici, Amelia Glaese, Zafarali Ahmed, Tyler Jackson, Shibl Mourad, and Doina Precup.
\newblock Androidenv: A reinforcement learning platform for android.
\newblock \emph{arXiv preprint arXiv:2105.13231}, 2021.

\bibitem[Wang et~al.(2025{\natexlab{a}})Wang, Wang, Deng, Xie, Li, Zhang, Li, Hua, Stoica, Chiang, Yang, Su, Zhang, Wang, Zhong, and Yu]{wang2025computer}
Bowen Wang, Xinyuan Wang, Jiaqi Deng, Tianbao Xie, Ryan Li, Yanzhe Zhang, Gavin Li, Toh~Jing Hua, Ion Stoica, Wei-Lin Chiang, Diyi Yang, Yu~Su, Yi~Zhang, Zhiguo Wang, Victor Zhong, and Tao Yu.
\newblock Computer agent arena: Compare \& test computer use agents on crowdsourced real-world tasks, 2025{\natexlab{a}}.

\bibitem[Wang et~al.(2024)Wang, Bai, Tan, Wang, Fan, Bai, Chen, Liu, Wang, Ge, Fan, Dang, Du, Ren, Men, Liu, Zhou, Zhou, and Lin]{wang2024qwen2vlenhancingvisionlanguagemodels}
Peng Wang, Shuai Bai, Sinan Tan, Shijie Wang, Zhihao Fan, Jinze Bai, Keqin Chen, Xuejing Liu, Jialin Wang, Wenbin Ge, Yang Fan, Kai Dang, Mengfei Du, Xuancheng Ren, Rui Men, Dayiheng Liu, Chang Zhou, Jingren Zhou, and Junyang Lin.
\newblock Qwen2-vl: Enhancing vision-language model's perception of the world at any resolution, 2024.
\newblock URL \url{https://arxiv.org/abs/2409.12191}.

\bibitem[Wang et~al.(2025{\natexlab{b}})Wang, Wang, Lu, Yang, Xie, Wang, Deng, Guo, Xu, Wu, Shen, Li, Li, Li, Chen, Zheng, Li, Lei, Cao, Fu, Shin, Shin, Hu, Wang, Chen, Ye, Zhang, Du, Hu, Chen, Zhou, Yao, Chen, Gu, Wang, Wang, Yang, Zhong, Sung, Charles, Yang, and Yu]{wang2025opencuaopenfoundationscomputeruse}
Xinyuan Wang, Bowen Wang, Dunjie Lu, Junlin Yang, Tianbao Xie, Junli Wang, Jiaqi Deng, Xiaole Guo, Yiheng Xu, Chen~Henry Wu, Zhennan Shen, Zhuokai Li, Ryan Li, Xiaochuan Li, Junda Chen, Boyuan Zheng, Peihang Li, Fangyu Lei, Ruisheng Cao, Yeqiao Fu, Dongchan Shin, Martin Shin, Jiarui Hu, Yuyan Wang, Jixuan Chen, Yuxiao Ye, Danyang Zhang, Dikang Du, Hao Hu, Huarong Chen, Zaida Zhou, Haotian Yao, Ziwei Chen, Qizheng Gu, Yipu Wang, Heng Wang, Diyi Yang, Victor Zhong, Flood Sung, Y.~Charles, Zhilin Yang, and Tao Yu.
\newblock Opencua: Open foundations for computer-use agents, 2025{\natexlab{b}}.
\newblock URL \url{https://arxiv.org/abs/2508.09123}.

\bibitem[Wu et~al.(2023)Wu, Wang, Shen, Peng, Nichols, and Bigham]{wu2023webui}
Jason Wu, Siyan Wang, Siman Shen, Yi-Hao Peng, Jeffrey Nichols, and Jeffrey~P Bigham.
\newblock Webui: A dataset for enhancing visual ui understanding with web semantics.
\newblock In \emph{Proceedings of the 2023 CHI Conference on Human Factors in Computing Systems}, pages 1--14, 2023.

\bibitem[Wu et~al.(2024)Wu, Wu, Xu, Wang, Sun, Jia, Cheng, Ding, Chen, Liang, and Qiao]{Wu2024OSATLASAF}
Zhiyong Wu, Zhenyu Wu, Fangzhi Xu, Yian Wang, Qiushi Sun, Chengyou Jia, Kanzhi Cheng, Zichen Ding, Liheng Chen, Paul~Pu Liang, and Yu~Qiao.
\newblock Os-atlas: A foundation action model for generalist gui agents.
\newblock \emph{ArXiv}, abs/2410.23218, 2024.
\newblock URL \url{https://api.semanticscholar.org/CorpusID:273696039}.

\bibitem[Xie et~al.(2024)Xie, Zhang, Chen, Li, Zhao, Cao, Hua, Cheng, Shin, Lei, Liu, Xu, Zhou, Savarese, Xiong, Zhong, and Yu]{xie2024osworldbenchmarkingmultimodalagents}
Tianbao Xie, Danyang Zhang, Jixuan Chen, Xiaochuan Li, Siheng Zhao, Ruisheng Cao, Toh~Jing Hua, Zhoujun Cheng, Dongchan Shin, Fangyu Lei, Yitao Liu, Yiheng Xu, Shuyan Zhou, Silvio Savarese, Caiming Xiong, Victor Zhong, and Tao Yu.
\newblock Osworld: Benchmarking multimodal agents for open-ended tasks in real computer environments, 2024.
\newblock URL \url{https://arxiv.org/abs/2404.07972}.

\bibitem[Xie et~al.(2025)Xie, Yuan, Zhang, Xiong, Shen, Zhou, Wang, Chen, Deng, Chen, Wang, Wu, Chen, Wang, Lu, Hu, and Yu]{osworld_verified}
Tianbao Xie, Mengqi Yuan, Danyang Zhang, Xinzhuang Xiong, Zhennan Shen, Zilong Zhou, Xinyuan Wang, Yanxu Chen, Jiaqi Deng, Junda Chen, Bowen Wang, Haoyuan Wu, Jixuan Chen, Junli Wang, Dunjie Lu, Hao Hu, and Tao Yu.
\newblock Introducing osworld-verified.
\newblock \emph{xlang.ai}, July 2025.
\newblock URL \url{https://xlang.ai/blog/osworld-verified}.

\bibitem[Xu et~al.(2024)Xu, Wang, Wang, Lu, Xie, Saha, Sahoo, Yu, and Xiong]{xu2024aguvis}
Yiheng Xu, Zekun Wang, Junli Wang, Dunjie Lu, Tianbao Xie, Amrita Saha, Doyen Sahoo, Tao Yu, and Caiming Xiong.
\newblock Aguvis: Unified pure vision agents for autonomous gui interaction.
\newblock \emph{arXiv preprint arXiv:2412.04454}, 2024.

\bibitem[Yang et~al.(2024)Yang, Wang, Li, Luo, Chen, Huang, and Li]{yang2024aria}
Yuhao Yang, Yue Wang, Dongxu Li, Ziyang Luo, Bei Chen, Chao Huang, and Junnan Li.
\newblock Aria-ui: Visual grounding for gui instructions.
\newblock \emph{arXiv preprint arXiv:2412.16256}, 2024.

\bibitem[Yao et~al.(2022)Yao, Chen, Yang, and Narasimhan]{yao2022webshop}
Shunyu Yao, Howard Chen, John Yang, and Karthik Narasimhan.
\newblock Webshop: Towards scalable real-world web interaction with grounded language agents.
\newblock \emph{Advances in Neural Information Processing Systems}, 35:\penalty0 20744--20757, 2022.

\bibitem[Yao et~al.(2024)Yao, Yu, Zhang, Wang, Cui, Zhu, Cai, Li, Zhao, He, Chen, Zhou, Zou, Zhang, Hu, Zheng, Zhou, Cai, Han, Zeng, Li, Liu, and Sun]{yao2024minicpmvgpt4vlevelmllm}
Yuan Yao, Tianyu Yu, Ao~Zhang, Chongyi Wang, Junbo Cui, Hongji Zhu, Tianchi Cai, Haoyu Li, Weilin Zhao, Zhihui He, Qianyu Chen, Huarong Zhou, Zhensheng Zou, Haoye Zhang, Shengding Hu, Zhi Zheng, Jie Zhou, Jie Cai, Xu~Han, Guoyang Zeng, Dahai Li, Zhiyuan Liu, and Maosong Sun.
\newblock Minicpm-v: A gpt-4v level mllm on your phone, 2024.
\newblock URL \url{https://arxiv.org/abs/2408.01800}.

\bibitem[Yu et~al.(2025)Yu, Yang, Wan, Song, Tang, Cheng, Liu, and Bai]{yu2025omniparser}
Wenwen Yu, Zhibo Yang, Jianqiang Wan, Sibo Song, Jun Tang, Wenqing Cheng, Yuliang Liu, and Xiang Bai.
\newblock Omniparser v2: Structured-points-of-thought for unified visual text parsing and its generality to multimodal large language models.
\newblock \emph{arXiv preprint arXiv:2502.16161}, 2025.

\bibitem[Zhang et~al.(2023)Zhang, Chen, and Yu]{zhang2023mobile}
Danyang Zhang, Lu~Chen, and Kai Yu.
\newblock Mobile-env: A universal platform for training and evaluation of mobile interaction.
\newblock \emph{arXiv preprint arXiv:2305.08144}, 2023.

\bibitem[Zhang et~al.(2024)Zhang, Wu, Teng, Liao, Xu, Xiao, Wei, and Tang]{zhang2024android}
Jiwen Zhang, Jihao Wu, Yihua Teng, Minghui Liao, Nuo Xu, Xiao Xiao, Zhongyu Wei, and Duyu Tang.
\newblock Android in the zoo: Chain-of-action-thought for gui agents.
\newblock \emph{arXiv preprint arXiv:2403.02713}, 2024.

\bibitem[Zheng et~al.(2024)Zheng, Gou, Kil, Sun, and Su]{zheng2024gpt}
Boyuan Zheng, Boyu Gou, Jihyung Kil, Huan Sun, and Yu~Su.
\newblock Gpt-4v (ision) is a generalist web agent, if grounded.
\newblock \emph{arXiv preprint arXiv:2401.01614}, 2024.

\bibitem[Zhou et~al.(2023)Zhou, Xu, Zhu, Zhou, Lo, Sridhar, Cheng, Bisk, Fried, Alon, et~al.]{zhou2023webarena}
Shuyan Zhou, Frank~F Xu, Hao Zhu, Xuhui Zhou, Robert Lo, Abishek Sridhar, Xianyi Cheng, Yonatan Bisk, Daniel Fried, Uri Alon, et~al.
\newblock Webarena: A realistic web environment for building autonomous agents.
\newblock \emph{arXiv preprint arXiv:2307.13854}, 2023.

\end{thebibliography}
